\PassOptionsToPackage{dvipsnames}{xcolor} %
\documentclass[]{fairmeta}

\newif\ifshowtodos  %
\showtodosfalse  %

\setcitestyle{sort}

\usepackage{amsmath}
\usepackage{amssymb}
\usepackage{mathtools}
\usepackage{xspace}
\usepackage{wrapfig}
\usepackage{makecell}
\usepackage{colortbl}
\usepackage[most]{tcolorbox}
\usepackage[shortlabels]{enumitem}
\usepackage{needspace}
\usepackage[dvipsnames]{xcolor}
\usepackage{fdsymbol} %
\usepackage{pdflscape} %

\usepackage{tikz}
\definecolor{ScratchEncoderBlue}{HTML}{416FF6}
\definecolor{ScratchEncoderBlueFill}{HTML}{C7D7FC}

\newlength{\paragraphabove}
\makeatletter
\renewcommand\paragraph{\@startsection{paragraph}{4}{\z@}{\paragraphabove}{-1em}{\normalfont\normalsize\bfseries}}
\makeatother

\makeatletter
\@ifundefined{ifshowtodos}{%
    \newif\ifshowtodos
    \showtodosfalse
}{}
\makeatother
\ifshowtodos
    \newcommand{\todotxt}[1]{#1}
\else
    \newcommand{\todotxt}[1]{}
\fi
\newcommand{\todo}[1]{}

\definecolor{PaperBlue100}{HTML}{C3D8FF}
\definecolor{PaperBlue200}{HTML}{B0CBFF}
\definecolor{PaperBlue300}{HTML}{9CBEFF}
\definecolor{PaperBlue400}{HTML}{88B0FF}
\definecolor{PaperBlue500}{HTML}{74A2FF}
\definecolor{PaperBlue600}{HTML}{5088FF}
\definecolor{PaperBlue700}{HTML}{2F70FF}
\definecolor{PaperGreen100}{HTML}{CBDFBE}
\definecolor{PaperGreen200}{HTML}{BBD4AA}
\definecolor{PaperGreen300}{HTML}{ABC997}
\definecolor{PaperGreen400}{HTML}{9BBE83}
\definecolor{PaperGreen500}{HTML}{83B476}
\definecolor{PaperGreen600}{HTML}{51A556}
\definecolor{PaperGreen700}{HTML}{299236}
\definecolor{PaperAmber100}{HTML}{F9CB8A}
\definecolor{PaperAmber200}{HTML}{F8C277}
\definecolor{PaperAmber300}{HTML}{F7BA64}
\definecolor{PaperAmber400}{HTML}{F6B150}
\definecolor{PaperAmber500}{HTML}{F5A83D}
\definecolor{PaperAmber600}{HTML}{EFA200}
\definecolor{PaperAmber700}{HTML}{EA9600}
\definecolor{PaperCoral100}{HTML}{FFC8C3}
\definecolor{PaperCoral200}{HTML}{FDB5AF}
\definecolor{PaperCoral300}{HTML}{FAA39C}
\definecolor{PaperCoral400}{HTML}{F69088}
\definecolor{PaperCoral500}{HTML}{F27D76}
\definecolor{PaperCoral600}{HTML}{F45A56}
\definecolor{PaperCoral700}{HTML}{EE343B}
\definecolor{PaperViolet100}{HTML}{E4CDF6}
\definecolor{PaperViolet200}{HTML}{DABDF1}
\definecolor{PaperViolet300}{HTML}{D1ADEC}
\definecolor{PaperViolet400}{HTML}{C79CE7}
\definecolor{PaperViolet500}{HTML}{B68BE0}
\definecolor{PaperViolet600}{HTML}{A169DA}
\definecolor{PaperViolet700}{HTML}{8F48D7}
\definecolor{PaperPink100}{HTML}{F5C9DD}
\definecolor{PaperPink200}{HTML}{F4B6D2}
\definecolor{PaperPink300}{HTML}{F1A2C7}
\definecolor{PaperPink400}{HTML}{EE8DBD}
\definecolor{PaperPink500}{HTML}{EB78B2}
\definecolor{PaperPink600}{HTML}{E24EA0}
\definecolor{PaperPink700}{HTML}{D72090}
\definecolor{PaperTeal100}{HTML}{B2E3DF}
\definecolor{PaperTeal200}{HTML}{99D9D4}
\definecolor{PaperTeal300}{HTML}{7ECFC9}
\definecolor{PaperTeal400}{HTML}{60C5BE}
\definecolor{PaperTeal500}{HTML}{39BAB4}
\definecolor{PaperTeal600}{HTML}{00A19B}
\definecolor{PaperTeal700}{HTML}{008681}
\definecolor{PaperOrange100}{HTML}{F8CEB2}
\definecolor{PaperOrange200}{HTML}{F6BD97}
\definecolor{PaperOrange300}{HTML}{F3AB7A}
\definecolor{PaperOrange400}{HTML}{F09A5C}
\definecolor{PaperOrange500}{HTML}{EB883B}
\definecolor{PaperOrange600}{HTML}{F7791A}
\definecolor{PaperOrange700}{HTML}{EB6200}
\definecolor{PaperSand100}{HTML}{E7D4BB}
\definecolor{PaperSand200}{HTML}{E0C6A3}
\definecolor{PaperSand300}{HTML}{D9B88B}
\definecolor{PaperSand400}{HTML}{D1AA73}
\definecolor{PaperSand500}{HTML}{C99B5A}
\definecolor{PaperSand600}{HTML}{BA7F14}
\definecolor{PaperSand700}{HTML}{9C6900}
\definecolor{PaperGray100}{HTML}{D5D8DB}
\definecolor{PaperGray200}{HTML}{C7CBD0}
\definecolor{PaperGray300}{HTML}{BABEC4}
\definecolor{PaperGray400}{HTML}{ADB1B9}
\definecolor{PaperGray500}{HTML}{A0A5AE}
\definecolor{PaperGray600}{HTML}{788190}
\definecolor{PaperGray700}{HTML}{4C5666}

\definecolor{navyblue}{HTML}{0071BC}
\definecolor{methodgreen}{RGB}{45,106,79}
\definecolor{fairblue}{HTML}{0064E0}
\definecolor{nyupurple}{HTML}{57068C}
\definecolor{pos}{RGB}{0,128,0}
\definecolor{neg}{RGB}{200,0,0}

\usepackage{tikz}
\makeatletter
\newcommand{\recipeglyph}[1]{\makebox[0.95em][l]{\ifcsname recipeglyph@#1\endcsname\csname recipeglyph@#1\endcsname\fi}}
\newcommand{\recipemark}[1]{\ifcsname recipeglyph@#1\endcsname\raisebox{\ifcsname recipemarkraise@#1\endcsname\csname recipemarkraise@#1\endcsname\else 0ex\fi}{\csname recipeglyph@#1\endcsname}\fi}
\newcommand{\recipemarkraise@scratch}{0.25ex}
\newcommand{\recipemarkraise@metaclip}{0.19ex}
\newcommand{\recipemarkraise@encoderfree}{0.27ex}
\newcommand{\recipemarkraise@infusionffn}{0.2ex}
\newcommand{\recipemarkraise@webssl}{0.1ex}
\newcommand{\recipemarkraise@cap}{0.1ex}
\newcommand{\recipemarkraise@cappa}{0.12ex}
\DeclareRobustCommand{\hlrow}[1]{{\setlength{\fboxsep}{1pt}\colorbox{PaperPink100!40}{\strut #1}}}
\DeclareRobustCommand{\selectedmark}{\tikz[baseline=-0.55ex]{\fill[PaperTeal700] (0,0) circle (0.62ex); \draw[white, line width=0.45pt, line cap=round, line join=round] (-0.32ex,-0.02ex) -- (-0.08ex,-0.26ex) -- (0.34ex,0.22ex);}}
\DeclareRobustCommand{\tokenglyph}[1]{\ifcsname tokenglyph@#1\endcsname\csname tokenglyph@#1\endcsname\fi}
\newcommand{\tokenglyph@vision}{\tikz[baseline=0pt]\fill[PaperBlue600, rounded corners=0.18ex] (0,0) rectangle (0.9ex,0.9ex);}
\newcommand{\tokenglyph@text}{\tikz[baseline=0pt]\fill[PaperCoral500, rounded corners=0.18ex] (0,0) rectangle (0.9ex,0.9ex);}
\newcommand{\panel}[1]{\textbf{(#1)}}
\newcommand{\recipeglyph@scratch}{\tikz[baseline=0pt]\fill[PaperBlue700] (0.36ex,0.36ex) circle (0.36ex);}
\newcommand{\recipeglyph@metaclip}{\tikz[baseline=0pt]\fill[PaperOrange700] (0.42ex,0) -- (0.84ex,0.42ex) -- (0.42ex,0.84ex) -- (0,0.42ex) -- cycle;}
\newcommand{\recipeglyph@encoderfree}{\tikz[baseline=0pt]\fill[PaperAmber700] (0,0) rectangle (0.68ex,0.68ex);}
\newcommand{\recipeglyph@infusionffn}{\tikz[baseline=0pt]\fill[PaperPink400] (0,0) -- (0.84ex,0) -- (0.42ex,0.76ex) -- cycle;}
\newcommand{\recipeglyph@infusionqkvo}{\tikz[baseline=0pt]\fill[PaperPink700] (0,0.76ex) -- (0.84ex,0.76ex) -- (0.42ex,0) -- cycle;}
\newcommand{\recipeglyph@webssl}{\tikz[baseline=0pt]{\fill[PaperGreen700] (0.27ex,0) rectangle (0.53ex,0.8ex); \fill[PaperGreen700] (0,0.27ex) rectangle (0.8ex,0.53ex);}}
\newcommand{\recipeglyph@cap}{\tikz[baseline=0pt]\fill[PaperSand500,shift={(0.42ex,0.4ex)}] (0:0.42ex) -- (60:0.42ex) -- (120:0.42ex) -- (180:0.42ex) -- (240:0.42ex) -- (300:0.42ex) -- cycle;}
\newcommand{\recipeglyph@cappa}{\tikz[baseline=0pt]\fill[PaperSand700,shift={(0.42ex,0.38ex)}] (90:0.44ex) -- (162:0.44ex) -- (234:0.44ex) -- (306:0.44ex) -- (18:0.44ex) -- cycle;}
\makeatother

\newtcolorbox{takeawaybox}[1][]{%
  enhanced jigsaw,
  colback=PaperTeal100!15,
  colframe=PaperTeal700,
  boxrule=0.55pt,
  arc=4pt,
  left=6pt, right=6pt,
  top=4pt, bottom=4pt,
  before skip=3pt,
  after skip=3pt,
  fontupper=\small,
  #1
}

\newcommand{\eb}[1]{}       %
\newcommand{\ch}[1]{}        %
\newcommand{\sx}[1]{}       %
\newcommand{\hu}[1]{}      %
\newcommand{\mr}[1]{}      %
\newcommand{\jv}[1]{}       %
\newcommand{\sq}[1]{}  %
\newcommand{\dl}[1]{}      %
\newcommand{\lb}[1]{}  %
\newcommand{\ca}[1]{}  %

\makeatletter
\DeclareRobustCommand\onedot{\futurelet\@let@token\@onedot}
\def\@onedot{\ifx\@let@token.\else.\null\fi\xspace}
\def\eg{e.g\onedot}          
\def\ie{i.e\onedot}          
           
         \def\vs{vs\onedot}

\makeatother


\newcommand{\equalfirstmark}{\textcolor{black}{*}}
\newcommand{\equalseniormark}{\textcolor{black}{\dagger}}

\newcommand{\nyutorchpath}{(0.25024,0.00034) .. controls (0.24906,0.00043) and (0.24701,0.00072) .. (0.24568,0.00098) -- (0.24328,0.00146) -- (0.24125,0.02087) .. controls (0.23646,0.06673) and (0.22868,0.14389) .. (0.21998,0.23189) .. controls (0.20622,0.37108) and (0.20501,0.38329) .. (0.20472,0.38527) .. controls (0.20450,0.38681) and (0.20466,0.38729) .. (0.20578,0.38840) -- (0.20710,0.38972) -- (0.24786,0.38959) .. controls (0.28367,0.38948) and (0.30689,0.38917) .. (0.30726,0.38880) .. controls (0.30732,0.38874) and (0.30579,0.37228) .. (0.30386,0.35223) .. controls (0.30192,0.33218) and (0.29882,0.29990) .. (0.29696,0.28051) .. controls (0.28789,0.18597) and (0.28135,0.12320) .. (0.27476,0.06745) .. controls (0.27224,0.04606) and (0.26703,0.00461) .. (0.26665,0.00276) .. controls (0.26650,0.00205) and (0.26602,0.00137) .. (0.26558,0.00125) .. controls (0.26301,0.00053) and (0.25453,0.00003) .. (0.25024,0.00034) -- cycle (0.20861,0.40735) .. controls (0.20857,0.40738) and (0.19995,0.40754) .. (0.18947,0.40771) .. controls (0.16844,0.40805) and (0.15770,0.40844) .. (0.15721,0.40890) .. controls (0.15636,0.40970) and (0.15594,0.42244) .. (0.15602,0.44528) -- (0.15610,0.47041) -- (0.16420,0.47057) .. controls (0.17628,0.47082) and (0.32632,0.47054) .. (0.32936,0.47026) .. controls (0.33080,0.47013) and (0.33776,0.46991) .. (0.34483,0.46976) .. controls (0.35190,0.46961) and (0.35786,0.46931) .. (0.35808,0.46909) .. controls (0.35874,0.46843) and (0.35873,0.42386) .. (0.35806,0.41659) .. controls (0.35773,0.41305) and (0.35736,0.40962) .. (0.35723,0.40896) -- (0.35698,0.40777) -- (0.29443,0.40764) .. controls (0.26003,0.40756) and (0.22667,0.40745) .. (0.22029,0.40740) .. controls (0.21391,0.40734) and (0.20865,0.40732) .. (0.20861,0.40735) -- cycle (0.16611,0.49218) .. controls (0.16506,0.49228) and (0.16152,0.49260) .. (0.15825,0.49289) .. controls (0.13538,0.49493) and (0.11011,0.50212) .. (0.08945,0.51247) .. controls (0.05490,0.52977) and (0.02708,0.55714) .. (0.01361,0.58707) .. controls (0.00465,0.60699) and (0.00034,0.62739) .. (0.00030,0.65014) .. controls (0.00027,0.66512) and (0.00148,0.67518) .. (0.00506,0.68962) .. controls (0.01061,0.71194) and (0.01940,0.73192) .. (0.04148,0.77240) .. controls (0.05714,0.80110) and (0.06653,0.82062) .. (0.07218,0.83620) .. controls (0.07700,0.84951) and (0.08062,0.86380) .. (0.08230,0.87616) -- (0.08312,0.88220) -- (0.08449,0.88149) .. controls (0.09108,0.87808) and (0.09819,0.86617) .. (0.10796,0.84217) .. controls (0.10971,0.83788) and (0.11337,0.82951) .. (0.11609,0.82358) .. controls (0.12522,0.80366) and (0.12702,0.79820) .. (0.12700,0.79051) .. controls (0.12699,0.78363) and (0.12571,0.78041) .. (0.11617,0.76339) .. controls (0.11121,0.75453) and (0.10846,0.74856) .. (0.10295,0.73469) .. controls (0.09509,0.71487) and (0.09082,0.69888) .. (0.08834,0.67993) .. controls (0.08721,0.67128) and (0.08722,0.65178) .. (0.08836,0.64323) .. controls (0.09354,0.60456) and (0.10932,0.57247) .. (0.13625,0.54584) .. controls (0.14904,0.53320) and (0.16119,0.52373) .. (0.19104,0.50312) .. controls (0.19820,0.49818) and (0.20388,0.49399) .. (0.20367,0.49380) .. controls (0.20253,0.49277) and (0.17313,0.49150) .. (0.16611,0.49218) -- cycle (0.29330,0.49370) -- (0.29084,0.49385) -- (0.29049,0.49847) .. controls (0.28994,0.50582) and (0.29060,0.52455) .. (0.29170,0.53289) .. controls (0.29627,0.56732) and (0.30893,0.59775) .. (0.33038,0.62590) .. controls (0.33657,0.63401) and (0.34096,0.63896) .. (0.34957,0.64753) .. controls (0.36036,0.65827) and (0.37017,0.66627) .. (0.38298,0.67479) .. controls (0.39200,0.68078) and (0.39760,0.68383) .. (0.42156,0.69578) .. controls (0.45515,0.71253) and (0.47082,0.72095) .. (0.48622,0.73050) .. controls (0.51966,0.75125) and (0.54218,0.77230) .. (0.55810,0.79766) .. controls (0.56197,0.80382) and (0.56807,0.81619) .. (0.57069,0.82315) .. controls (0.57240,0.82772) and (0.57299,0.82877) .. (0.57421,0.82948) .. controls (0.57546,0.83021) and (0.57588,0.83025) .. (0.57698,0.82975) -- (0.57828,0.82916) -- (0.57797,0.82080) .. controls (0.57575,0.75933) and (0.56498,0.70943) .. (0.54369,0.66197) .. controls (0.52177,0.61310) and (0.48680,0.56674) .. (0.45192,0.54032) .. controls (0.41688,0.51378) and (0.37122,0.49761) .. (0.32197,0.49430) .. controls (0.31415,0.49377) and (0.29768,0.49343) .. (0.29330,0.49370) -- cycle (0.24190,0.49791) .. controls (0.21565,0.50259) and (0.19035,0.51779) .. (0.15706,0.54889) .. controls (0.12844,0.57563) and (0.11200,0.60335) .. (0.10607,0.63489) .. controls (0.10312,0.65056) and (0.10273,0.67175) .. (0.10512,0.68675) .. controls (0.10760,0.70234) and (0.11341,0.71913) .. (0.12320,0.73904) .. controls (0.13154,0.75598) and (0.14282,0.77494) .. (0.15992,0.80076) .. controls (0.17521,0.82384) and (0.18466,0.83668) .. (0.21043,0.86940) .. controls (0.23579,0.90159) and (0.24319,0.91148) .. (0.25342,0.92684) .. controls (0.26917,0.95049) and (0.28014,0.97209) .. (0.28670,0.99237) .. controls (0.28845,0.99778) and (0.28985,1.00000) .. (0.29151,1.00000) .. controls (0.29235,1.00000) and (0.30143,0.98189) .. (0.30761,0.96789) .. controls (0.31940,0.94117) and (0.33040,0.91129) .. (0.33944,0.88132) .. controls (0.34681,0.85693) and (0.34861,0.84839) .. (0.34864,0.83769) .. controls (0.34867,0.83058) and (0.34866,0.83052) .. (0.34714,0.82743) .. controls (0.34493,0.82295) and (0.33921,0.81636) .. (0.32909,0.80665) .. controls (0.32421,0.80197) and (0.31827,0.79600) .. (0.31589,0.79337) .. controls (0.31351,0.79075) and (0.30565,0.78228) .. (0.29842,0.77454) .. controls (0.29119,0.76681) and (0.28301,0.75802) .. (0.28024,0.75500) .. controls (0.24928,0.72132) and (0.23083,0.69320) .. (0.22378,0.66892) .. controls (0.21423,0.63605) and (0.21560,0.58625) .. (0.22710,0.54845) .. controls (0.23005,0.53878) and (0.23239,0.53354) .. (0.23784,0.52451) .. controls (0.24275,0.51636) and (0.24559,0.51110) .. (0.24682,0.50786) .. controls (0.24766,0.50565) and (0.24879,0.50041) .. (0.24880,0.49869) .. controls (0.24881,0.49725) and (0.24870,0.49714) .. (0.24726,0.49718) .. controls (0.24641,0.49721) and (0.24399,0.49753) .. (0.24190,0.49791) -- cycle (0.26747,0.50655) .. controls (0.25579,0.51827) and (0.24503,0.54375) .. (0.23885,0.57436) .. controls (0.23614,0.58775) and (0.23570,0.59270) .. (0.23572,0.60939) .. controls (0.23574,0.62580) and (0.23631,0.63217) .. (0.23907,0.64681) .. controls (0.24134,0.65880) and (0.24789,0.68271) .. (0.25011,0.68707) .. controls (0.25171,0.69023) and (0.26213,0.70550) .. (0.26810,0.71345) .. controls (0.28189,0.73180) and (0.29624,0.74803) .. (0.32436,0.77712) .. controls (0.34753,0.80109) and (0.35371,0.80756) .. (0.36172,0.81627) .. controls (0.38078,0.83696) and (0.39507,0.85477) .. (0.40671,0.87234) .. controls (0.41420,0.88365) and (0.41907,0.89208) .. (0.42522,0.90440) .. controls (0.42969,0.91336) and (0.43184,0.91677) .. (0.43254,0.91604) .. controls (0.43313,0.91544) and (0.43754,0.90755) .. (0.43969,0.90328) .. controls (0.45221,0.87830) and (0.46352,0.83728) .. (0.46874,0.79790) .. controls (0.47260,0.76876) and (0.47323,0.75126) .. (0.47067,0.74400) .. controls (0.46972,0.74130) and (0.46643,0.73795) .. (0.46180,0.73498) .. controls (0.45773,0.73236) and (0.44632,0.72666) .. (0.43374,0.72096) .. controls (0.42916,0.71888) and (0.42142,0.71498) .. (0.41654,0.71229) .. controls (0.40217,0.70435) and (0.37159,0.68658) .. (0.36566,0.68272) .. controls (0.35400,0.67512) and (0.34121,0.66450) .. (0.33177,0.65458) .. controls (0.29791,0.61905) and (0.27805,0.57108) .. (0.27281,0.51218) .. controls (0.27236,0.50705) and (0.27180,0.50286) .. (0.27157,0.50286) .. controls (0.27134,0.50286) and (0.26950,0.50452) .. (0.26747,0.50655) -- cycle}

\DeclareRobustCommand{\fairmark}{\raisebox{0.05em}{\includegraphics[height=0.45em,trim=0 0 2590 0,clip]{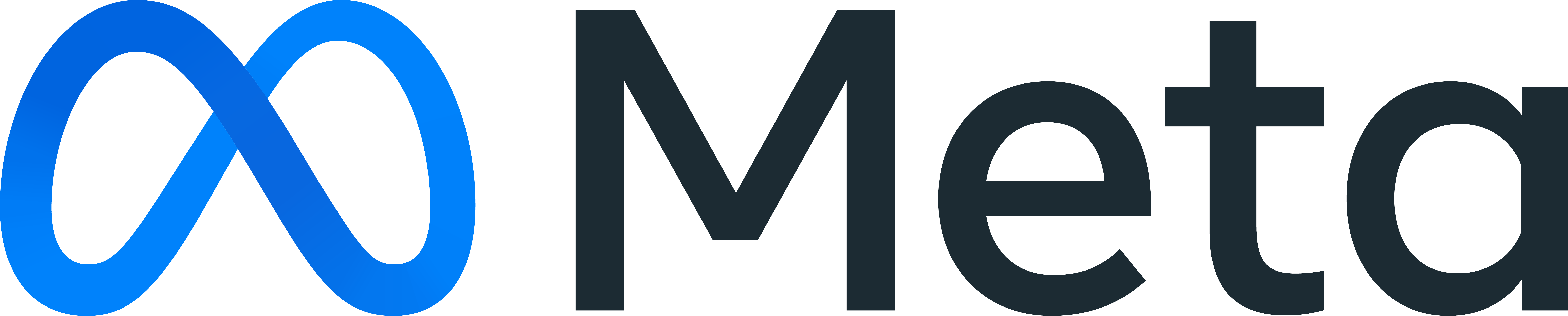}}}
\DeclareRobustCommand{\nyumark}{\tikz[baseline=0.125em,x=0.85em,y=0.85em]\fill[nyupurple,even odd rule] \nyutorchpath;}
\newcommand{\marksep}{\,}  %

\usepackage{etoolbox}
\newcommand{\titlelogos}{\includegraphics[width=1.5cm]{assets/logometa.pdf}\hspace{0.5cm}\raisebox{-0.068cm}[\height][0pt]{\includegraphics[height=0.438cm]{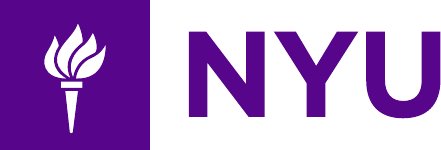}}}
\makeatletter
\@ifundefined{mymaketitle}{}{\patchcmd{\mymaketitle}{\includegraphics[width=1.5cm]{assets/logometa.pdf}}{\titlelogos}{}{\PackageWarning{preamble}{title-box logo patch failed}}}
\makeatother

\newcolumntype{g}{>{\columncolor{gray!30}}c}     %

\newcommand{\cmark}{\ensuremath{\checkmark}}    %

\newcommand{\boldcolortext}[2]{\textcolor{#1}{\textbf{#2}}}

\creflabelformat{equation}{#2#1#3}%
\hypersetup{
  pdftitle={Unbundling Multimodal Vision Encoding},
  pdfauthor={Ellis Brown, Chris Hoang, Shengyi Qian, Shang-Wen Li, Mike Lewis, Marjan Ghazvininejad, Luke Zettlemoyer, Lucas Beyer, Mengye Ren, Jakob Verbeek, Saining Xie, Hu Xu},
  pdfdisplaydoctitle=true,
}
\makeatletter
\renewcommand{\tokenglyph@text}{\tikz[baseline=0pt]\fill[PaperGreen500, rounded corners=0.18ex] (0,0) rectangle (0.9ex,0.9ex);}
\makeatother

\DeclareRobustCommand{\ours}{in-fusion vision capacity\xspace}
\DeclareRobustCommand{\oursffn}{in-fusion (FFN-only)\xspace}
\DeclareRobustCommand{\oursqkvo}{in-fusion\xspace}
\DeclareRobustCommand{\oursstyled}{In-fusion vision capacity\xspace}
\DeclareRobustCommand{\oursffnstyled}{In-fusion (FFN-only)\xspace}
\DeclareRobustCommand{\oursqkvostyled}{In-fusion\xspace}

\title{Unbundling Multimodal Vision Encoding}

\author[\equalfirstmark\marksep\fairmark\marksep\nyumark]{Ellis~Brown}
\author[\equalfirstmark\marksep\fairmark\marksep\nyumark]{Chris~Hoang}
\author[\fairmark]{\\[0.25em] Shengyi~Qian}
\author[\fairmark]{Shang-Wen~Li}
\author[\fairmark]{Mike~Lewis}
\author[\fairmark]{Marjan~Ghazvininejad}
\author[\fairmark]{Luke~Zettlemoyer}
\author[\fairmark]{Lucas~Beyer}
\author[\equalseniormark\marksep\nyumark]{\\[0.25em] Mengye~Ren}
\author[\equalseniormark\marksep\fairmark]{Jakob~Verbeek}
\author[\equalseniormark\marksep\nyumark]{Saining~Xie}
\author[\equalseniormark\marksep\fairmark]{Hu~Xu}

\affiliation[\fairmark]{\textcolor{fairblue}{Meta FAIR}}
\affiliation[\nyumark]{\textcolor{nyupurple}{New York University}}

\contribution[\equalfirstmark]{equal contribution, ordered randomly}
\contribution[\equalseniormark]{equal advising}

\abstract{

How multimodal language models should encode visual input remains up for debate.
Most models use a pretrained vision encoder, while some recent recipes train the encoder from scratch with the language model or remove it entirely.
Vision encoders bundle together three design choices:
    a \emph{visual prior} from upstream pretraining,
    \emph{dedicated vision capacity},
    and \emph{pre-fusion placement} of that capacity
    (\ie, before the multimodal backbone).
Because published models also differ in data, architecture, and compute, comparing them cannot isolate the effect of each choice.
We disentangle these choices via a controlled framework where we pretrain all models from scratch on common data and compare them on visual question answering and language benchmarks at matched \emph{lifetime compute}.
We find that the visual prior is not necessary; an encoder trained jointly from scratch is on par with a pretrained one, without a separate training stage.
What matters instead is dedicated vision capacity: an encoder-free model trails on visual understanding even when its backbone is enlarged by the encoder's parameter count.
Yet that capacity need not precede multimodal fusion, as vision-only modules added \emph{in-fusion} to early backbone layers perform comparably to a pre-fusion encoder and can lower time to first token.
These results also hold across backbone sizes, when encoder pretraining is treated as free, and when starting from a pretrained language model.
Overall, our findings favor dedicated vision capacity trained jointly with the backbone and placed early within it.

}

\date{November 12, 2026}

\metadata[Blog]{\url{https://unbundling.github.io}}

\begin{document}
\maketitle

\vspace{3em}

\begin{figure}[hb!]
  \centering
  \newcommand{\figonelink}[5]{\node[anchor=south west, inner sep=0] at (#2,#3) {\hyperref[#1]{\phantom{\rule{#4\textwidth}{#5\textwidth}}}};}
  \begin{tikzpicture}
    \node[anchor=south west, inner sep=0] (figone) at (0,0) {\includegraphics[width=\textwidth,trim=40 40 40 40,clip]{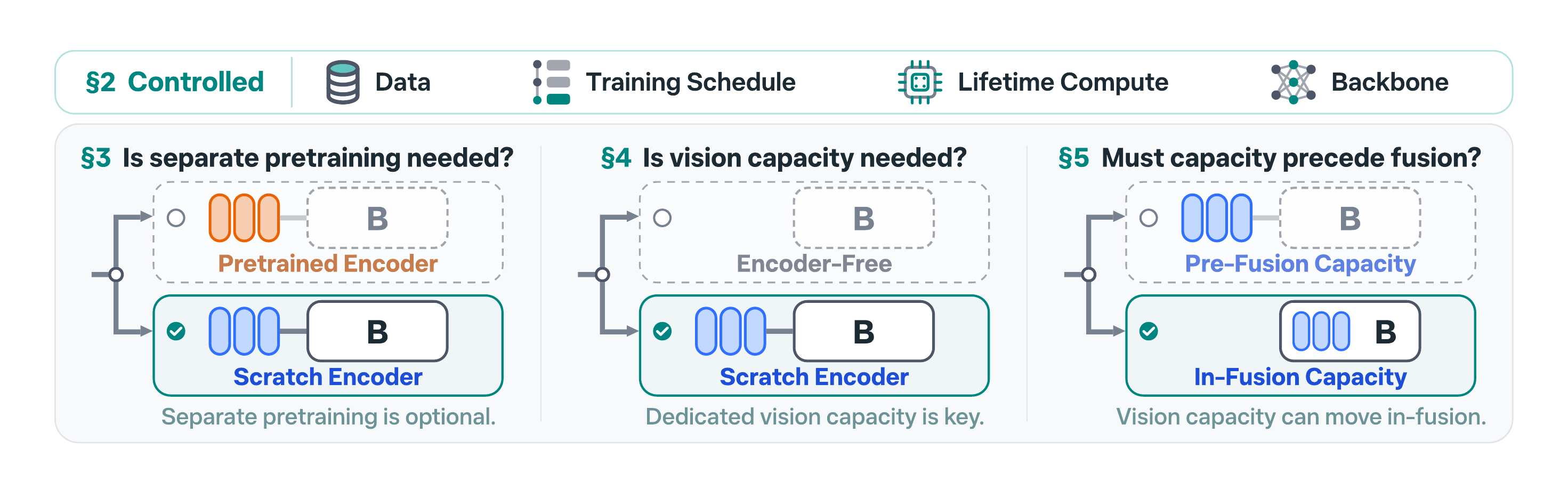}};
    \begin{scope}[x={(figone.south east)}, y={(figone.north west)}]
      \figonelink{sec:protocol}{0.0233}{0.8395}{0.1299}{0.0394}
      \figonelink{sec:encoders}{0.0199}{0.6553}{0.3019}{0.0382}
      \figonelink{sec:encoder-free}{0.3716}{0.6553}{0.2567}{0.0382}
      \figonelink{sec:capacity}{0.6809}{0.6553}{0.2951}{0.0382}
    \end{scope}
  \end{tikzpicture}
  \caption{
    \textbf{A controlled study of multimodal vision encoding.}
    (Top) Every recipe shares one backbone architecture (``B'') and trains all modules from scratch on common data at matched lifetime compute~(\S\ref{sec:protocol}).
    (Bottom) Each panel unbundles one vision encoder design choice, marks the favored recipe (\selectedmark), and states its takeaway; \boldcolortext{PaperOrange700}{orange} marks a pretrained prior and \boldcolortext{PaperBlue700}{blue} dedicated vision capacity.
    \S\ref{sec:practical} goes beyond lifetime compute by treating encoder pretraining as free, adding vision to a pretrained language model, and measuring training and inference speed.
  }
  \label{fig:intro:encoder-overview}
\end{figure}

\clearpage
{
  \hypersetup{linkcolor=black}
  \tableofcontents
}
\clearpage

\section{Introduction}
\label{sec:intro}

Most multimodal models acquire visual understanding from a vision encoder pretrained for a separate objective, such as image recognition or image--text alignment, rather than from being trained directly with the multimodal backbone that uses its features~\citep{he2016deep,radford2021clip,tschannen2025siglip2,simeoni2025dinov3,tschannen2023captioners}.
Multimodal training has progressed from freezing this encoder~\citep{alayrac2022flamingo,li2023blip2,liu2023visual,muse-glimmer} to updating it~\citep{beyer2024paligemma,bai2025qwen3vl}, but encoder pretraining has remained a separate stage.
Recent recipes bring vision and language closer by discarding parts of the pretrained encoder setup: Kimi~K3 trains a vision encoder from scratch jointly with the multimodal backbone~\citep{kimiteam2026kimik3}, and encoder-free models directly process pixel patches, forcing visual encoding to be learned entirely within the backbone~\citep{adept2023fuyu,diao2024eve,liu2026tuna2,gemmateam2026gemma4,thinkingmachines2026inkling}.
Each recipe implies a different answer to what the vision encoder contributes, if anything at all.
We ask which of its inherited design choices benefit multimodal models trained entirely from scratch.

The predominant multimodal vision encoding recipes bundle together three design choices:
  (i)~a \emph{visual prior} from upstream pretraining,
  (ii)~\emph{dedicated vision capacity} (\ie, parameters that only vision tokens use),
  and (iii)~\emph{pre-fusion placement} of that capacity in a separate encoder that operates before vision and text tokens interact in the backbone---rather than operating \emph{in-fusion}, within the shared backbone.
A pretrained encoder keeps all three, a from-scratch encoder gives up only the prior, and an encoder-free model gives up all three.

Existing comparisons cannot isolate these choices, for two reasons.
First, they entangle the choices with each other and the rest of the recipe.
Published models differ in data, training stages, compute, and architecture.
While controlled studies hold multimodal training data and compute fixed, their comparisons among pretrained encoders~\citep{mckinzie2024mm1,tong2024cambrian,fan2025webssl,li2025openvision} vary \emph{which visual prior} an encoder imports rather than \emph{whether it needs one} in the first place.
Studies of encoder-based \vs encoder-free models either change all three choices~\citep{tong2026beyondlm,liu2026tuna2} or change capacity and placement together~\citep{shukor2025scaling}.
Second, comparisons with pretrained encoders control only multimodal training, treating encoder pretraining as free even though it adds substantial compute and data of its own (\eg, SigLIP~2 trains on 40B image--text samples;~\citealp{tschannen2025siglip2}), and several studies attribute the encoder's advantage to its pretraining~\citep{lei2025sail,diao2025evev2}.

In this work, we conduct a controlled study that unbundles the design choices of multimodal vision encoding one at a time (\cref{fig:intro:encoder-overview}):
    (i) \textit{does the vision encoder need separate pretraining?}
    (ii) \textit{does vision encoding need dedicated capacity?}
    and (iii) \textit{does that capacity need to precede multimodal fusion?}
We train every model entirely from scratch, language backbone included, on a common pool of data with one encoder architecture, one dense backbone architecture, and one optimization recipe.
We evaluate models on downstream visual question-answering (VQA) and language benchmarks at backbone sizes from 0.6B to 6.4B.
Because recipes spend compute in different stages, we compare them at matched \emph{lifetime compute}: the total compute of all training stages, including any pretraining of the encoder or language model.

Our experiments answer these questions and test whether the answers hold beyond lifetime compute:
\begin{enumerate}[noitemsep,topsep=0pt,parsep=0pt,partopsep=0pt,leftmargin=1.5em]
  \item \textbf{Separate pretraining is optional} (\S\ref{sec:encoders}).
    At matched lifetime compute, a from-scratch encoder is on par with encoders pretrained on the same data.
    Even if we \emph{ignore upstream compute}  (\S\ref{sec:practical:free}), more encoder pretraining improves recognition but not VQA.
  \item \textbf{Dedicated vision capacity is key} (\S\ref{sec:encoder-free}).
    An encoder-free model trails an encoder-based model on visual understanding, even when its backbone is enlarged by the encoder's parameter count.
    The encoder's advantage thus lies in capacity dedicated to vision rather than in total parameter count.
  \item \textbf{Vision capacity can move in-fusion} (\S\ref{sec:capacity}).
    Adding vision-only capacity to the backbone's earliest layers is comparable to adding a from-scratch pre-fusion encoder.
    Scaling laws show in-fusion tracking the encoder, whereas encoder-free trails it at every size (\S\ref{sec:capacity:sizes}).
  \item \textbf{The findings hold beyond lifetime compute} (\S\ref{sec:practical}).
    Starting from a pretrained 1.7B language model, in-fusion stays comparable to the from-scratch encoder while encoder-free falls further behind.
    In-fusion can also lower time to first token, since no separate encoder must run before the backbone.
\end{enumerate}

Together, these findings show that multimodal models benefit from dedicated vision capacity, which however needs neither separate pretraining nor pre-fusion placement.

\FloatBarrier

\section{A controlled protocol for comparing multimodal recipes}
\label{sec:protocol}

We propose a protocol that varies the three design choices a vision encoder bundles, while holding fixed the data sources, optimization settings, compute, and backbone architecture. This isolates their effects when comparing current multimodal model recipes.
Conclusions about these design choices cannot be drawn from comparing published models, because they differ in several factors at once---\eg, they build on different pretrained language models and vision encoders that each vary in training data and compute.
We therefore use no external pretrained models: we train every component from scratch, with common data sources and training setups, and compare recipes at matched \textit{lifetime compute}, the total compute of all training stages.

\subsection{Recipes and the fusion boundary}
\label{sec:protocol:canonical}

\setlength{\columnsep}{12pt}%
\begin{wraptable}{r}{0.42\linewidth}
  \vspace{-1.096em}
  \centering
  \small
  \caption{
    \textbf{Unbundling the encoder's design choices.} \linebreak
    A pretrained encoder bundles all three; in-fusion keeps only dedicated vision capacity.
  }
  \label{tab:protocol:recipes}
  \vspace{-0.13em}
  
\providecommand{\dyes}{$\checkmark$}
\providecommand{\dno}{\textcolor{PaperGray500}{$\times$}}
\setlength{\tabcolsep}{2.4pt}
\renewcommand{\arraystretch}{1.05}
\begin{tabular}{@{}lccc@{}}
  \toprule
  Recipe & \makecell{Prior} & \makecell{Capacity} & \makecell{Pre-fusion} \\
  \midrule
  \recipeglyph{metaclip}Pretrained encoder & \dyes & \dyes & \dyes \\
  \recipeglyph{scratch}Scratch encoder & \dno & \dyes & \dyes \\
  \recipeglyph{encoderfree}Encoder-free & \dno & \dno & \dno \\
  \recipeglyph{infusionqkvo}In-fusion & \dno & \dyes & \dno \\
  \bottomrule
\end{tabular}

  \vspace{-0.56em}
\end{wraptable}

All recipes differ only in how they encode visual input.

\paragraph{Backbone.}
Our backbones use an architecture modeled after Qwen3~\citep{yang2025qwen3}: a dense, decoder-only, pre-norm Transformer~\citep{vaswani2017attention} with RMSNorm~\citep{zhang2019rmsnorm}, RoPE~\citep{su2024roformer}, SwiGLU feed-forward layers~\citep{shazeer2020glu}, and grouped-query attention~\citep{ainslie2023gqa} with query--key normalization~\citep{dehghani2023vit22b} (\cref{sec:appx:architecture}).
Our main comparisons (\cref{sec:encoders,sec:encoder-free,sec:capacity}) use the Qwen3 1.7B and 4B configurations with dedicated vision capacity fixed at 0.4B parameters, and our scaling study (\cref{sec:capacity:sizes}) uses a 0.6B--6.4B ladder that grows vision capacity in proportion to the backbone.
Training uses autoregressive cross-entropy on text tokens; vision tokens precede the text they condition and have no loss.
Each image's vision tokens attend to each other bidirectionally and text tokens attend causally to all earlier tokens, including vision tokens (\cref{sec:appx:data-mixture}).

\paragraph{Fusion boundary.}
We describe how image patches reach the backbone relative to the \emph{multimodal fusion boundary}, the first layer in which vision and text tokens can interact: vision-only computation before it is \emph{pre-fusion} and computation inside the shared backbone is \emph{in-fusion} (\cref{fig:intro:encoder-overview,tab:protocol:recipes}).
Scratch encoders~(\recipemark{scratch}) process image patches pre-fusion with a randomly initialized 27-layer, 0.4B-parameter Vision Transformer~\citep{dosovitskiy2021image,alabdulmohsin2023getting} modeled after MoonViT-V2 (Kimi~K3's encoder; \citealp{kimiteam2026kimik3}), whose output is linearly projected into the backbone.
Pretrained encoders~(\recipemark{metaclip}) share this architecture, but we first pretrain them ourselves on our common data (\cref{sec:protocol:data}) rather than taking external checkpoints.
Encoder-free models~(\recipemark{encoderfree}) map patches directly into the backbone with a linear patch embedding~\citep{adept2023fuyu,gemmateam2026gemma4}, followed by RMSNorm and a learned scalar gate~\citep{qi2025beyond,li2026unseen}.
In-fusion models~(\recipemark{infusionqkvo}) instead add vision-only modules inside the backbone (\cref{sec:capacity}).
Every recipe trains on $224\times224$ images, read as 256 vision tokens using $14\times14$ patches.
Instruction tuning for evaluation raises the resolution to $448\times448$ (1,024 vision tokens), as in PaliGemma's short high-resolution stage~\citep{beyer2024paligemma}, since OCR and chart reading needs fine detail~\citep{tong2024cambrian,beyer2024vitspeed}.

\setlength{\columnsep}{0.65cm}%
\subsection{Common data}
\label{sec:protocol:data}
Text-only sequences come from Nemotron-CC v2/v2.1~\citep{nvidia2025nemotronnano2,nvidia2025nemotron3nano}, and image--text sequences use MetaCLIP 1.2 pairs (English subset)~\citep{xu2023metaclip}.
Encoder pretraining uses these pairs with their original alt-text, as in each objective's original setup (\cref{sec:encoders}; \eg, CapPa,~\citealp{tschannen2023captioners}), while multimodal training uses captions regenerated by Qwen2.5-VL-32B~\citep{bai2025qwen25vl} and shared by every recipe.
Multimodal training mixes text-only and image--text sequences 70:30, so recipes with a pretrained encoder spend more lifetime compute on image--text data.
This matches the language share \citet{han2026physics} found optimal and gave the best visual understanding in our sweep at a small cost to text (\cref{sec:encoders:mixtures}).

\subsection{Lifetime compute}
\label{sec:protocol:accounting}
Multimodal models often build on separately pretrained components whose training can even exceed the multimodal stage.
For example, SigLIP~2, a commonly used encoder, is trained on 40B image--text samples~\citep{tschannen2025siglip2}, which translates to 10 trillion image tokens, while language models are typically also trained on trillions of text tokens~\citep{touvron2023llama,yang2025qwen3}.
Counting only multimodal training therefore favors recipes that spend more compute in these earlier stages.
To make such comparisons fair, we propose \emph{lifetime compute}, the total training compute of every stage that produces the final model,
\begin{equation}
  C_{\mathrm{lifetime}} = \sum_{s \in \mathcal{S}} C_s,
  \label{eq:lifetime-compute}
\end{equation}
where $\mathcal{S}$ is the set of training stages and $C_s$ is the training compute of stage $s$.
Here, $\mathcal{S}$ contains multimodal training, any text-only training before it and, for pretrained encoders, encoder pretraining, but not instruction tuning.
We report compute in FLOPs; because recipes differ in cost per token, matched compute means that they process different numbers of tokens.
\Cref{sec:appx:lifetime-flops} describes the compute accounting and converts each budget into token-equivalents, the number of tokens a text-only model of the same size would process with the same compute.
Lifetime compute models a project that allocates one budget across every training stage, which lets us separate the contribution of each design choice.
In practice, pretrained encoders are often available for free, so we also compare encoders with their pretraining compute not counted (\cref{sec:practical:free}).

\subsection{Training schedule}
\label{sec:protocol:length}
To compare recipes at many training lengths from one run~\citep{zhai2022scaling}, we use a warmup--stable--decay schedule~\citep{hu2024minicpm,wen2024wsd,hagele2024scaling}, where the learning rate stays constant after warmup and any checkpoint of that run can be decayed to 5\% of its peak over the final 10\% of training (\cref{sec:appx:optimization}).
Our main comparisons span compute budgets from $6.5\times10^{20}$ to $2.9\times10^{22}$~FLOPs, or 56B to 1.11T token-equivalents (\cref{tab:protocol:budgets}).
We index budgets by tokens per parameter (TPP): token-equivalents divided by the backbone's non-embedding parameters, \ie, all parameters except the tied token embedding (1.4B at 1.7B and 3.6B at 4B; backbones are named by their total parameters).
Budgets reach 788 TPP at 1.7B and 306 TPP at 4B.
All recipes share one peak learning rate per backbone size, scaled with the inverse square root of its total parameter count ($3\times10^{-4}$ at 1.7B and $2\times10^{-4}$ at 4B).
At 1.7B, the encoder's lead over encoder-free and parity with in-fusion hold from half to twice that learning rate (\cref{sec:appx:scaling:lr}).

\subsection{Evaluation}
\label{sec:protocol:evaluation}
We evaluate recipes on text and visual understanding benchmarks and report the following metrics.
\textbf{Avg.\ Text} averages four category means over 12 language benchmarks: commonsense reasoning (COPA, HellaSwag, PIQA, Social IQa, and WinoGrande~\citep{roemmele2011copa,zellers2019hellaswag,bisk2020piqa,sap2019socialiqa,sakaguchi2020winogrande}), science question answering (ARC-Easy, ARC-Challenge, and OpenBookQA~\citep{clark2018arc,mihaylov2018openbookqa}), broad knowledge (MMLU~\citep{hendrycks2021mmlu}), and reading comprehension (RACE-middle, RACE-high, and BoolQ~\citep{lai2017race,clark2019boolq}).
\textbf{Avg.\ VQA} averages the general, knowledge, OCR and chart, and vision-centric category means of the 16 Cambrian visual question-answering benchmarks~\citep{tong2024cambrian}, after instruction tuning each checkpoint for three epochs on Cambrian-7M with $448\times448$ images.
A text-only model at the same lifetime compute serves as a reference for Avg.\ Text (\cref{sec:appx:scaling:text}).
\Cref{sec:encoder-free,sec:capacity:sizes} also compare held-out caption loss on images never seen in training (\cref{sec:appx:perplexity}).
\Cref{sec:appx:evaluation} details the benchmarks, and \cref{sec:appx:per-benchmark} gives every benchmark score.

\section{Does the vision encoder need separate pretraining?}
\label{sec:encoders}

The first design choice a vision encoder bundles is a visual prior imported through separate pretraining, which recent models~\citep{kimiteam2026kimik3} forgo by training the encoder from scratch with the language model.
Under a fixed lifetime compute budget, encoder pretraining directly competes with multimodal training for FLOPs.
Whether the prior is needed also shapes the remainder of our study, because encoder-free architectures cannot easily import one.
We test whether the prior helps by comparing encoders pretrained with different objectives against the same encoder trained from scratch with the backbone.

\paragraph{Setup.}
We pretrain a separate encoder with each of four objectives and the same compute: contrastive image--text alignment (MetaCLIP;~\citealp{xu2023metaclip}), self-supervised learning (WebSSL;~\citealp{fan2025webssl}), and captioning (Cap and CapPa;~\citealp{tschannen2023captioners}).
Pretrained encoders are updated during multimodal training and compared with an encoder trained from scratch at matched lifetime compute; we also test a frozen MetaCLIP encoder.
MetaCLIP, an instance of widely used CLIP-style pretraining~\citep{radford2021clip,zhai2023sigmoid,tschannen2025siglip2}, serves as the pretrained reference in the remaining experiments.
\Cref{sec:appx:encoder-objectives} details the pretraining objectives and \cref{sec:appx:architecture} the encoder architecture.

\begin{figure}[t]
  \centering
  \setlength{\parskip}{0pt}%
  \captionsetup[sub]{skip=0pt,justification=centering}%
  \includegraphics{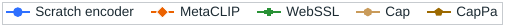}\par
  \nointerlineskip\vspace{3pt}%
  \includegraphics[width=\linewidth]{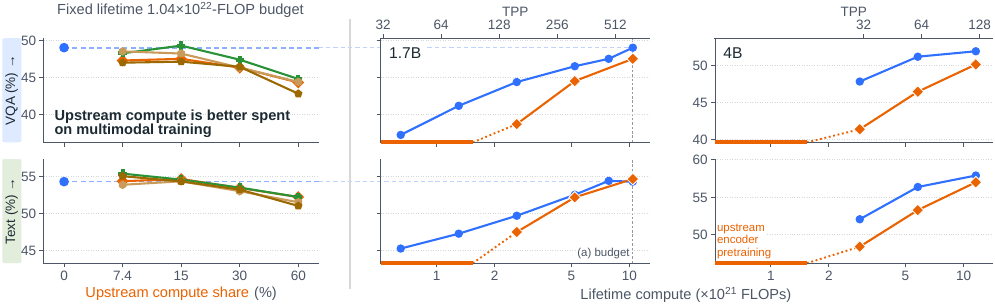}\par
  \nointerlineskip\vspace{2pt}%
  \hspace*{0.0047\linewidth}%
  \begin{minipage}[t]{0.3158\linewidth}
    \subcaption{1.7B backbone}
    \label{fig:encoders:upstream:a}
  \end{minipage}%
  \hspace*{0.0572\linewidth}%
  \begin{minipage}[t]{0.6200\linewidth}
    \subcaption{From-scratch versus MetaCLIP pretrained}
    \label{fig:encoders:upstream:b}
  \end{minipage}%
  \hspace*{0.0023\linewidth}\par
  \vspace{-0.5em}
  \caption{
    \textbf{Encoder pretraining is not beneficial at matched lifetime compute.}
    \panel{a} At 1.7B and $1.04 \times10^{22}$-FLOP lifetime budget, spending a larger share of compute on encoder pretraining is comparable to the from-scratch encoder (horizontal \textcolor{PaperBlue700}{\textbf{dashed blue}}) up to 15\% and lowers performance beyond it.
    \panel{b} MetaCLIP pretraining uses $1.54\times10^{21}$~FLOPs at every backbone size (\textcolor{PaperOrange700}{\textbf{orange floor}}); the from-scratch encoder matches or exceeds MetaCLIP at every budget. The vertical \textcolor{PaperGray600}{\textbf{dashed gray}} line at 1.7B marks the budget of \panel{a} and the top axis gives the from-scratch encoder's TPP.
  }
  \label{fig:encoders:upstream}
  \label{fig:appx:pretrained-lifetime}%
\end{figure}

\paragraph{A from-scratch encoder is on par with pretrained ones.}
With a 1.7B backbone and a $1.04 \times10^{22}$ FLOP budget, all four pretrained encoders are comparable to the from-scratch encoder when pretraining takes 7.4\% or 15\% of the budget (e.g., Cap sees 1B or 2B image--text examples).
Larger upstream shares hurt multimodal performance: at a 60\% share, both averages fall clearly below the from-scratch encoder (\cref{fig:encoders:upstream:a}).
Across backbone sizes, the from-scratch encoder matches or exceeds MetaCLIP at every budget (\cref{fig:encoders:upstream:b}).
\Cref{sec:appx:prior} gives the per-objective results, including frozen encoders, the full pretraining sweep, the results at 4B, and compositional image--text matching on SugarCrepe~\citep{hsieh2023sugarcrepe}.

Encoder pretraining does not help even when its compute is free (\cref{sec:practical:free}).
More pretraining raises ImageNet accuracy but not Avg.\ VQA, and pretrained encoders are only comparable to the from-scratch encoder at matched multimodal compute.
Delaying vision also hurts, as training on text alone for the first 25--75\% of training, with the total number of image--text examples held fixed, clearly lowers Avg.\ VQA (\cref{tab:understanding:schedule}), consistent with the vision laziness that \citet{han2026physics} observe after text-only training.

\vspace{0.5em}

\begin{takeawaybox}
    \textbf{Takeaway 1: Encoder pretraining offers little advantage.}
    At matched lifetime compute, an encoder trained from scratch together with the backbone is on par with encoders pretrained on the same data.
\end{takeawaybox}

\suppressfloats[t]%
\begin{figure}[t]
  \centering
  \setlength{\parskip}{0pt}%
  \captionsetup[sub]{skip=0pt,justification=centering}%
  \includegraphics[width=\linewidth]{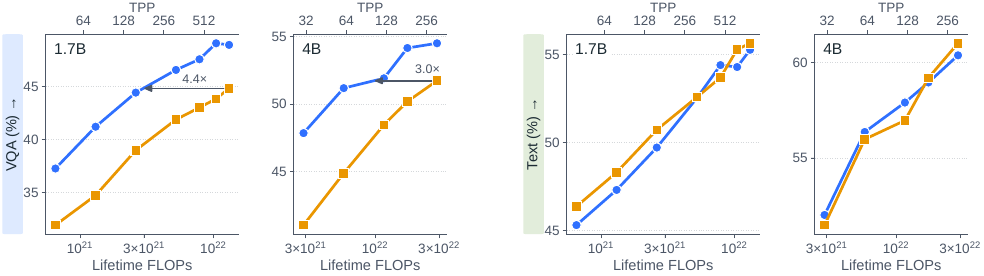}\par
  \nointerlineskip\vspace{-9pt}%
  \rlap{\makebox[\linewidth][c]{\smash{\raisebox{-23.8pt}{\includegraphics{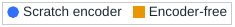}}}}}%
  \hspace*{0.0058\linewidth}%
  \begin{minipage}[t]{0.4621\linewidth}
    \subcaption{VQA}
    \label{fig:encoder-free:gap:vqa}
  \end{minipage}%
  \hspace*{0.0677\linewidth}%
  \begin{minipage}[t]{0.4621\linewidth}
    \subcaption{Text}
    \label{fig:encoder-free:gap:text}
  \end{minipage}%
  \hspace*{0.0023\linewidth}\par
  \caption{
    \textbf{Encoder-free trails the encoder on Avg.\ VQA, while Avg.\ Text stays comparable.}
    Each checkpoint is decayed, and arrows mark the compute factor encoder-free needs to match the encoder.
    \Cref{sec:appx:encoder-free-categories} gives per-category results.
  }
  \label{fig:encoder-free:gap}
  \label{fig:scaling:trajectory-encoder-free}
\end{figure}
\section{Does vision encoding need dedicated capacity?}
\label{sec:encoder-free}

The from-scratch encoder of \cref{sec:encoders} forgoes the imported prior but keeps the second design choice: dedicated vision capacity.
\emph{Encoder-free} models drop this capacity entirely and project image patches directly into the shared backbone~\citep{adept2023fuyu,diao2024eve,liu2026tuna2,gemmateam2026gemma4,thinkingmachines2026inkling}.
Dropping the encoder, however, also removes a substantial number of parameters and changes how raw patches enter the backbone, so a gap between the two recipes could reflect fewer total parameters, a weak patch embedding, or missing dedicated capacity.
In this section, we test whether visual understanding needs this dedicated capacity, leaving the study of its precise placement to \cref{sec:capacity}.

\paragraph{Setup.}
\enlargethispage{\baselineskip}%
We compare the from-scratch encoder (\cref{sec:encoders}) with the encoder-free recipe (\cref{sec:protocol:canonical}), which maps each patch into the backbone with a linear patch embedding followed by RMSNorm and a learned gate.
As in every recipe, vision tokens attend to one another bidirectionally in the backbone (\cref{sec:protocol:canonical}).
Both recipes share common backbones, data, and lifetime compute budgets, and we compare them at 1.7B and 4B.

\paragraph{Encoder-free trails encoder on visual understanding.}
\Cref{fig:encoder-free:gap} shows that Avg.\ VQA clearly suffers from a missing encoder, across scales and budgets, while Avg.\ Text stays comparable.
This also holds when scaling the encoder's size (\cref{sec:capacity:sizes}).
This result differs from \citet{shukor2025scaling}, who find that encoder-free models match or slightly outperform encoder-based ones, mainly at smaller sizes.
We identify two choices in their setup that work against the encoder (\cref{sec:appx:shukor}).
First, their encoder's learning rate may have been too small.
Second, they fit their scaling laws to an average of text, interleaved, and caption losses, which drowns out the signal from vision.
In-distribution held-out caption loss instead shows the encoder-free deficit and tracks Avg.\ VQA across recipes, sizes, and budgets (\cref{fig:scaling:loss-vqa}), whereas their averaged loss hides it.

\paragraph{Matching total parameters does not overcome encoder-free's deficit.}
We next widen or deepen the encoder-free backbone by 0.4B parameters to match the vision parameters added by the encoder (\cref{tab:encoder-free:steelman}; 4B in \cref{tab:appx:encoder-free:4b}).
None of the variants comes close to closing the gap on Avg.\ VQA, though all improve Avg.\ Text, since text tokens also use the added parameters.
The remaining gap in visual understanding is largest on OCR and chart benchmarks.
The same parameters are thus more useful as a vision encoder.
As none of these variants improves Avg.\ VQA, encoder-free hereafter keeps the plain backbone shared across all recipes.

\begin{table}[h!]
  \centering
  \footnotesize
  \caption{
    \textbf{Matching total parameters does not close the encoder-free gap.}
    All rows use a 1.7B backbone and $2.6\times10^{21}$ lifetime FLOPs.
    Indented rows ($+$) add the encoder's 0.4B parameters to the backbone.
    Bold: within one point of the best.
  }
  \label{tab:encoder-free:steelman}
  
\setlength{\tabcolsep}{3.0pt}
\renewcommand{\arraystretch}{1.05}
\begin{tabular}{@{}lrrrrrrrrrrrr@{}}
  \toprule
  \multicolumn{1}{>{\columncolor{PaperGray100!55}[.5\tabcolsep][.5\tabcolsep]}c}{\textbf{Architecture}}
    & \multicolumn{2}{>{\columncolor{PaperGray100!55}[.5\tabcolsep][.5\tabcolsep]}c}{\textbf{Added params (B)}}
    & \multicolumn{5}{>{\columncolor{PaperBlue100!55}[.5\tabcolsep][.5\tabcolsep]}c}{\textbf{VQA (\%) $\uparrow$}}
    & \multicolumn{5}{>{\columncolor{PaperGreen100!55}[.5\tabcolsep][.5\tabcolsep]}c}{\textbf{Text (\%) $\uparrow$}} \\
  \cmidrule(l{.5\tabcolsep}r{.5\tabcolsep}){1-1}\cmidrule(l{.5\tabcolsep}r{.5\tabcolsep}){2-3}\cmidrule(l{.5\tabcolsep}r{.5\tabcolsep}){4-8}\cmidrule(l{.5\tabcolsep}r{.5\tabcolsep}){9-13}
  Variant & Vision-only & Shared & \multicolumn{1}{c}{Avg} & \multicolumn{1}{c}{Gen} & \multicolumn{1}{c}{Knw} & \multicolumn{1}{c}{OCR} & \multicolumn{1}{c}{Vis} & \multicolumn{1}{c}{Avg} & \multicolumn{1}{c}{Com} & \multicolumn{1}{c}{Sci} & \multicolumn{1}{c}{Knw} & \multicolumn{1}{c}{Rdg} \\
  \midrule
  \recipeglyph{scratch}Scratch encoder & 0.4 & -- & \textbf{44.4} & \textbf{57.2} & \textbf{35.7} & \textbf{38.2} & \textbf{46.5} & 49.7 & \textbf{63.9} & 45.0 & 36.7 & 53.4 \\
  \recipeglyph{encoderfree}Encoder-free & -- & -- & 39.0 & 50.7 & \textbf{35.0} & 25.2 & 45.0 & \textbf{50.7} & \textbf{64.1} & \textbf{45.8} & 38.2 & 54.8 \\
  \recipeglyph{}\hspace{0.7em}+ wider & -- & 0.4 & 38.6 & 48.5 & 34.2 & 29.3 & 42.2 & \textbf{51.3} & \textbf{64.0} & \textbf{46.5} & 38.0 & \textbf{56.6} \\
  \recipeglyph{}\hspace{0.7em}+ deeper & -- & 0.4 & 38.7 & 50.0 & 33.2 & 26.3 & 45.4 & \textbf{51.2} & \textbf{64.2} & \textbf{45.8} & 39.1 & \textbf{55.9} \\
  \recipeglyph{}\hspace{0.7em}+ wider and deeper & -- & 0.4 & 38.8 & 51.6 & \textbf{34.8} & 25.1 & 43.7 & \textbf{51.4} & \textbf{64.0} & \textbf{45.7} & \textbf{40.3} & \textbf{55.9} \\
  \arrayrulecolor{PaperGray400}\cmidrule{1-13}\arrayrulecolor{black}
  \multicolumn{13}{@{}l}{\textit{Richer patch embedding}} \\
  \recipeglyph{}Two-layer projection & $<$0.01 & -- & 40.5 & 50.8 & 33.9 & 30.5 & \textbf{46.7} & 49.7 & \textbf{63.2} & \textbf{45.6} & 36.3 & 53.5 \\
  \recipeglyph{}Hierarchical MLP & 0.04 & -- & 35.9 & 50.8 & 33.1 & 15.3 & 44.5 & 50.3 & \textbf{63.5} & 45.1 & \textbf{39.3} & 53.4 \\
  \bottomrule
\end{tabular}

\end{table}

\paragraph{A richer patch embedding does not close the gap.}
We also study replacing the encoder-free linear patch embedding with two richer ones: a normalized two-layer projection inspired by Gemma~4~\citep{gemmateam2026gemma4}, and a four-layer hierarchical MLP~\citep{touvron2022three}, as in Inkling~\citep{thinkingmachines2026inkling}.
Trained on $224\times224$ images, neither comes close to the encoder (\cref{tab:encoder-free:steelman}), though the hierarchical MLP matches it when compressing $640\times640$ images into the same 256 tokens (\cref{sec:appx:encoder-free-tokenization}).
Subsequent encoder-free experiments keep the default patch embedding, so dedicated capacity is placed deliberately.

\vspace{0.5em}

\begin{takeawaybox}
  \textbf{Takeaway 2: Visual understanding benefits from dedicated vision capacity.}
  Encoder-free models trail the from-scratch encoder on Avg.\ VQA.
  Adding backbone capacity with the same encoder parameter budget does not close the gap.
\end{takeawaybox}

\clearpage

\section{Does dedicated capacity need to precede multimodal fusion?}
\label{sec:capacity}

\Cref{sec:encoder-free} shows that visual understanding benefits from dedicated capacity.
We now ask whether that capacity must precede multimodal fusion.
The encoder's parameters are both \emph{dedicated}, used only by vision tokens, and \emph{pre-fusion}, processing vision tokens before they interact with text.
We test whether pre-fusion placement is needed by moving the encoder's parameter budget in-fusion, into vision-only modules inside the shared backbone (\cref{sec:capacity:infusion}).
\Cref{sec:capacity:design} studies where this capacity should sit, which tokens should use it, and which modules it adds.
\Cref{sec:capacity:fusion,sec:capacity:sizes} then compare it with the encoder over training and with scale.

\begin{figure}[t]
  \centering
  \setlength{\parskip}{0pt}%
  \captionsetup[sub]{skip=2pt,justification=centering}%
  \includegraphics[width=\linewidth]{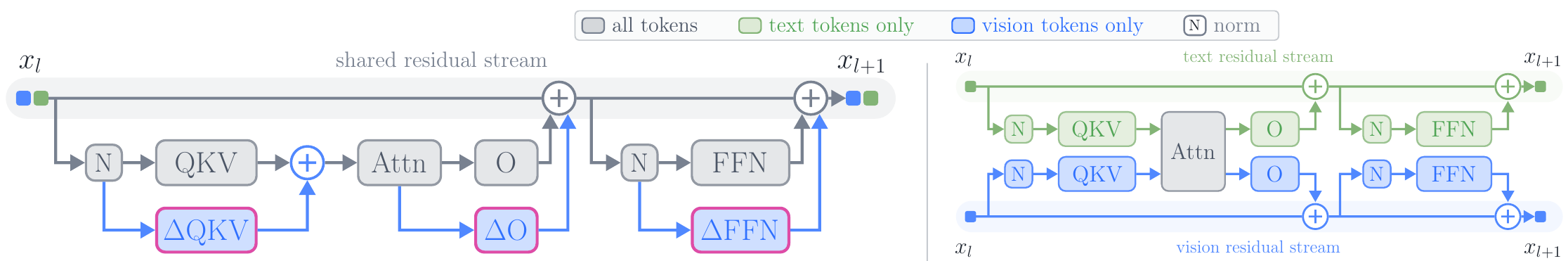}\par
  \nointerlineskip\vspace{2pt}%
  \hspace*{0.0028\linewidth}%
  \begin{minipage}[t]{0.5696\linewidth}
    \subcaption{\textcolor{PaperPink700}{\textbf{In-fusion}}:~vision modules add to a shared residual stream}
    \label{fig:capacity:block:infusion}
  \end{minipage}%
  \hspace*{0.0373\linewidth}%
  \begin{minipage}[t]{0.3875\linewidth}
    \subcaption{\textcolor{PaperGray700}{\textbf{MoT}: separate streams, joint attention}}
    \label{fig:capacity:block:mot}
  \end{minipage}%
  \hspace*{0.0028\linewidth}\par
  \caption{
      \textbf{In-fusion vision capacity adds to the shared backbone instead of replacing it.}
      \panel{a} All text~\tokenglyph{text} and vision tokens~\tokenglyph{vision} are processed by shared modules (\textcolor{PaperGray600}{\textbf{gray}});
        vision tokens \textit{also} pass through vision-only modules (\textcolor{PaperBlue700}{\textbf{blue}} with \textcolor{PaperPink700}{\textbf{pink}} outline) that add to the shared outputs;
        \oursffn omits the vision attention projections ($\Delta$QKV, $\Delta$O).
      \panel{b} MoT uses entirely separate modules for text and vision, so the modality streams interact only through attention.
    }
    \label{fig:capacity:block}
\end{figure}
\subsection{Moving the encoder's capacity inside the backbone}
\label{sec:capacity:infusion}

\oursstyled adds vision-only modules to selected backbone layers.
In each of these layers, the attention projections and the feed-forward network (FFN) keep their shared weights and add vision-only counterparts, whose outputs add to the shared outputs after each module (\cref{fig:capacity:block}).
Each vision-only module thus acts as an additive expert routed by modality: vision tokens use it alongside the shared weights.
Unlike mixture-of-transformers~\citep[MoT;][]{liang2025mixture}, which gives text and vision tokens entirely separate parameters, in-fusion still passes every token through the shared parameters.
\Cref{sec:capacity:late-onset} compares the two when starting from pretrained language models.

For token $i$ with vision indicator $m_i\in\{0,1\}$, each attention projection $W_P$ gains a vision-only counterpart $\Delta W_P$ that applies only to vision tokens,
\begin{equation}
  P(\mathbf{z}_i) = W_P\,\mathbf{z}_i + m_i\,\Delta W_P\,\mathbf{z}_i, \qquad P\in\{Q,K,V,O\},
  \label{eq:capacity:infusion-attn}
\end{equation}
where $\mathbf{z}_i$ is the projection's input (the normalized hidden state for $Q$, $K$, $V$ and the attention output for $O$).
Attention thus remains joint across vision and text tokens.
The FFN gains a vision-only branch,
\begin{equation}
  \mathbf{h}_i' = \mathbf{h}_i + F\!\left(N(\mathbf{h}_i)\right) + m_i\,\Delta F\!\left(N(\mathbf{h}_i)\right),
  \label{eq:capacity:infusion-ffn}
\end{equation}
where $F$ is the shared FFN, $N$ is its normalization, and $\Delta F$ is a vision-only FFN of the same form that reads the same normalized input (\cref{sec:vis-capacity:parameterization}).
The vision-only projections $\Delta W_P$ and the output projection of $\Delta F$ are zero-initialized, so training starts from the shared model.
Because these modules act only when $m_i=1$, text-only inputs are processed exactly as they would have been by the shared backbone.

We use two variants: ``\oursqkvo'', our default, adds vision FFN branches of smaller hidden width and vision-only counterparts of all four attention projections, while ``\oursffn'' adds only vision FFN branches.
We size both variants to match the encoder's vision-only parameter budget (\cref{tab:capacity:qkvo}), so comparisons with the encoder vary only where dedicated capacity is placed.
Existing vision-specific architectures do not fit this comparison: they either duplicate vision weights that scale with the backbone~\citep{wang2024cogvlm,liang2025mixture} or use vision adapters that rely on an encoder~\citep{dong2024internlmxcomposer2,wang2025vora}.

\clearpage

\subsection{Where and how to add in-fusion capacity}
\label{sec:capacity:design}

\begin{table}[t]
  \centering
  \small
  \caption{
    \textbf{Added in-fusion capacity is most effective when placed early and reserved for vision tokens.}
    Every variant adds 0.4B parameters to the 1.7B backbone.
    Schematics show backbone layers from early to late; gray serves all tokens, blue only vision tokens, and wider layers are taller.
    \hlrow{Highlighted}: our default placement.
    Bold: within one point of the best.
  }
  \label{tab:capacity:design}
  \resizebox{\linewidth}{!}{
\setlength{\tabcolsep}{3.0pt}
\renewcommand{\arraystretch}{1.05}
\begin{tabular}{@{}llllrrrrrrrrrr@{}}
  \toprule
  \multicolumn{4}{>{\columncolor{PaperGray100!55}[.5\tabcolsep][.5\tabcolsep]}c}{\textbf{Architecture}}
    & \multicolumn{5}{>{\columncolor{PaperBlue100!55}[.5\tabcolsep][.5\tabcolsep]}c}{\textbf{VQA (\%) $\uparrow$}}
    & \multicolumn{5}{>{\columncolor{PaperGreen100!55}[.5\tabcolsep][.5\tabcolsep]}c}{\textbf{Text (\%) $\uparrow$}} \\
  \cmidrule(l{.5\tabcolsep}r{.5\tabcolsep}){1-4}\cmidrule(l{.5\tabcolsep}r{.5\tabcolsep}){5-9}\cmidrule(l{.5\tabcolsep}r{.5\tabcolsep}){10-14}
  Variant & Design & Used by & Placement & \multicolumn{1}{c}{Avg} & \multicolumn{1}{c}{Gen} & \multicolumn{1}{c}{Knw} & \multicolumn{1}{c}{OCR} & \multicolumn{1}{c}{Vis} & \multicolumn{1}{c}{Avg} & \multicolumn{1}{c}{Com} & \multicolumn{1}{c}{Sci} & \multicolumn{1}{c}{Knw} & \multicolumn{1}{c@{}}{Rdg} \\
  \midrule
  \rowcolor{PaperGray100!50}
  \recipeglyph{scratch}Scratch encoder & \hspace{-3.42ex}\raisebox{-0.5ex}{\includegraphics[height=1.9ex]{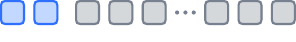}} & Vision & Pre-fusion & \textbf{44.4} & \textbf{57.2} & 35.7 & \textbf{38.2} & 46.5 & 49.7 & \textbf{63.9} & 45.0 & 36.7 & 53.4 \\
  \rowcolor{PaperGray100!50}
  \recipeglyph{encoderfree}Encoder-free & \raisebox{-0.5ex}{\includegraphics[height=1.9ex]{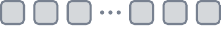}} & -- & -- & 39.0 & 50.7 & 35.0 & 25.2 & 45.0 & \textbf{50.7} & \textbf{64.1} & \textbf{45.8} & 38.2 & \textbf{54.8} \\
  \addlinespace[2pt]
  \multicolumn{14}{@{}l}{\textit{Where to add (vision tokens)}} \\
  \recipeglyph{}\oursffnstyled{} & \raisebox{-0.5ex}{\includegraphics[height=1.9ex]{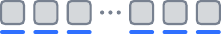}} & Vision & Every layer (narrow) & 39.4 & 46.4 & 33.4 & 31.3 & 46.6 & \textbf{50.7} & \textbf{64.5} & \textbf{45.6} & 39.8 & 53.1 \\
  \recipeglyph{}\oursffnstyled{} & \raisebox{-0.5ex}{\includegraphics[height=1.9ex]{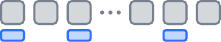}} & Vision & Spread (8 of 28) & 37.6 & 46.7 & 33.1 & 28.1 & 42.4 & 50.0 & \textbf{63.9} & \textbf{45.6} & 40.0 & 50.5 \\
  \recipeglyph{}\oursffnstyled{} & \raisebox{-0.5ex}{\includegraphics[height=1.9ex]{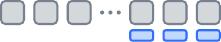}} & Vision & Layers 21--28 & 35.1 & 42.6 & 31.4 & 23.3 & 43.3 & 49.8 & \textbf{64.1} & 44.6 & 38.5 & 52.1 \\
  \rowcolor{PaperPink100!40}
  \recipeglyph{infusionffn}\oursffnstyled{} & \raisebox{-0.5ex}{\includegraphics[height=1.9ex]{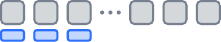}} & Vision & Layers 1--8 & \textbf{45.0} & \textbf{57.0} & \textbf{36.7} & 36.0 & \textbf{50.4} & \textbf{50.4} & 63.0 & \textbf{45.3} & \textbf{40.6} & 52.9 \\
  \arrayrulecolor{PaperGray400}\cmidrule{1-14}\arrayrulecolor{black}
  \multicolumn{14}{@{}l}{\textit{All-token controls (layers 1--8)}} \\
  \recipeglyph{}Shared FFN widening & \raisebox{-0.5ex}{\includegraphics[height=1.9ex]{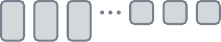}} & All & Layers 1--8 & 40.1 & 51.5 & 34.1 & 29.6 & 45.1 & \textbf{51.2} & \textbf{64.5} & \textbf{45.6} & \textbf{40.6} & \textbf{54.2} \\
  \recipeglyph{}Separate FFN branch & \raisebox{-0.5ex}{\includegraphics[height=1.9ex]{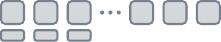}} & All & Layers 1--8 & 42.3 & 54.1 & \textbf{37.4} & 30.8 & 46.7 & \textbf{51.2} & \textbf{64.8} & \textbf{46.3} & \textbf{41.6} & 52.0 \\
  \bottomrule
\end{tabular}

}
\end{table}

We compare in-fusion designs that reallocate the encoder's 0.4B vision parameters within the backbone.

\paragraph{Placement of capacity within fusion.}
We study placement with the FFN-only variant, because FFN branches can be narrowed to fit any placement.
Full-rank vision-only attention projections cannot, because adding them to every layer would take 88\% of the 0.4B budget at 1.7B and exceed it at 4B.
With the budget fixed, we compare four placements (\cref{tab:capacity:design}): a narrow branch in every layer, or eight wider branches spread evenly across depth, in the last eight layers, or in the first eight layers.
The first three variants do no better than the encoder-free model on Avg.\ VQA, whereas placing the modules in the first eight layers matches the scratch encoder.
Early placement also helps most on OCR and chart questions, which depend on fine visual detail, whereas late placement does not.
Dedicated in-fusion capacity is therefore most useful where the backbone first processes vision tokens, similar to an encoder.
We hypothesize that vision tokens benefit most from vision-only processing while they are still close to raw patch features.
Only early placement concentrates the budget there, whereas the every-layer and spread placements dilute it across the backbone.
Layerwise representation similarity shows that the in-fusion models change vision tokens mostly within the first eight blocks, where their vision-only modules sit, and little afterward, while the encoder-free model keeps changing them through shared blocks 4--13 (\cref{sec:appx:layer-probes}).
Therefore, we use the first eight layers from here on.

\paragraph{Reserving the capacity for vision tokens matters.}
To check that the gain comes from reserving this capacity for vision tokens, we also give the same parameters in the first eight layers to all tokens (\cref{tab:capacity:design}).
Widening the shared FFNs barely improves over the encoder-free model.
A separate branch used by all tokens recovers some of the vision-only branch's gain, but still trails it even with twice the training (\cref{tab:appx:capacity:routing-longer}).
The all-token variants slightly improve on Avg.\ Text, because text also uses the added parameters.
We hypothesize that a vision-only branch gives vision tokens processing of their own without competing with text for the same parameters, while both still share the backbone's attention and FFNs.

\begin{wrapfigure}[12]{r}{0.35\linewidth}
  \vspace{-1em}
  \centering
  \includegraphics[width=\linewidth]{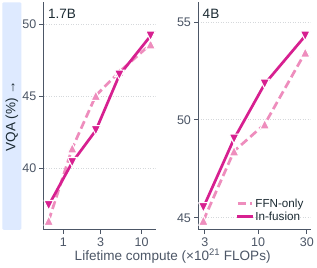}
  \vspace{-1.8em}
  \caption{\textbf{\oursqkvostyled is comparable to FFN-only at 1.7B and slightly ahead at every 4B budget.}}
  \label{fig:capacity:qkvo-scale}
  \vspace{-1.2em}
\end{wrapfigure}
\paragraph{Adding vision-only attention projections.}
With the capacity confined to the first eight layers, the QKVO attention projections fit within the budget (0.1B at 1.7B), so ``\oursqkvo'' adds them alongside narrower FFN branches.
It is comparable to FFN-only at 1.7B and slightly ahead at every 4B budget (\cref{fig:capacity:qkvo-scale}), and its vision-only attention projections help on their own at 4B (\cref{sec:vis-capacity:qkvo-controls}), so we use this variant as our default for the remainder of the paper.

\subsection{Can in-fusion capacity match a pre-fusion encoder?}
\label{sec:capacity:fusion}

\begin{figure}[t]
  \centering
  \setlength{\parskip}{0pt}%
  \captionsetup[sub]{skip=0pt,justification=centering}%
  \includegraphics[width=\linewidth]{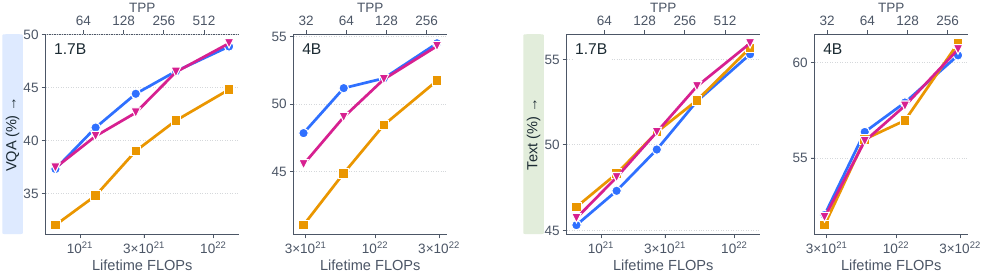}\par
  \nointerlineskip\vspace{-9pt}%
  \rlap{\makebox[\linewidth][c]{\smash{\raisebox{-23.8pt}{\includegraphics{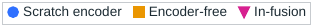}}}}}%
  \hspace*{0.0058\linewidth}%
  \begin{minipage}[t]{0.4621\linewidth}
    \subcaption{VQA}
    \label{fig:scaling:trajectory-infusion:vqa}
  \end{minipage}%
  \hspace*{0.0677\linewidth}%
  \begin{minipage}[t]{0.4621\linewidth}
    \subcaption{Text}
    \label{fig:scaling:trajectory-infusion:text}
  \end{minipage}%
  \hspace*{0.0023\linewidth}\par
  \caption{
    \textbf{In-fusion is comparable to the pre-fusion encoder.}
    Encoder-free is shown for reference.
    \Cref{tab:capacity:endpoints} gives per-category results at the longest budget.
  }
  \label{fig:scaling:trajectory-infusion}
\end{figure}

With the same dedicated budget on either side of the multimodal fusion boundary, we compare in-fusion with a scratch encoder at matched lifetime compute over training at 1.7B and 4B (\cref{fig:scaling:trajectory-infusion}).
\oursqkvostyled is comparable to the encoder throughout.
Like the encoder, \oursqkvo clearly leads encoder-free at both sizes; \cref{sec:capacity:sizes} compares the recipes across backbone sizes.
Part of the \oursqkvo budget sits in vision-only attention projections, which help on their own in additional 4B ablations (\cref{sec:vis-capacity:qkvo-controls}).
Dedicated vision capacity therefore does not need to precede fusion to benefit visual understanding.
In-fusion can also lower time to first token for image requests, since no encoder runs before the backbone (\cref{sec:practical:serving}).

\subsection{Results hold with scale}
\label{sec:capacity:sizes}

\begin{figure}[h!]
  \centering
  \setlength{\parskip}{0pt}%
  \captionsetup[sub]{skip=0pt,justification=centering}%
  \includegraphics[width=\linewidth]{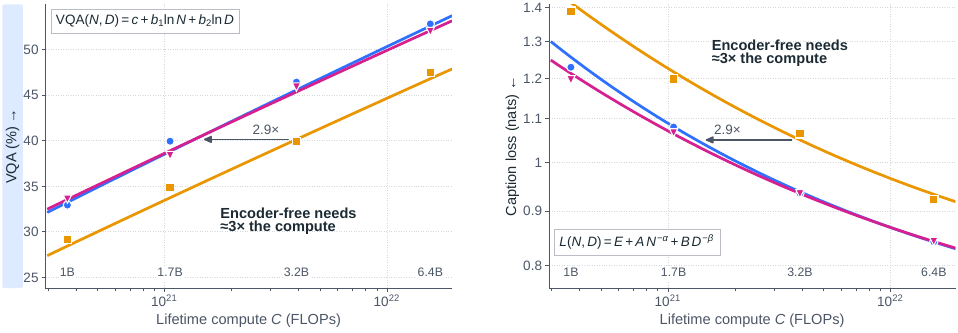}\par
  \nointerlineskip\vspace{-9pt}%
  \rlap{\makebox[\linewidth][c]{\smash{\raisebox{-24pt}{\includegraphics{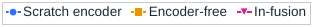}}}}}%
  \hspace*{0.0479\linewidth}%
  \begin{minipage}[t]{0.4240\linewidth}
    \subcaption{VQA}
    \label{fig:capacity:sizes:vqa}
  \end{minipage}%
  \hspace*{0.1017\linewidth}%
  \begin{minipage}[t]{0.4241\linewidth}
    \subcaption{Caption loss}
    \label{fig:capacity:sizes:loss}
  \end{minipage}%
  \hspace*{0.0023\linewidth}\par
  \caption{
    \textbf{Encoder-free trails the encoder at every size, while in-fusion tracks it.}
    Points and curves shown are 64 TPP.
    Each recipe's log-linear VQA fit \panel{a} and power-law loss fit \panel{b} use all of its runs, across sizes and TPPs (\cref{sec:appx:size-ladder}).
  }
  \label{fig:capacity:sizes}
\end{figure}

So far, we have compared the recipes at 1.7B and 4B with 0.4B of dedicated vision capacity.
To see how they behave with scale, we train them on a ladder of five backbone sizes from 0.6B to 6.4B (setup in \cref{sec:appx:size-ladder}).
The dedicated vision capacity grows with the backbone, keeping its share of parameters from the 1.7B model, as in jointly scaled models~\citep{chen2023pali,tian2025navil}.
For each recipe, we fit Avg.\ VQA, caption loss, and text loss as functions of model size $N$ and token-equivalents $D$ on all of its ladder runs.
At every size, encoder-free needs approximately three times the encoder's compute to match it on Avg.\ VQA and caption loss, while in-fusion tracks the encoder (\cref{fig:capacity:sizes}).
On text loss and Avg.\ Text, the three recipes are comparable (\cref{tab:appx:size-ladder-fits,tab:appx:size-ladder-scores}).
Taken together, the findings of \cref{sec:encoder-free,sec:capacity} hold with scale.

\vspace{0.5em}

\begin{takeawaybox}
  \textbf{Takeaway 3: Dedicated vision capacity can move into the backbone's earliest layers.}
  Placed in the earliest layers, vision-only modules with the encoder's budget are comparable to a from-scratch encoder at every backbone size.
\end{takeawaybox}

\vspace{0.5em}

\clearpage

\section{Vision encoding beyond lifetime compute}
\label{sec:practical}

Our main investigations (\cref{sec:encoders,sec:encoder-free,sec:capacity}) are deliberately controlled: they train every component from scratch and charge each stage's compute to one lifetime budget, which enables us to isolate the three design choices of a vision encoder.
In practice, however, multimodal models are often built from separately pretrained components: a language model and, usually, a vision encoder.
Inference speed for serving also shapes these design decisions.
We find that a ``free'' pretrained encoder still does not help a backbone trained from scratch (\cref{sec:practical:free}), and that instead midtraining a 1.7B text-only model leaves the comparisons of \cref{sec:encoders,sec:encoder-free,sec:capacity} unchanged (\cref{sec:capacity:late-onset}).
Finally, in-fusion can lower the time to first token for image requests (\cref{sec:practical:serving}).

\subsection{Zero-cost pretrained encoders}
\label{sec:practical:free}
\begin{figure}[t]
  \centering
  \begin{minipage}[t]{0.56\linewidth}
    \centering
    \includegraphics[width=\linewidth]{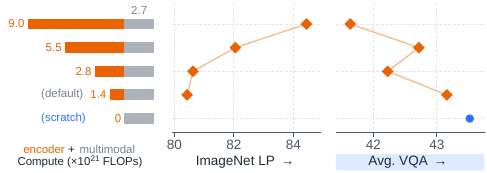}
      \caption{
        \textbf{Better encoder $\nRightarrow$ better VQA, even if encoder pretraining is free.}
        We compare MetaCLIP encoders pretrained with $1.4$--$9.0\times10^{21}$~FLOPs (\textcolor{PaperOrange700}{\textbf{orange}}) followed by identical $2.7\times10^{21}$-FLOP multimodal stages (\textcolor{PaperGray600}{\textbf{gray}}) with a 1.7B backbone.
        ImageNet linear-probe accuracy is measured after pretraining, and VQA after the multimodal stage.
        \textcolor{PaperBlue700}{\textbf{Blue}} shows a scratch encoder (no pretraining).
      }
      \label{fig:practical:free-pretraining}
  \end{minipage}\hfill
  \begin{minipage}[t]{0.40\linewidth}
    \centering
    \includegraphics[width=\linewidth]{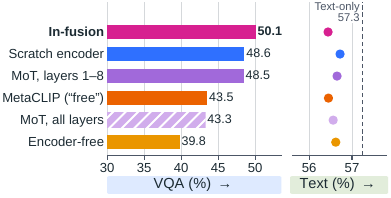}
    \caption{
      \textbf{In midtraining, in-fusion, early-layer MoT, and the from-scratch encoder are comparable on VQA.}
      All-layer MoT, pretrained MetaCLIP encoder, and encoder-free trail behind.
      Each recipe continues from a 1.7B text-only model at 1T token-equivalents to 1.33T.
    }
    \label{fig:capacity:late-onset}
  \end{minipage}
\end{figure}

\Cref{sec:encoders} charges encoder pretraining to the lifetime budget, but in practice a pretrained encoder is often available for free.
Here, we hold multimodal training fixed at $2.7\times10^{21}$ FLOPs and vary MetaCLIP pretraining from $1.4\times10^{21}$ to $9.0\times10^{21}$ FLOPs.
This raises the encoder's ImageNet linear-probe accuracy before multimodal training from 80.4\% to 84.4\% (\cref{fig:practical:free-pretraining}; \cref{sec:appx:upstream-fixed}), yet Avg.\ VQA is highest after the shortest pretraining, and an encoder trained from scratch with the same multimodal training matches it.
Thus, even when pretraining is free, more of it does not help.

\subsection{Midtraining from pretrained language models}
\label{sec:capacity:late-onset}

Many multimodal models add vision to a language model pretrained on text alone~\citep{alayrac2022flamingo,liu2023visual,tong2024cambrian,bai2025qwen3vl}.
We call this stage of continued training on multimodal data \emph{midtraining}.
We test whether the comparisons of \cref{sec:encoders,sec:encoder-free,sec:capacity} hold in this setting.
We pretrain our own 1.7B language model on 1T tokens of the same text data and then midtrain it to 1.33T token-equivalents with each recipe at equal compute (\cref{fig:capacity:late-onset}).
Following DeepSeek-VL~\citep{lu2024deepseekvl}, midtraining keeps 70\% text data; as in \cref{sec:practical:free}, MetaCLIP's encoder pretraining is not counted.
Besides the four recipes, we compare MoT (\cref{sec:capacity:infusion}, initialized from the language model) and a variant that copies only its first eight layers, matching the encoder's 0.4B vision-only parameters.

\oursqkvostyled, MoT layers 1--8, and the from-scratch encoder perform comparably, and all lead encoder-free.
The full MoT model falls behind despite 3.5 times as many vision-only parameters---adding vision-only copies of the later layers to MoT layers 1--8 hurts, consistent with late placement not helping in \cref{sec:capacity:design}.
A free MetaCLIP encoder, which matches the from-scratch encoder when the backbone is also trained from scratch (\cref{sec:practical:free}), trails it here, mostly on OCR, chart, and vision-centric questions.
\enlargethispage{\baselineskip}%
Tuning each encoder's learning rate does not change these conclusions (\cref{sec:appx:midtraining-lr}).
We hypothesize that the from-scratch encoder learns features suited to the pretrained language model from its first update, whereas MetaCLIP's contrastive features, which miss fine visual detail~\citep{tong2024eyes}, must be reshaped by a language model that already leans on text.
Every recipe keeps Avg.\ Text within a point of a text-only control.
Thus, the conclusions from our controlled experiments (\cref{sec:encoders,sec:encoder-free,sec:capacity}) carry over to midtraining, where a free pretrained encoder trails the from-scratch encoder, encoder-free trails on VQA, and in-fusion is comparable to the pre-fusion encoder.

\subsection{Training and inference speed}
\label{sec:practical:serving}
\begin{wrapfigure}{r}{0.42\linewidth}
  \vspace{-1.2em}
  \centering
  \includegraphics[width=\linewidth]{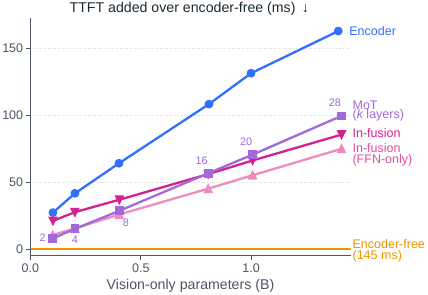}
  \vspace{-1.4em}
    \caption{
    \textbf{In-fusion serves images faster than the encoder at every vision parameter budget.}
    Designs are parameter-matched on the 1.7B backbone.
    }
    \label{fig:capacity:systems}
  \vspace{-0.3em}
\end{wrapfigure}
A separate encoder must finish processing each image before the backbone can start, which adds a serial step to inference.
Following the inference benchmark of Mono-InternVL~\citep{luo2025monointernvl}, we measure time to first token (TTFT) on one H200 with 16 concurrent requests, each with one 448-pixel image and 256 text tokens (\cref{sec:appx:inference-benchmark}).
At 1.7B with 0.4B vision-only parameters, \oursqkvo and its FFN-only variant are clearly faster than the encoder (\cref{fig:capacity:systems}) at comparable Avg.\ VQA
(\cref{tab:capacity:endpoints}).
MoT, which copies the first $k$ backbone layers for vision tokens, is about as fast as in-fusion up to 0.4B vision-only parameters, but its added TTFT grows faster with the budget, while staying below the encoder's TTFT.
Encoder-free is fastest but clearly trails on Avg.\ VQA (\cref{tab:capacity:endpoints}).
In-fusion's TTFT advantage over the encoder grows with the vision-only budget (\cref{fig:capacity:systems}), which scales with the backbone (\cref{sec:capacity:sizes}); at a fixed budget, vision adds nearly the same time at 1.7B and 4B (\cref{tab:appx:ttft-fixed-backbone}).
Earlier encoder-free models report larger TTFT reductions, up to 69\%~\citep{luo2025monointernvl,luo2025monointernvl15}, although their gains also reflect other efficiency changes.
Inference benefits also depend on deployment details.
Under continuous batching of mixed text and image requests, in-fusion's vision-only modules run in the same forward passes as text-only requests, whereas a separate encoder adds latency only to image requests.
Finally, both in-fusion variants train as fast as encoder-free, whereas the encoder recipe is 3\% slower at 4B (\cref{tab:appx:capacity:systems-4b}).

\begin{takeawaybox}
  \textbf{Takeaway 4: Our findings hold beyond lifetime compute, and in-fusion can be faster to serve.}
  A free pretrained encoder still does not help, the comparisons carry over to midtraining, and in-fusion lowers TTFT at comparable VQA.
\end{takeawaybox}

\section{Related work}
\label{sec:related-work}

\paragraph{Multimodal model recipes.}
Most multimodal models connect a separately pretrained vision encoder~\citep{radford2021clip,tschannen2025siglip2,bolya2025perception}, kept frozen~\citep{liu2023visual,li2023blip2,muse-glimmer} or updated~\citep{beyer2024paligemma,steiner2024paligemma2,bai2025qwen3vl}, to a pretrained language model.
More recent models train the encoder from scratch with the language model~\citep{kimiteam2026kimik3,tian2025navil}, embed image patches without an encoder~\citep{adept2023fuyu,diao2026neo,liu2026tuna2,gemmateam2026gemma4,thinkingmachines2026inkling}, or quantize images with a separately trained tokenizer~\citep{chameleonteam2024chameleon,wang2026emu3}.
Inkling was released as ``encoder-free'', yet its successor, Inkling~2~Small, uses a 292M-parameter vision encoder~\citep{thinkingmachines2026inkling,thinkingmachines2026inkling2small}.
Because these models also differ in data, pretrained components, and compute, comparing them directly does not isolate a single design choice.
We therefore train every stage of each recipe class from scratch on common data sources and compare the recipes at matched lifetime compute (\cref{sec:protocol}).

\paragraph{Controlled studies.}
Prior controlled studies of encoder pretraining typically start from a pretrained language model, count only multimodal training, and compare encoder pretraining objectives and data~\citep{karamcheti2024prismatic,mckinzie2024mm1,tong2024cambrian,laurencon2024whatmatters,fan2025webssl,li2025openvision} rather than studying whether separate pretraining is needed at all.
The few works that compare pretrained and from-scratch encoders vary in their findings, possibly because they control neither the encoder's pretraining data nor lifetime compute~\citep{tian2025navil,kimiteam2026kimik3}.
Studies raise a similar question for language pretraining, finding that adding vision before text-only pretraining ends, rather than after it, improves visual understanding~\citep{keh2025vlms,han2026physics,kimiteam2026kimik25}, as in our delayed-vision experiments (\cref{tab:understanding:schedule}).
\enlargethispage{\baselineskip}%
Encoder pretraining also confounds encoder \vs encoder-free comparisons, which typically pit a pretrained encoder against an encoder-free model~\citep{beyer2024paligemma,diao2024eve,lei2025sail,diao2025evev2,tong2026beyondlm,liu2026tuna2,chen2026encoderfree}.
Such comparisons remove separate pretraining, dedicated capacity, and pre-fusion placement at once, and often attribute the encoder's advantage to pretraining data they do not control.
\citet{shukor2025scaling}, who train both designs from scratch, find that encoder-free models match or slightly outperform an encoder.
In contrast, we find that the encoder leads on VQA and held-out caption loss, which we trace to their setup (\cref{sec:appx:shukor}).

\paragraph{Dedicated vision capacity.}
Testing whether dedicated capacity must precede fusion requires moving the encoder's capacity inside the backbone, which the encoder \vs encoder-free comparisons cannot accomplish.
Prior works on fusion depth come closer, varying where modalities begin to interact~\citep{nagrani2021attention,hao2026unix} or where encoder features enter the backbone~\citep{meng2024deepstack}, but none moves a fixed amount of vision-only capacity into layers that vision and text share.
Other multimodal architectures add vision-specific parameters to the backbone, but not in a form that fits this comparison.
Modality-expert designs duplicate the backbone's weights for vision tokens, typically in every layer, and route vision tokens through these copies instead of the shared weights~\citep{bao2022vlmo,lin2024moma,liang2025mixture}.
Their vision-only capacity is thus set by the backbone's size and spread over its depth rather than held at an encoder's budget early in the model, and most also build on a pretrained language model~\citep{wang2024cogvlm,luo2025monointernvl,tian2025navil,diao2025evev2,shi2024lmfusion}.
Adapter architectures fine-tune a pretrained model through small parallel or low-rank modules shared by all tokens~\citep{he2022unified,chen2022adaptformer,hu2022lora}, and their multimodal variants keep an encoder alongside the adapters~\citep{dong2024internlmxcomposer2,shapourian2026zaya1vl} or distill from one~\citep{wang2025vora}.
The encoders in these works still supply pre-fusion capacity or a pretrained prior, the two factors our comparison aims to remove.
We instead give vision tokens their own attention projections and feed-forward branches in selected backbone layers.
Their outputs add to the shared ones and together match an encoder's parameter budget, so only the placement of this capacity changes (\cref{sec:capacity}).
Placed in the first eight layers, in-fusion capacity is comparable to a pre-fusion encoder on VQA, so dedicated vision capacity helps whether it sits pre-fusion or in-fusion.

\section{Discussion}
\label{sec:discussion}
\label{sec:conclusion}

With multimodal models revisiting the vision encoder, we asked which of its design choices are still beneficial.
We proposed unbundling the encoder into three choices: a visual prior from separate pretraining, dedicated vision capacity, and pre-fusion placement of that capacity.
We tested each by training every recipe from scratch on common data, with one architecture and optimization setup, and comparing recipes on text and visual understanding benchmarks at matched lifetime compute.

\paragraph{Takeaways from unbundling.}
Of the three choices an encoder bundles, only dedicated vision capacity maintains its advantage.
Separate pretraining is unnecessary: an encoder trained jointly with the backbone is on par with pretrained ones, and longer pretraining improves recognition but not VQA, even when it is free (\cref{sec:encoders,sec:practical:free}).
Dedicated capacity matters: with the encoder's parameters added to their shared backbone, encoder-free models gain on Avg.\ Text but still trail the encoder on Avg.\ VQA (\cref{sec:encoder-free}), most on OCR and chart questions (\cref{tab:encoder-free:steelman}).
This capacity belongs early, but it need not precede the backbone: vision-only modules in the first backbone layers are comparable to a pre-fusion encoder (\cref{sec:capacity}).
For recipes that drop the encoder, this suggests reallocating its parameters as vision-only capacity in the earliest layers rather than as shared parameters.

\paragraph{Limitations and future work.}
\label{sec:limitations}
To keep comparisons controlled, we pretrain encoders on the same images as multimodal training, so our results do not rule out a benefit from released encoders trained on additional data.
Encoder pretraining uses the images' original alt-text, as in standard image--text pretraining; pretraining on our regenerated captions instead changes little (\cref{tab:appx:caption-source}).
Our experiments are on dense backbones with up to 6.4B parameters; the most compelling next step is extending them to sparse mixture-of-experts backbones, whose experts could naturally host in-fusion's modality-routed capacity.
Multimodal training also uses $224\times224$ MetaCLIP images, leaving higher-resolution training and more text-rich data to future work.
The text--image data mixture was selected with the from-scratch encoder at 1.7B, so mixtures tuned for each recipe remain underexplored (\cref{sec:encoders:mixtures}).
Finally, how architecture interacts with interleaved image--text and video data remains open.

\paragraph{Conclusion.}
We presented a controlled study that disentangles the design choices behind how multimodal models process visual input, studying each choice in isolation instead of as part of a bundled recipe.
Rather than assembling multimodal models from components optimized for different objectives, our findings favor training them end-to-end as one system.

\section*{Acknowledgments}
We thank
Jack Lu,
Philippe Hansen-Estruch, Fei Zhang,
Lili Yu,
Shengbang Tong, John Nguyen, and David Fan
for helpful discussions and feedback.

\phantomsection
\addcontentsline{toc}{section}{References}

\bibliographystyle{assets/plainnat}
\bibliography{paper}

\clearpage
\beginappendix
\crefalias{section}{appendix}\crefalias{subsection}{subappendix}\crefalias{subsubsection}{subsubappendix}

\addtocontents{toc}{\protect\setcounter{tocdepth}{1}}

\phantomsection
\addcontentsline{toc}{section}{Appendix}

\makeatletter
\setlength{\@fptop}{0pt}
\setlength{\@fpsep}{\floatsep}
\setlength{\@fpbot}{0pt plus 1fil}
\makeatother
\widowpenalty=10000
\clubpenalty=10000
\ifdefined\appxonly

  \FloatBarrier
\else
  
\phantomsection
\label{app:appendix}
\label{sec:appx:overview}

\begin{itemize}[topsep=0pt,parsep=0pt,partopsep=0pt,leftmargin=1.5em]
  \item[\S\ref{sec:appx:setup}] \textbf{Controlled protocol} (\cref{sec:protocol}): architectures, encoder design selection, optimization, training data and mixture (including when image--text data enters training), and compute accounting.
  \item[\S\ref{sec:appx:evaluation}] \textbf{Evaluation} (\cref{sec:protocol:evaluation}): VQA benchmarks with instruction tuning and the blank-image control, language benchmarks and the text-only reference, held-out losses, and ImageNet probes.
  \item[\S\ref{sec:appx:prior}] \textbf{Separate encoder pretraining} (\cref{sec:encoders,sec:practical:free}): objectives, upstream budgets, and pretraining at fixed multimodal compute.
  \item[\S\ref{sec:appx:dedicated-capacity}] \textbf{Dedicated vision capacity} (\cref{sec:encoder-free}): the encoder's advantage by VQA category, shared-capacity controls, richer patch embeddings, and the encoder learning rate behind the comparison with \citet{shukor2025scaling}.
  \item[\S\ref{sec:appx:visual-capacity}] \textbf{In-fusion vision capacity} (\cref{sec:capacity}): implementation and sizing; routing, depth, and attention-projection controls; the longest budget; and where vision tokens are processed.
  \item[\S\ref{sec:appx:scaling}] \textbf{Decay, learning rate, and caption loss} (\cref{sec:protocol:length,sec:encoder-free}): protocol checks, with every decayed endpoint, learning rate, and held-out caption loss as a proxy for Avg.\ VQA.
  \item[\S\ref{sec:appx:ladder-small}] \textbf{Scaling ladder} (\cref{sec:capacity:sizes}): the ladder's setup, the scaling laws fitted to it, and their predictions for runs they were not fitted on.
  \item[\S\ref{sec:appx:systems-full}] \textbf{Beyond lifetime compute: midtraining and inference} (\cref{sec:practical}): midtraining and its vision learning rate, training time, and inference latency.
  \item[\S\ref{sec:appx:per-benchmark}] \textbf{Per-benchmark results}: every VQA and text benchmark for each main-text exhibit.
\end{itemize}
Tables abbreviate the Avg.\ Text categories as Com (commonsense reasoning), Sci (science question answering), Knw (broad knowledge), and Rdg (reading comprehension), and the Avg.\ VQA categories as Gen (general), Knw (knowledge), OCR (OCR and chart), and Vis (vision-centric).
\cref{sec:appx:evaluation} defines them.

  \FloatBarrier

\section{Controlled protocol}
\label{sec:appx:setup}

Unless stated otherwise, every comparison of the 1.7B and 4B backbones uses the settings below, which detail the protocol of \cref{sec:protocol}.

\subsection{Architectures}
\label{sec:appx:architecture}
\label{sec:appx:model} %
\label{sec:appx:recipes} %

The multimodal recipes differ in where their vision-only capacity sits and how it is trained: in a pre-fusion vision encoder that is pretrained and then frozen or updated, or trained from scratch, encoder-free with no module besides a patch embedding, or vision-only modules inside the shared backbone (\ours; \cref{sec:vis-capacity:parameterization}).
Unless a variable is the stated ablation, comparisons hold the backbone, the 256 vision tokens per image in multimodal training, the data, the optimization recipe, and lifetime compute fixed.
\Cref{tab:appx:architectures} lists the backbone and encoder architectures.

\paragraph{Backbone.}
The shared backbone is a dense decoder-only Transformer that follows the Qwen3-1.7B and Qwen3-4B configurations~\citep{yang2025qwen3} and is trained from random initialization.
Its FFNs use SwiGLU~\citep{shazeer2020glu}, and its attention uses grouped key--value heads with RMSNorm on queries and keys.
The vision-only FFN branches of \ours use the same form and read the same normalized input as the shared FFN (\cref{sec:vis-capacity:parameterization}).

\begin{table}[!htbp]
  \centering
  \small
  \caption{
    \textbf{Every multimodal recipe shares one backbone configuration per scale, and the encoder adds a 0.4B ViT.}
    Parameter counts exclude the linear projection from the encoder into the backbone.
  }
  \label{tab:appx:architectures}
  
\setlength{\tabcolsep}{4pt}
\renewcommand{\arraystretch}{1.08}
\newcommand{\allscales}[1]{\multicolumn{2}{c@{\hspace{2em}}}{#1}}
\begin{tabular}{@{}lcc@{\hspace{2em}}c@{}}
  \toprule
  & \multicolumn{2}{>{\columncolor{PaperGray100!55}[.5\tabcolsep][.5\tabcolsep]}c@{\hspace{2em}}}{\textbf{Backbone}}
    & \multicolumn{1}{>{\columncolor{PaperGray100!55}[.5\tabcolsep][.5\tabcolsep]}c}{\textbf{Vision encoder}} \\
  \cmidrule(l{.5\tabcolsep}r{.5\tabcolsep}){2-3}\cmidrule(l{.5\tabcolsep}r{.5\tabcolsep}){4-4}
  & 1.7B & 4B & 0.4B \\
  \midrule
  Configuration & Qwen3-1.7B & Qwen3-4B & MoonViT-V2 \\
  Layers & 28 & 36 & 27 \\
  Width & 2,048 & 2,560 & 1,024 \\
  Query heads & 16 & 32 & 12 \\
  FFN width & 6,144 & 9,728 & 4,096 \\
  Tied embeddings & \cmark & \cmark & -- \\
  Parameters (B) & 1.72 & 4.02 & 0.40 \\
  Non-embedding parameters (B) & 1.41 & 3.63 & -- \\
  \arrayrulecolor{PaperGray400}\cmidrule{1-4}\arrayrulecolor{black}
  & \allscales{\textit{Shared}} & \\
  \cmidrule(r){2-3}
  Key--value heads & \allscales{8} & 12 \\
  Head dimension & \allscales{128} & 128 \\
  FFN activation & \allscales{SwiGLU} & GELU \\
  Normalization & \allscales{RMSNorm} & RMSNorm \\
  Position encoding & \allscales{1D RoPE} & learned 2D + 2D RoPE \\
  Input & \allscales{152K-token vocabulary} & $14\times14$ patches \\
  Tokens & \allscales{4,096 per sequence} & 256 per image \\
  \bottomrule
\end{tabular}

\end{table}

\paragraph{Vision encoder.}
The encoder follows MoonViT-V2, the vision encoder of Kimi K3~\citep{kimiteam2026kimik3}, in depth, patch size, normalization, and size: 27 layers, patch size 14, RMSNorm, and about 0.4B parameters in total.
It has width 1,024 with 4,096-wide MLPs and uses learned spatial position embeddings combined with interleaved two-dimensional rotary position embeddings.
During multimodal training, each image is resized bilinearly to a shorter side of 224 pixels, center-cropped to $224\times224$, and divided into $14\times14$ patches, producing 256 vision tokens.
For an image $x$, the encoder recipe computes
\begin{equation}
  v_{\mathrm{enc}}(x)
  = W_{\mathrm{proj}} f_{\phi}(x),
  \label{eq:appx:encoder-path}
\end{equation}
where $f_{\phi}$ is the vision encoder and the linear projector $W_{\mathrm{proj}}$ maps its output width to the backbone width.
Pretrained variants (\cref{sec:encoders}) use the same architecture and differ only in the pretraining objective.
MetaCLIP and WebSSL use a class token during upstream training; we discard it and pass only the 256 patch tokens to the backbone~\citep{liu2023visual}.

\paragraph{Encoder-free patch embedding.}
The encoder-free recipe projects each flattened $14\times14$ RGB patch linearly into the backbone width, as in Fuyu and Gemma~4~\citep{adept2023fuyu,gemmateam2026gemma4}, and applies RMSNorm and a learned scalar gate~\citep{qi2025beyond,li2026unseen}, giving 256 backbone tokens per image:
\begin{equation}
  v_{\mathrm{free}}(x)
  = g\,\operatorname{RMSNorm}
    \left(W_{\mathrm{patch}}\operatorname{Patchify}(x)+p_{2\mathrm{D}}\right),
  \label{eq:appx:encoder-free-path}
\end{equation}
where $p_{2\mathrm{D}}$ is a fixed two-dimensional sine--cosine position embedding and the learned scalar gate $g$ is initialized to $0.1$.
The in-fusion models of \cref{sec:vis-capacity:parameterization} use the same patch embedding.

\FloatBarrier
\subsection{Encoder design selection}
\label{sec:encoders:encoder-architecture}

We chose the from-scratch encoder's shape and size at matched lifetime compute, and every main comparison uses the resulting 27-layer, 0.4B encoder.
The shape and size comparisons use the 1.7B backbone at $2.6\times10^{21}$~FLOPs (\cref{tab:architecture:encoder-architecture}).

The configurations in \cref{tab:architecture:encoder-architecture} vary the structure of the dedicated visual computation.
Half-Depth, Wide, and Shallow + Wide are pre-fusion encoders that change the default encoder's depth, width, or both.
The backbone-style encoder is also a pre-fusion encoder.
It has its own patch embedding and eight Transformer blocks shaped like those of the 1.7B backbone, and it connects to the backbone through the linear projection of \cref{eq:appx:encoder-path}.
The Default, Shallow + Wide, and backbone-style variants use about the same vision-only parameter budget; the Half-Depth and Wide variants test smaller and larger budgets, respectively.
The ``Size'' section of \cref{tab:architecture:encoder-architecture} scales the default encoder's depth and width together, from Small (0.05B) to Huge (0.92B).
At matched lifetime compute, encoders with more parameters process fewer tokens.

\begin{table}[!htbp]
  \centering
  \small
  \caption{
    \textbf{The default 27-layer, 0.4B encoder gives the strongest VQA and comparable performance on text.}
    Larger encoders do not automatically translate to stronger visual understanding.
    All rows use a 1.7B backbone at $2.6\times10^{21}$~FLOPs of lifetime compute.
    Bold: within one point of the best.
  }
  \label{tab:architecture:encoder-architecture}
  \label{tab:architecture:encoder-scaling} %
  
\setlength{\tabcolsep}{2.5pt}
\renewcommand{\arraystretch}{1.05}
\begin{tabular}{@{}lrrrrrrrrrrrrr@{}}
  \toprule
  \multicolumn{4}{>{\columncolor{PaperGray100!55}[.5\tabcolsep][.5\tabcolsep]}c}{\textbf{Architecture}}
    & \multicolumn{5}{>{\columncolor{PaperBlue100!55}[.5\tabcolsep][.5\tabcolsep]}c}{\textbf{VQA (\%) $\uparrow$}}
    & \multicolumn{5}{>{\columncolor{PaperGreen100!55}[.5\tabcolsep][.5\tabcolsep]}c}{\textbf{Text (\%) $\uparrow$}} \\
  \cmidrule(l{.5\tabcolsep}r{.5\tabcolsep}){1-4}\cmidrule(l{.5\tabcolsep}r{.5\tabcolsep}){5-9}\cmidrule(l{.5\tabcolsep}r{.5\tabcolsep}){10-14}
  Variant & Depth & Width & \makecell{Vision-only\\params (B)}
    & \multicolumn{1}{c}{Avg}
    & \multicolumn{1}{c}{Gen}
    & \multicolumn{1}{c}{Knw}
    & \multicolumn{1}{c}{OCR}
    & \multicolumn{1}{c}{Vis}
    & \multicolumn{1}{c}{Avg}
    & \multicolumn{1}{c}{Com}
    & \multicolumn{1}{c}{Sci}
    & \multicolumn{1}{c}{Knw}
    & \multicolumn{1}{c@{}}{Rdg}
  \\
  \midrule
  \rowcolor{PaperGray100!50}
  \recipeglyph{scratch}Large (default) & 27 & 1,024 & 0.40 & \textbf{44.4} & \textbf{57.2} & 35.7 & \textbf{38.2} & 46.5 & 49.7 & 63.9 & 45.0 & 36.7 & 53.4 \\
  \addlinespace[2pt]
  \multicolumn{14}{@{}l}{\textit{Shape}} \\
  \recipeglyph{}Half-Depth & 14 & 1,024 & 0.21 & 43.0 & 54.3 & 35.0 & 34.8 & \textbf{47.8} & 49.6 & 63.6 & \textbf{45.8} & 37.5 & 51.5 \\
  \recipeglyph{}Wide & 27 & 2,048 & 1.26 & \textbf{44.0} & 56.1 & 35.1 & \textbf{37.8} & \textbf{47.1} & 49.2 & 63.6 & 44.8 & 35.2 & 53.2 \\
  \recipeglyph{}Shallow + Wide & 9 & 2,048 & 0.42 & 42.1 & 54.9 & 35.6 & 34.2 & 43.9 & \textbf{49.9} & \textbf{64.1} & 44.8 & 36.3 & 54.2 \\
  \recipeglyph{}Backbone-style & 8 & 2,048 & 0.40 & \textbf{43.9} & \textbf{56.9} & \textbf{38.2} & \textbf{37.7} & 42.8 & \textbf{50.1} & 63.6 & 44.1 & 37.7 & \textbf{54.9} \\
  \arrayrulecolor{PaperGray400}\cmidrule{1-14}\arrayrulecolor{black}
  \multicolumn{14}{@{}l}{\textit{Size}} \\
  \recipeglyph{}Small & 12 & 512 & 0.05 & 41.1 & 52.9 & 35.6 & 33.2 & 42.5 & \textbf{50.9} & 62.9 & \textbf{45.8} & \textbf{39.1} & \textbf{55.9} \\
  \recipeglyph{}Base & 18 & 768 & 0.15 & 42.5 & 54.6 & 34.8 & 34.5 & 46.2 & \textbf{49.9} & \textbf{64.8} & \textbf{46.1} & 35.4 & 53.2 \\
  \recipeglyph{}Huge & 40 & 1,280 & 0.92 & 42.6 & 55.0 & 35.8 & 36.1 & 43.4 & \textbf{50.7} & \textbf{65.1} & 44.8 & 37.8 & \textbf{55.1} \\
  \bottomrule
\end{tabular}

\end{table}

The default encoder is comparable to Wide on VQA while using fewer vision-only parameters.
Avg.\ Text does not follow encoder size: Small and Huge score highest, and the default slightly lower.

\paragraph{At 4B.}
We further ablate encoder size at the 4B model scale, using the same from-scratch recipe (\cref{tab:appx:encoder-scaling-4b}).
The default Large encoder is strongest on both Avg.\ Text and Avg.\ VQA.
At both scales, the default encoder gives the highest Avg.\ VQA, and Avg.\ Text does not follow encoder size.

\begin{table}[!htbp]
  \centering
  \small
  \caption{
    \textbf{On the 4B backbone, the Large encoder attains the strongest VQA.}
    All variants are trained to $5.8\times10^{21}$~FLOPs of lifetime compute.
    Bold: within one point of the best.
  }
  \label{tab:appx:encoder-scaling-4b}
  
\setlength{\tabcolsep}{3.5pt}
\renewcommand{\arraystretch}{1.05}
\begin{tabular}{@{}lrrrrrrrrrrr@{}}
  \toprule
  \multicolumn{2}{>{\columncolor{PaperGray100!55}[.5\tabcolsep][.5\tabcolsep]}c}{\textbf{Architecture}}
    & \multicolumn{5}{>{\columncolor{PaperBlue100!55}[.5\tabcolsep][.5\tabcolsep]}c}{\textbf{VQA (\%) $\uparrow$}}
    & \multicolumn{5}{>{\columncolor{PaperGreen100!55}[.5\tabcolsep][.5\tabcolsep]}c}{\textbf{Text (\%) $\uparrow$}} \\
  \cmidrule(l{.5\tabcolsep}r{.5\tabcolsep}){1-2}\cmidrule(l{.5\tabcolsep}r{.5\tabcolsep}){3-7}\cmidrule(l{.5\tabcolsep}r{.5\tabcolsep}){8-12}
  Variant & \makecell{Vision-only\\params (B)}
    & \multicolumn{1}{c}{Avg}
    & \multicolumn{1}{c}{Gen}
    & \multicolumn{1}{c}{Knw}
    & \multicolumn{1}{c}{OCR}
    & \multicolumn{1}{c}{Vis}
    & \multicolumn{1}{c}{Avg}
    & \multicolumn{1}{c}{Com}
    & \multicolumn{1}{c}{Sci}
    & \multicolumn{1}{c}{Knw}
    & \multicolumn{1}{c@{}}{Rdg}
  \\
  \midrule
  \recipeglyph{}Base & 0.15 & 48.5 & 58.7 & 41.8 & 40.6 & 52.7 & \textbf{55.9} & \textbf{67.2} & \textbf{49.0} & \textbf{49.2} & 58.1 \\
  \rowcolor{PaperGray100!50}
  \recipeglyph{scratch}Large (default) & 0.40 & \textbf{51.2} & \textbf{62.7} & \textbf{44.5} & \textbf{43.5} & \textbf{54.0} & \textbf{56.4} & \textbf{67.5} & \textbf{49.8} & \textbf{49.4} & \textbf{58.8} \\
  \recipeglyph{}Huge & 0.92 & 47.6 & 58.7 & 42.2 & 39.7 & 49.6 & \textbf{55.8} & \textbf{67.0} & \textbf{49.3} & 47.9 & \textbf{59.2} \\
  \bottomrule
\end{tabular}

\end{table}

\paragraph{At 6.4B.}
On the 6.4B backbone of the size ladder (\cref{sec:appx:size-ladder}), we also keep the 0.4B encoder instead of scaling it to 1.7B with the backbone.
It is comparable to the scaled encoder on both averages at every budget (\cref{tab:appx:encoder-size-6b}), so the default encoder does not limit the encoder recipe at larger backbones either.

\begin{table}[!htbp]
  \centering
  \small
  \caption{
    \textbf{At 6.4B, keeping the encoder at 0.4B is comparable to scaling it to 1.7B.}
    All runs follow the size ladder's recipe (\cref{sec:appx:size-ladder}).
    Bold: within one point of the best.
  }
  \label{tab:appx:encoder-size-6b}
  
\setlength{\tabcolsep}{4pt}
\renewcommand{\arraystretch}{1.05}
\begin{tabular}{@{}rrrrrrrr@{}}
  \toprule
  \multicolumn{2}{>{\columncolor{PaperGray100!55}[.5\tabcolsep][.5\tabcolsep]}c}{\textbf{Setup}} & \multicolumn{3}{>{\columncolor{PaperBlue100!55}[.5\tabcolsep][.5\tabcolsep]}c}{\textbf{VQA (\%) $\uparrow$}} & \multicolumn{3}{>{\columncolor{PaperGreen100!55}[.5\tabcolsep][.5\tabcolsep]}c}{\textbf{Text (\%) $\uparrow$}} \\
  \cmidrule(l{.5\tabcolsep}r{.5\tabcolsep}){1-2}\cmidrule(l{.5\tabcolsep}r{.5\tabcolsep}){3-5}\cmidrule(l{.5\tabcolsep}r{.5\tabcolsep}){6-8}
  TPP & FLOPs & \makecell[r]{0.4B\\encoder} & \makecell[r]{1.7B\\encoder} & \makecell[r]{Encoder-\\free} & \makecell[r]{0.4B\\encoder} & \makecell[r]{1.7B\\encoder} & \makecell[r]{Encoder-\\free} \\
  \midrule
  16 & $3.90{\times}10^{21}$ & \textbf{45.9} & \textbf{46.0} & 40.4 & \textbf{52.3} & \textbf{53.0} & \textbf{52.6} \\
  32 & $7.81{\times}10^{21}$ & \textbf{49.7} & \textbf{50.2} & 44.3 & \textbf{56.4} & \textbf{56.4} & \textbf{56.4} \\
  64 & $1.56{\times}10^{22}$ & \textbf{53.1} & \textbf{52.8} & 47.5 & \textbf{59.9} & \textbf{59.1} & \textbf{58.9} \\
  \bottomrule
\end{tabular}

\end{table}

\FloatBarrier
\subsection{Optimization settings}
\label{sec:appx:optimization}

\Cref{tab:appx:hyperparameters} lists the optimization settings for multimodal training and instruction tuning.
We train the Transformer weight matrices with Muon~\citep{jordan2024muon}, applied per head to the query, key, and value projections~\citep{kimiteam2026kimik3}, and the embeddings, output heads, normalization parameters, and biases with AdamW.
The peak learning rate scales with the inverse square root of the backbone's total parameter count, and training follows a warmup--stable--decay schedule~\citep{hu2024minicpm,wen2024wsd,hagele2024scaling}.
Checkpoints taken before the final decay use a constant learning rate.
The decay lowers the learning rate linearly to 5\% of its peak over the last 10\% of updates, so a run whose stable phase ends at compute $C$ finishes at $C/0.9$ (for example, $1.17\times10^{22}$ and $1.30\times10^{22}$~FLOPs at 1.7B and $2.61\times10^{22}$ and $2.90\times10^{22}$~FLOPs at 4B; \cref{tab:protocol:budgets}).

\begin{table}[!htbp]
  \centering
  \small
  \caption{
    \textbf{The 1.7B and 4B backbones share one optimization recipe, with the peak learning rate scaled to backbone size.}
    Muon and AdamW use the same peak learning rate.
  }
  \label{tab:appx:hyperparameters}
  
\setlength{\tabcolsep}{6pt}
\renewcommand{\arraystretch}{1.08}
\begin{tabular}{@{}lcc@{}}
  \toprule
  & \multicolumn{1}{>{\columncolor{PaperGray100!55}[.5\tabcolsep][.5\tabcolsep]}c}{\textbf{Multimodal training}}
    & \multicolumn{1}{>{\columncolor{PaperGray100!55}[.5\tabcolsep][.5\tabcolsep]}c}{\textbf{Instruction tuning}} \\
  \midrule
  Optimizer & Muon, AdamW & AdamW \\
  Peak learning rate & $3\times10^{-4}$ (1.7B), $2\times10^{-4}$ (4B) & $1\times10^{-5}$ \\
  Warmup updates & 4,000 & 1,000 \\
  Schedule & warmup--stable--decay & cosine \\
  Final learning rate & 5\% of peak & 5\% of peak \\
  Weight decay & 0.1 & 0.1 \\
  Gradient-norm clip & 1.0 & 5.0 \\
  Batch (sequences $\times$ length) & $1{,}024\times4{,}096$ & $256\times4{,}096$ \\
  Data & text and image--text, 70:30 & Cambrian-7M, 3 epochs \\
  \bottomrule
\end{tabular}

\end{table}

\FloatBarrier
\subsection{Training data and mixture}
\label{sec:appx:data-mixture}

Text-only sequences come from Nemotron-CC v2~\citep{nvidia2025nemotronnano2} and v2.1~\citep{nvidia2025nemotron3nano}, and image--text sequences use MetaCLIP 1.2 image--text pairs (English subset)~\citep{xu2023metaclip}.
Every training sequence has 4,096 tokens and packs several documents.
Attention is bidirectional among the vision tokens of each image and causal otherwise.
It stays within each document in image--text sequences and in every sequence of the text-only reference, whereas the text sequences of image-trained models attend across packed documents.
Multimodal training uses a 70:30 mixture of text and image--text sequences, which we select below (\cref{tab:architecture:mixtures}).
Each image is processed at $224\times224$ resolution and yields 256 vision tokens (\cref{sec:appx:architecture}).
An image--text sequence packs about 12.8 MetaCLIP 1.2 pairs, each one image and its caption, so vision tokens fill about 80\% of an image--text sequence and, under the 70:30 mixture, about 24\% of all training tokens.
Below, the data ablation varies the mixture and the study of when image--text data enters training uses a 90:10 mixture; the richer patch embeddings of \cref{sec:appx:encoder-free-tokenization} read higher-resolution images.

\paragraph{Recaptioning.}
Multimodal training uses captions regenerated once for the MetaCLIP 1.2 images by Qwen2.5-VL-32B-Instruct~\citep{bai2025qwen25vl} and shared by every recipe, while encoder pretraining uses the original alt-text (\cref{sec:protocol}).
We use nucleus sampling with temperature 0.6 and top-$p$ 0.95, generating at most 128 new tokens.
The system prompt is:
\begin{quote}
\ttfamily\small
You are an expert image captioner. Write one specific,
information-dense caption (1{\rmfamily\textendash}2 sentences). Do not begin with
\textquotesingle A\textquotesingle, \textquotesingle An\textquotesingle, or \textquotesingle The\textquotesingle. Name the main subject first, then key
objects, attributes, actions, and scene context. Avoid hedging and
meta phrases.
\end{quote}
The user prompt is:
\begin{quote}
\ttfamily\small
Generate a caption for this image.
\end{quote}

\paragraph{Caption source for encoder pretraining.}
To ablate the choice of regenerated captions versus alt-text, we pretrain the CapPa encoder for $7.7\times10^{20}$~FLOPs on either the alt-text or our regenerated captions and train it with the 1B ladder backbone (\cref{tab:appx:size-ladder}), a 0.4B encoder, and the 70:30 mixture, against the from-scratch encoder, to 32 and 64 tokens per parameter (\cref{tab:appx:caption-source}).
The regenerated-caption encoder scores slightly higher Avg.\ VQA than the alt-text encoder and matches the from-scratch encoder.

\begin{table}[!htbp]
  \centering
  \small
  \caption{
    \textbf{Pretraining the encoder on regenerated captions slightly outperforms alt-text.}
    All variants are trained to either $1.8\times10^{20}$ or $3.7\times10^{20}$~FLOPs of multimodal training (32 and 64 TPP, respectively).
  }
  \label{tab:appx:caption-source}
  
\renewcommand{\arraystretch}{1.05}
\setlength{\tabcolsep}{4pt}
\begin{tabular}{@{}lrrrrrrrrrrr@{}}
  \toprule
  \multicolumn{2}{>{\columncolor{PaperGray100!55}[.5\tabcolsep][.5\tabcolsep]}c}{\textbf{Setup}}
    & \multicolumn{5}{>{\columncolor{PaperBlue100!55}[.5\tabcolsep][.5\tabcolsep]}c}{\textbf{VQA (\%) $\uparrow$}}
    & \multicolumn{5}{>{\columncolor{PaperGreen100!55}[.5\tabcolsep][.5\tabcolsep]}c}{\textbf{Text (\%) $\uparrow$}} \\
  \cmidrule(l{.5\tabcolsep}r{.5\tabcolsep}){1-2}\cmidrule(l{.5\tabcolsep}r{.5\tabcolsep}){3-7}\cmidrule(l{.5\tabcolsep}r{.5\tabcolsep}){8-12}
  Encoder & TPP & \multicolumn{1}{c}{Avg} & \multicolumn{1}{c}{Gen} & \multicolumn{1}{c}{Knw} & \multicolumn{1}{c}{OCR} & \multicolumn{1}{c}{Vis} & \multicolumn{1}{c}{Avg} & \multicolumn{1}{c}{Com} & \multicolumn{1}{c}{Sci} & \multicolumn{1}{c}{Knw} & \multicolumn{1}{c@{}}{Rdg} \\
  \midrule
  \rowcolor{PaperGray100!50}
  \recipeglyph{scratch}Scratch encoder & 32 & 28.6 & 33.8 & 27.5 & 9.7 & 43.4 & 42.8 & 56.5 & 39.6 & 26.4 & 48.6 \\
  \recipeglyph{cappa}CapPa, alt-text & 32 & 28.7 & 36.5 & 26.4 & 8.0 & 43.9 & 43.5 & 57.1 & 40.7 & 26.8 & 49.2 \\
  \recipeglyph{cappa}CapPa, recaptioned & 32 & 29.2 & 37.6 & 26.9 & 8.8 & 43.5 & 42.9 & 57.9 & 39.4 & 26.2 & 48.1 \\
  \midrule
  \rowcolor{PaperGray100!50}
  \recipeglyph{scratch}Scratch encoder & 64 & 34.6 & 43.0 & 27.9 & 22.1 & 45.2 & 44.7 & 58.9 & 41.6 & 27.3 & 50.9 \\
  \recipeglyph{cappa}CapPa, alt-text & 64 & 33.2 & 42.2 & 28.1 & 19.1 & 43.5 & 44.7 & 59.2 & 41.1 & 27.6 & 50.7 \\
  \recipeglyph{cappa}CapPa, recaptioned & 64 & 33.9 & 43.1 & 28.8 & 20.7 & 42.9 & 44.9 & 59.1 & 42.1 & 27.8 & 50.4 \\
  \bottomrule
\end{tabular}

\end{table}

\paragraph{Data mixture.}
\label{sec:encoders:mixtures}

We select the 70:30 mixture of text and image--text sequences by an ablation on the 1.7B backbone with a from-scratch vision encoder, consistent with controlled studies of data balance in early multimodal training~\citep{han2026physics}.
At the same lifetime compute ($2.34\times10^{21}$~FLOPs, at a constant learning rate), we compare it with 100:0, 90:10, and 50:50 mixtures (\cref{tab:architecture:mixtures}).
The 70:30 mixture gives the best Avg.\ VQA with slight degradation to text, whereas 90:10 keeps Avg.\ Text closer to the text-only model.

\begin{table}[!htbp]
  \centering
  \small
  \caption{
    \textbf{The 70:30 mixture favors visual understanding, whereas 90:10 preserves more language ability.}
    All mixtures use a 1.7B backbone at $2.34\times10^{21}$~FLOPs of lifetime compute at a constant learning rate.
    For 100:0, images are replaced with zeros during instruction tuning and evaluation, so its starred VQA entries are blank-image controls (\cref{sec:appx:image-controls}).
  }\label{tab:architecture:mixtures}
  
\renewcommand{\arraystretch}{1.05}
\setlength{\tabcolsep}{4pt}
\begin{tabular}{@{}lrrrrrrrrrr@{}}
  \toprule
  \multicolumn{1}{>{\columncolor{PaperGray100!55}[.5\tabcolsep][.5\tabcolsep]}c}{\textbf{Data mixture}}
    & \multicolumn{5}{>{\columncolor{PaperBlue100!55}[.5\tabcolsep][.5\tabcolsep]}c}{\textbf{VQA (\%) $\uparrow$}}
    & \multicolumn{5}{>{\columncolor{PaperGreen100!55}[.5\tabcolsep][.5\tabcolsep]}c}{\textbf{Text (\%) $\uparrow$}} \\
  \cmidrule(l{.5\tabcolsep}r{.5\tabcolsep}){1-1}\cmidrule(l{.5\tabcolsep}r{.5\tabcolsep}){2-6}\cmidrule(l{.5\tabcolsep}r{.5\tabcolsep}){7-11}
  Text/image--text & \multicolumn{1}{c}{Avg} & \multicolumn{1}{c}{Gen} & \multicolumn{1}{c}{Knw} & \multicolumn{1}{c}{OCR} & \multicolumn{1}{c}{Vis} & \multicolumn{1}{c}{Avg} & \multicolumn{1}{c}{Com} & \multicolumn{1}{c}{Sci} & \multicolumn{1}{c}{Knw} & \multicolumn{1}{c@{}}{Rdg} \\
  \midrule
  \rowcolor{PaperGray100!50}
  100:0 & 28.0\rlap{$^*$} & 33.4\rlap{$^*$} & 31.3\rlap{$^*$} & 3.7\rlap{$^*$} & 43.5\rlap{$^*$} & 50.1 & 64.3 & 44.8 & 36.2 & 55.3 \\
  90:10 & 41.2 & 53.1 & 34.5 & 31.0 & 46.3 & \textbf{49.3} & 62.8 & 44.3 & \textbf{36.3} & \textbf{53.9} \\
  70:30 & \textbf{43.4} & \textbf{55.2} & \textbf{36.0} & \textbf{37.0} & 45.6 & 48.0 & 62.6 & \textbf{44.5} & 31.7 & 53.2 \\
  50:50 & 43.0 & 54.3 & 35.2 & 35.9 & \textbf{46.4} & 47.7 & \textbf{63.7} & 44.2 & 31.4 & 51.5 \\
  \bottomrule
\end{tabular}

\end{table}

We use 70:30 for the main comparisons as it gives a strong balance of visual understanding and language ability.

\paragraph{When image--text data enters.}
\label{sec:appx:onset}
\label{sec:encoders:curricula} %
We also vary the point in training at which image--text data enters.
Every schedule allocates 10\% of lifetime training examples to image--text data, a 90:10 mixture rather than the constant 70:30 mixture above.
We compare schedules that begin image--text training at different points while keeping the total number of image--text examples and lifetime compute fixed (\cref{fig:appx:curriculum-schedules}).
Later schedules continue the language model and optimizer, add a randomly initialized vision encoder and projector, and warm up the new parameters for 4,000 updates.
Starting image--text training at the first update gives the strongest visual understanding.
Starting at 25\%, 50\%, or 75\% of training lowers Avg.\ VQA, with the largest drops on general and on OCR and chart questions, while Avg.\ Text stays comparable (\cref{tab:understanding:schedule}).

\begin{figure}[!htbp]
  \centering
  \includegraphics[width=\linewidth]{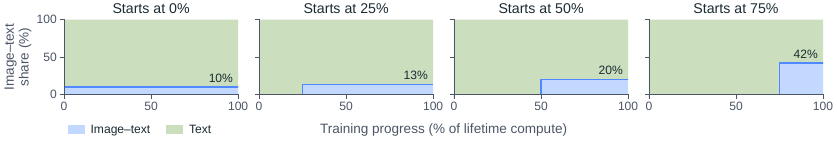}
  \caption{\textbf{Later starts concentrate the same image--text data into fewer updates.}
  Every schedule sees the same number of image--text examples, so the blue areas are about equal.
  Labels give each schedule's final image--text fraction.}
  \label{fig:appx:curriculum-schedules}
\end{figure}

\begin{table}[!htbp]
  \centering
  \small
  \caption{
    \textbf{Starting image--text training from the first update gives the strongest visual understanding.}
    All rows use a 1.7B backbone at $2.6\times10^{21}$~FLOPs of lifetime compute and a 90:10 mixture overall, scheduled as in \cref{fig:appx:curriculum-schedules}.
    Bold: within one point of the best.
  }
  \label{tab:understanding:schedule}
  
\setlength{\tabcolsep}{4.0pt}
\renewcommand{\arraystretch}{1.05}
\begin{tabular}{@{}lrrrrrrrrrr@{}}
  \toprule
  \multicolumn{1}{>{\columncolor{PaperGray100!55}[.5\tabcolsep][.5\tabcolsep]}c}{\textbf{Image--text start}}
    & \multicolumn{5}{>{\columncolor{PaperBlue100!55}[.5\tabcolsep][.5\tabcolsep]}c}{\textbf{VQA (\%) $\uparrow$}}
    & \multicolumn{5}{>{\columncolor{PaperGreen100!55}[.5\tabcolsep][.5\tabcolsep]}c}{\textbf{Text (\%) $\uparrow$}} \\
  \cmidrule(l{.5\tabcolsep}r{.5\tabcolsep}){1-1}\cmidrule(l{.5\tabcolsep}r{.5\tabcolsep}){2-6}\cmidrule(l{.5\tabcolsep}r{.5\tabcolsep}){7-11}
  {} & \multicolumn{1}{c}{Avg} & \multicolumn{1}{c}{Gen} & \multicolumn{1}{c}{Knw} & \multicolumn{1}{c}{OCR} & \multicolumn{1}{c}{Vis} & \multicolumn{1}{c}{Avg} & \multicolumn{1}{c}{Com} & \multicolumn{1}{c}{Sci} & \multicolumn{1}{c}{Knw} & \multicolumn{1}{c@{}}{Rdg} \\
  \midrule
  From first update & \textbf{42.5} & \textbf{55.1} & \textbf{36.0} & \textbf{32.5} & \textbf{46.2} & 50.8 & \textbf{64.6} & \textbf{46.2} & 38.0 & \textbf{54.4} \\
  25\% start & 35.0 & 43.8 & 31.9 & 19.4 & 44.8 & \textbf{52.0} & \textbf{64.6} & \textbf{46.5} & \textbf{41.9} & \textbf{55.0} \\
  50\% start & 35.1 & 43.8 & 34.0 & 17.0 & \textbf{45.5} & \textbf{51.0} & \textbf{64.2} & \textbf{46.3} & 40.2 & 53.1 \\
  75\% start & 34.4 & 42.8 & 33.7 & 17.2 & 44.0 & 50.8 & \textbf{64.6} & \textbf{46.0} & 39.2 & 53.4 \\
  \bottomrule
\end{tabular}

\end{table}

\FloatBarrier
\subsection{Compute accounting}
\label{sec:appx:lifetime-flops}
\label{sec:appx:compute-conversion} %

Lifetime compute (\cref{eq:lifetime-compute}) sums the training compute of every stage that produces the final model: multimodal training for every recipe, encoder pretraining for pretrained encoders, and, in midtraining (\cref{sec:capacity:late-onset}), the text-only pretraining of the language model.
As with fine-tuning in transfer evaluation, instruction tuning is a fixed downstream step applied to every checkpoint to measure VQA (\cref{sec:appx:evaluation}), so it is not counted.
Each term uses the same algorithmic count of forward- and backward-pass matrix multiplication and convolution FLOPs for the modules active in that stage~\citep{kaplan2020scaling,hoffmann2022chinchilla}.
The count excludes activation recomputation~\citep{chowdhery2022palm}, optimizer updates, communication, elementwise operations, and input processing.
A frozen encoder receives no backward pass during multimodal training, so it is charged only its forward pass, one third of its training cost.
The free-encoder comparison of \cref{sec:practical:free} instead matches multimodal training compute and does not charge encoder pretraining.

\paragraph{Charging pretrained checkpoints.}
A pretrained encoder is charged the full compute of its upstream training up to the checkpoint used.
Because a MetaCLIP example costs less compute than a captioning example, the same upstream compute buys MetaCLIP about 12\% more epochs.
The objective comparisons hold upstream compute equal across objectives, so they use MetaCLIP checkpoints matched to the compute of 1, 2, 4, and 8 captioning epochs (\cref{sec:appx:encoder-objectives}).
The sweep of \cref{sec:appx:upstream-fixed} compares MetaCLIP only with itself, so it uses MetaCLIP's own 2-, 4-, 8-, and 13-epoch checkpoints.

\paragraph{Token-equivalent budgets.}

To provide intuition about compute scales, we also express each budget in \emph{token-equivalents}, the number of tokens that a text-only model of the same backbone size would train on with the same compute:
\begin{equation}
  H_{\mathrm{eq}} = \frac{C_{\mathrm{lifetime}}}{c_{\mathrm{LM}}},
  \label{eq:appx:equivalent-horizon}
\end{equation}
where $c_{\mathrm{LM}}$ is that text-only model's training FLOPs per token, $1.17\times10^{10}$ at 1.7B and $2.61\times10^{10}$ at 4B.
For example, a budget of 200B token-equivalents is $2.34\times10^{21}$~FLOPs at 1.7B, and \cref{tab:protocol:budgets} gives every decayed budget in FLOPs and tokens per parameter; a constant-learning-rate checkpoint sits at 0.9 times its decayed budget (\cref{sec:appx:optimization}).

\begin{table}[!htbp]
  \centering
  \small
  \caption{
    \textbf{Compute budgets can be expressed as token-equivalents, FLOPs, or TPP.}
    TPP divides by each backbone's non-embedding parameter count.
  }
  \label{tab:protocol:budgets}
  
\setlength{\tabcolsep}{5pt}
\renewcommand{\arraystretch}{1.05}
\begin{tabular}{@{}lrrrr@{}}
  \toprule
   & \multicolumn{2}{>{\columncolor{PaperGray100!55}[.5\tabcolsep][.5\tabcolsep]}c}{\textbf{1.7B}} & \multicolumn{2}{>{\columncolor{PaperGray100!55}[.5\tabcolsep][.5\tabcolsep]}c}{\textbf{4B}} \\
  \cmidrule(l{.5\tabcolsep}r{.5\tabcolsep}){2-3}\cmidrule(l{.5\tabcolsep}r{.5\tabcolsep}){4-5}
  Budget & \multicolumn{1}{c}{FLOPs} & \multicolumn{1}{c}{TPP} & \multicolumn{1}{c}{FLOPs} & \multicolumn{1}{c@{}}{TPP} \\
  \midrule
  55.6B & $6.5{\times}10^{20}$ & 39 & $1.45{\times}10^{21}$ & 15 \\
  111.1B & $1.30{\times}10^{21}$ & 79 & $2.90{\times}10^{21}$ & 31 \\
  222.2B & $2.60{\times}10^{21}$ & 158 & $5.79{\times}10^{21}$ & 61 \\
  444.4B & $5.21{\times}10^{21}$ & 315 & $1.16{\times}10^{22}$ & 122 \\
  666.7B & $7.81{\times}10^{21}$ & 473 & $1.74{\times}10^{22}$ & 183 \\
  888.9B & $1.04{\times}10^{22}$ & 631 & $2.32{\times}10^{22}$ & 245 \\
  1.11T & $1.30{\times}10^{22}$ & 788 & $2.90{\times}10^{22}$ & 306 \\
  \bottomrule
\end{tabular}

\end{table}

Architecture and data mixture change the number of updates at fixed compute.
For instance, at $2.34\times10^{21}$~FLOPs, the 1.7B text-only model takes 47.7K updates of $1{,}024\times4{,}096$ tokens (200B tokens), and the from-scratch encoder recipe with the 70:30 mixture takes 45.3K, because the encoder's computation also counts toward the budget.

  \FloatBarrier
  
\section{Evaluation}
\label{sec:appx:evaluation}

This section describes the two main evaluation metrics, Avg.\ VQA and Avg.\ Text.
We also define the held-out losses that \cref{sec:encoder-free,sec:capacity:sizes,sec:appx:scaling:loss} use as well as two additional metrics, out-of-distribution caption loss and ImageNet linear-probe accuracy.

\subsection{VQA benchmarks}
\label{sec:appx:understanding}
\label{sec:appx:image-controls}

\paragraph{Visual instruction tuning.}
We use instruction tuning as an evaluation protocol~\citep{tong2024cambrian}.
Every multimodal model checkpoint is instruction-tuned with one fixed recipe before we measure its visual understanding.
The recipe runs three epochs of Cambrian-7M~\citep{tong2024cambrian} before evaluation.
Multimodal training reads $224\times224$ images, but instruction tuning and evaluation raise the resolution to $448\times448$ (1,024 vision tokens), since OCR and chart reading needs fine detail~\citep{tong2024cambrian,beyer2024vitspeed}.
Exhibits whose captions say so instead retain the multimodal-training resolution of $224\times224$ (256 vision tokens), otherwise with the same recipe.
The encoder's learned position table is interpolated to the larger patch grid, and the fixed sine--cosine positions of encoder-free and in-fusion models are recomputed for it.
Cambrian-7M is a curated instruction-tuning mixture spanning general visual tasks, OCR and charts, counting, math, science, code, and language-only data.
\Cref{tab:appx:hyperparameters} provides the optimizer settings and batch size.
Instruction tuning updates the same parameters as multimodal training.
Therefore, a frozen encoder, e.g., the frozen MetaCLIP encoder in \cref{tab:encoders:objectives}, stays frozen.

\paragraph{Benchmarks and Avg.\ VQA.}
Evaluation follows the Cambrian-1 visual understanding suite~\citep{tong2024cambrian}, using greedy decoding with at most 256 output tokens.
Each task is capped at its first 5{,}000 examples in dataset order.
\Cref{tab:appx:vqa-benchmarks} lists the 16 benchmarks in four categories (general, knowledge, OCR and chart, and vision-centric), with the split, metric, and number of examples for each.
Exact match compares the normalized parsed answer or option letter with the reference answers, relaxed accuracy tolerates small numeric errors, and VQA accuracy is the consensus metric~\citep{antol2015vqa} that TextVQA uses.
Each category mean averages its four benchmarks.
The suite contains 15 benchmarks scored in $[0,100]$ and MME-Perception (MME-P), whose ten subtasks each score out of 200.
We divide the MME-P total by 20 to put it on the same scale, and Avg.\ VQA is the mean of the 16 scores, which equals the mean of the four category means.

\begin{table}[!htbp]
  \centering
  \small
  \caption{
    \textbf{Avg.\ VQA weights four categories of visual understanding benchmarks,}
    Each category contains four tasks.
    MME-P is normalized to $[0,100]$ before averaging.
    Examples counts the questions evaluated.
  }\label{tab:appx:vqa-benchmarks}
  
\setlength{\tabcolsep}{4.0pt}
\renewcommand{\arraystretch}{1.08}
\begin{tabular}{@{}ll@{\hspace{1.5em}}llr@{}}
  \toprule
  Benchmark & Reference & Split & Metric & Examples \\
  \midrule
  \multicolumn{5}{@{}l}{\textit{General (Gen)}} \\
  MME-P & \citet{fu2023mme} & test & MME score $/\,20$ & 2{,}114 \\
  MMBench-EN & \citet{liu2023mmbench} & dev & Exact match & 4{,}329 \\
  SEED-Bench & \citet{li2023seedbench} & test & Exact match & 5{,}000 \\
  GQA & \citet{hudson2019gqa} & test-dev & Exact match & 5{,}000 \\
  \arrayrulecolor{PaperGray400}\cmidrule{1-5}\arrayrulecolor{black}
  \multicolumn{5}{@{}l}{\textit{Knowledge (Knw)}} \\
  ScienceQA (image) & \citet{lu2022scienceqa} & test & Exact match & 2{,}017 \\
  MMMU & \citet{yue2024mmmu} & validation & Exact match & 900 \\
  MathVista & \citet{lu2024mathvista} & testmini & Exact match & 936 \\
  AI2D & \citet{kembhavi2016ai2d} & test & Exact match & 3{,}088 \\
  \arrayrulecolor{PaperGray400}\cmidrule{1-5}\arrayrulecolor{black}
  \multicolumn{5}{@{}l}{\textit{OCR and chart (OCR)}} \\
  ChartQA & \citet{masry2022chartqa} & test & Relaxed accuracy & 2{,}500 \\
  OCRBench & \citet{liu2024ocrbench} & test & Exact match & 1{,}000 \\
  TextVQA & \citet{singh2019textvqa} & validation & VQA accuracy & 5{,}000 \\
  DocVQA & \citet{mathew2021docvqa} & validation & Exact match & 5{,}000 \\
  \arrayrulecolor{PaperGray400}\cmidrule{1-5}\arrayrulecolor{black}
  \multicolumn{5}{@{}l}{\textit{Vision-centric (Vis)}} \\
  MMVP & \citet{tong2024eyes} & full & Exact match & 300 \\
  RealWorldQA & \citet{xai2024grok15v} & test & Exact match & 765 \\
  CV-Bench-2D & \citet{tong2024cambrian} & test & Exact match & 1{,}438 \\
  CV-Bench-3D & \citet{tong2024cambrian} & test & Exact match & 1{,}200 \\
  \bottomrule
\end{tabular}

\end{table}

\paragraph{Blank-image control.}
The \emph{blank-image} control separates visual understanding from what the question alone supports.
It replaces every image with zeros during both instruction tuning and evaluation while keeping the vision-token positions.
We use it only for the text-only (100:0) row of the data-mixture ablation (\cref{tab:architecture:mixtures}), which has no trained visual pathway.
At 1.7B, this row reaches 28 Avg.\ VQA, a measure of how much the questions alone support~\citep{goyal2017making,chen2024mmstar,brown2025benchmark}.

\subsection{Language benchmarks}
\label{sec:appx:language}

Language evaluations use the multimodal checkpoint without instruction tuning.
\Cref{tab:appx:language-benchmarks} lists the 12 benchmarks, their categories, and how each is scored.
All tasks rank the answer choices by log-likelihood; MMLU uses five subject-matched demonstrations and averages its 57 subjects equally, and the other tasks are zero-shot.
Avg.\ Text is the mean of the four category scores, each the mean of its benchmarks, with ARC-Easy and ARC-Challenge averaged into one ARC score and RACE-middle and RACE-high combined by test-question count.

\begin{table}[!htbp]
  \centering
  \small
  \caption{
    \textbf{Avg.\ Text averages four categories of text benchmarks.}
  }\label{tab:appx:language-benchmarks}
  
\setlength{\tabcolsep}{4.0pt}
\renewcommand{\arraystretch}{1.08}
\begin{tabular}{@{}ll@{\hspace{1.5em}}cl@{}}
  \toprule
  Task & Reference & Shots & Scoring \\
  \midrule
  \multicolumn{4}{@{}l}{\textit{Commonsense reasoning (Com)}} \\
  COPA & \citet{roemmele2011copa} & 0 & Summed log-likelihood \\
  HellaSwag & \citet{zellers2019hellaswag} & 0 & Mean token log-likelihood \\
  PIQA & \citet{bisk2020piqa} & 0 & Mean token log-likelihood \\
  Social IQa & \citet{sap2019socialiqa} & 0 & Summed log-likelihood \\
  WinoGrande & \citet{sakaguchi2020winogrande} & 0 & Summed log-likelihood \\
  \arrayrulecolor{PaperGray400}\cmidrule{1-4}\arrayrulecolor{black}
  \multicolumn{4}{@{}l}{\textit{Science question answering (Sci)}} \\
  ARC-Easy & \citet{clark2018arc} & 0 & Mean token log-likelihood \\
  ARC-Challenge & \citet{clark2018arc} & 0 & Mean token log-likelihood \\
  OpenBookQA & \citet{mihaylov2018openbookqa} & 0 & Mean token log-likelihood \\
  \arrayrulecolor{PaperGray400}\cmidrule{1-4}\arrayrulecolor{black}
  \multicolumn{4}{@{}l}{\textit{Broad knowledge (Knw)}} \\
  MMLU (57 subjects) & \citet{hendrycks2021mmlu} & 5 & Likelihood ranking, macro-average \\
  \arrayrulecolor{PaperGray400}\cmidrule{1-4}\arrayrulecolor{black}
  \multicolumn{4}{@{}l}{\textit{Reading comprehension (Rdg)}} \\
  BoolQ & \citet{clark2019boolq} & 0 & Summed log-likelihood \\
  RACE-middle (1{,}436 test) & \citet{lai2017race} & 0 & Summed log-likelihood \\
  RACE-high (3{,}498 test) & \citet{lai2017race} & 0 & Summed log-likelihood \\
  \bottomrule
\end{tabular}

\end{table}

\subsection{Text-only reference}
\label{sec:appx:scaling:text}
At the same text-token budget, the multimodal recipes score about as well on Avg.\ Text as a text-only reference trained on the same text: on average 0.1 point below it at 1.7B and 0.6 below at 4B (\cref{fig:appx:text-cost}).
The encoder, encoder-free, and in-fusion recipes behave alike.
At matched lifetime compute, the reference leads by 0.5 to 3.9 points (\cref{tab:appx:ladder}), because multimodal training gives 30\% of its sequences to image--text data and so sees about 30\% fewer text tokens.
Vision therefore costs text little beyond the text data it displaces.

\begin{figure}[!htbp]
  \centering
  \includegraphics[width=\linewidth]{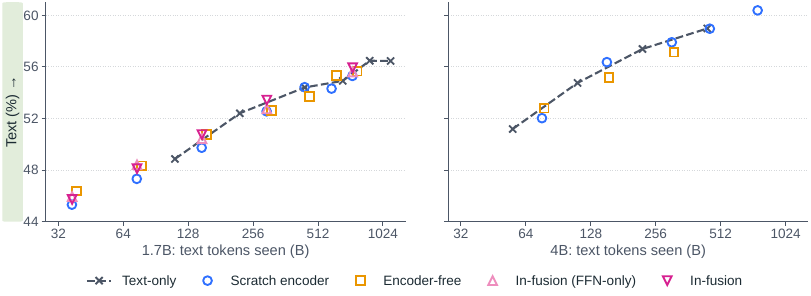}
  \caption{
    \textbf{At the same text-token budget, the multimodal recipes score about as well on Avg.\ Text as the text-only reference.}
    Each point is one decayed endpoint, placed by the text-only tokens it trained on (70\% of multimodal training).
  }
  \label{fig:appx:text-cost}
\end{figure}

\subsection{Held-out losses}
\label{sec:appx:perplexity}

We measure losses on two kinds of held-out data, both as per-token negative log-likelihood in nats.
The held-out text and caption losses score unseen samples of the training distributions, and \cref{sec:encoder-free,sec:capacity:sizes,sec:appx:scaling:loss} use them.
Out-of-distribution caption loss scores caption sets outside those distributions, and \cref{tab:capacity:qkvo-mechanisms} reports it.

\paragraph{In-distribution.}
Held-out text loss is the per-token negative log-likelihood on held-out text from the training text corpus, weighted as in the training mixture.
It is scored with the attention masking each model was trained with.
Held-out caption loss is the per-token negative log-likelihood of the regenerated captions of MetaCLIP 1.2 images never used in training, so it samples the training caption distribution.

\paragraph{Out-of-distribution.}
Out-of-distribution caption loss averages the caption loss of three sets of 5{,}000 images with equal weight: the COCO validation images with the first of their human reference captions~\citep{lin2014coco}, MetaCLIP 1.2 pairs with their original alt-text~\citep{xu2023metaclip}, and images from a licensed stock-photo collection.
It is scored with the attention masking each model was trained with.

\subsection{ImageNet linear probes}
\label{sec:appx:probe}
We use three linear probes on ImageNet-1K~\citep{deng2009imagenet}.
The \emph{downstream} probe reads the multimodal model with all weights frozen.
It mean-pools the backbone's final normalized vision-token states and fits an $\ell_2$-regularized linear classifier on 50 training images per class.
A fixed 10\% holdout selects the regularization strength from $\{10^{-4},10^{-3},10^{-2}\}$, and the classifier is then refit on all 50,000 images.
We report top-1 accuracy on the full validation split.
The \emph{upstream} probe (\cref{tab:appx:upstream-imagenet}) fits the same classifier to a frozen pretrained encoder before multimodal training.
It mean-pools the encoder's 256 final normalized patch tokens of a $224\times224$ center crop; for WebSSL, it reads the exponential-moving-average teacher.
The \emph{full-data} probe applies the DINOv2 linear evaluation~\citep{oquab2024dinov2} to the frozen pretrained encoder, training on the full ImageNet-1K training set.
It gives the upstream accuracy in \cref{fig:practical:free-pretraining} and \cref{sec:appx:upstream-fixed}.

  \FloatBarrier

\section{Separate encoder pretraining}
\label{sec:appx:prior}

This section details the pretrained-encoder comparisons of \cref{sec:encoders} and the sweep behind the zero-cost pretrained-encoder comparison of \cref{sec:practical:free}.

\subsection{Objectives and upstream training}
\label{sec:appx:encoder-objectives}

We compare four upstream objectives on the same vision encoder architecture (\cref{sec:appx:architecture}).
MetaCLIP~\citep{xu2023metaclip} uses a symmetric image--text contrastive objective~\citep{radford2021clip}.
If $z_i$ and $w_i$ are normalized image and text representations in a batch of size $B$, its loss is
\begin{equation}
  \mathcal{L}_{\mathrm{clip}}
  = -\frac{1}{2B}\sum_{i=1}^{B}
  \left[
    \log\frac{\exp(z_i^\top w_i/\tau)}
                   {\sum_j\exp(z_i^\top w_j/\tau)}
    +
    \log\frac{\exp(w_i^\top z_i/\tau)}
                   {\sum_j\exp(w_i^\top z_j/\tau)}
  \right].
  \label{eq:appx:metaclip}
\end{equation}
Here $\tau$ is a learned softmax temperature, initialized at 0.07 and kept at or above 0.01.
Cap~\citep{tschannen2023captioners} uses a causal decoder during upstream pretraining to predict caption tokens from the encoder output $f_{\phi}(x)$ (\cref{eq:appx:encoder-path}),
\begin{equation}
  \mathcal{L}_{\mathrm{cap}}
  = -\mathbb{E}_{(x,y)}
    \left[\frac{1}{|y|}\sum_j
      \log p(y_j\mid f_{\phi}(x),y_{<j})\right].
  \label{eq:appx:cap}
\end{equation}
CapPa~\citep{tschannen2023captioners} combines this causal objective with fully masked parallel caption prediction.
If $\mathcal{L}_{\mathrm{par}}$ is the loss that predicts all caption positions in parallel from the image and a fully masked caption input, CapPa's expected loss is
\begin{equation}
  \mathcal{L}_{\mathrm{cappa}}
  = 0.25\,\mathcal{L}_{\mathrm{cap}}
  + 0.75\,\mathcal{L}_{\mathrm{par}}.
  \label{eq:appx:cappa}
\end{equation}
WebSSL~\citep{fan2025webssl} follows DINOv2~\citep{oquab2024dinov2} and combines class-token self-distillation~\citep{caron2021dino} and KoLeo regularization~\citep{sablayrolles2019koleo} with iBOT masked-patch prediction~\citep{zhou2022ibot},
\begin{equation}
  \mathcal{L}_{\mathrm{webssl}}
  = \lambda_{\mathrm{DINO}}\mathcal{L}_{\mathrm{DINO}}
  + \lambda_{\mathrm{iBOT}}\mathcal{L}_{\mathrm{iBOT}}
  + \lambda_{\mathrm{KoLeo}}\mathcal{L}_{\mathrm{KoLeo}}.
  \label{eq:appx:webssl}
\end{equation}
We use the DINOv2 weights $\lambda_{\mathrm{DINO}}=\lambda_{\mathrm{iBOT}}=1$ and $\lambda_{\mathrm{KoLeo}}=0.1$.
The DINO term sums over the global and local crops, and the KoLeo term over the two global crops.
We discard the text tower, caption decoder, and self-supervised prediction heads before multimodal training.
We follow the original objective definitions.
MetaCLIP, Cap, and CapPa train on the original alt-text of the MetaCLIP 1.2 image--text pairs, and WebSSL trains on their images alone.
Sharing one data source isolates the objective, since the pretraining data alone changes the learned representations~\citep{li2023internet,fan2025webssl}.
\Cref{tab:appx:upstream-training} lists the upstream training settings.
MetaCLIP's text tower is a 12-layer Transformer of width 768 with 12 attention heads over 77 text tokens.
The Cap and CapPa decoder follows the published CapPa decoder: 12 layers of width 1,024 with 16 attention heads, a 4,096-wide MLP, and cross-attention to the encoder's patch tokens in every layer, over captions of up to 64 tokens.
\begin{table}[!htbp]
  \centering
  \small
  \caption{
    \textbf{Every objective pretrains the same encoder on the same data and compute.}
    The training schedules span the entire horizon of each encoder pretraining run.
  }
  \label{tab:appx:upstream-training}
  \setlength{\tabcolsep}{5pt}
  \begin{tabular}{@{}lccc@{}}
    \toprule
    & MetaCLIP & WebSSL & Cap and CapPa \\
    \midrule
    Global batch (examples) & 32,768 & 3,072 & 8,192 \\
    Optimizer & Muon, AdamW & Muon, AdamW & Muon, AdamW \\
    Peak learning rate & $4\times10^{-4}$ & $3.5\times10^{-4}$ & $2.5\times10^{-4}$ \\
    Warmup updates & 2,000 & 100,000 & 10,000 \\
    Schedule & cosine to 0 & cosine to $10^{-6}$ & cosine to 0 \\
    Weight decay & 0.2 & 0.04 to 0.2 (cosine) & $10^{-4}$ \\
    Gradient-norm clip & none & 3.0 & 1.0 \\
    Image crops & \makecell{one $224\times224$\\random resized crop} & \makecell{two $224\times224$ global,\\eight $98\times98$ local} & \makecell{one $224\times224$\\random resized crop} \\
    \bottomrule
  \end{tabular}
\end{table}

We compare the objectives at matched upstream compute.
Each method uses the first checkpoint that reaches the compute of two epochs of the Cap objective over the image--text pairs ($1.54\times10^{21}$ FLOPs).
For Cap and CapPa, this is 2 billion image--caption examples.
MetaCLIP and WebSSL reach the same compute point after different numbers of examples.
\Cref{sec:appx:objectives} uses these checkpoints at both backbone sizes.
\Cref{sec:appx:objectives} also uses the first checkpoints that reach the compute of one, four, and eight Cap epochs ($0.77$, $3.08$, and $6.16\times10^{21}$~FLOPs), the 7.4\%, 30\%, and 60\% shares of \cref{fig:encoders:upstream:a}.
The sweep of \cref{sec:appx:upstream-fixed} compares MetaCLIP only with itself, so it uses MetaCLIP's own epoch checkpoints instead (\cref{sec:appx:lifetime-flops}).

The ImageNet columns of \cref{tab:appx:upstream-imagenet} give the probe accuracy of each upstream encoder at this matched compute, before multimodal training.
MetaCLIP has the highest and WebSSL the lowest top-1 accuracy, yet after multimodal training WebSSL reaches the highest Avg.\ VQA of the four (\cref{sec:appx:objectives}), so their ImageNet recognition ranking does not carry over to Avg.\ VQA.

\FloatBarrier

\subsection{Comparison of encoder objectives}
\label{sec:appx:objectives}

At 1.7B, \cref{tab:encoders:objectives} compares the from-scratch encoder with encoders pretrained on each of the four objectives and updated during multimodal training, and with a frozen MetaCLIP encoder.
The updated pretrained rows are the points at a 15\% pretraining share in \cref{fig:encoders:upstream:a}.
The from-scratch encoder and WebSSL lead on Avg.\ VQA, with the other objectives 0.7 to 1.8 points behind, and all updated encoders are comparable on Avg.\ Text.
MetaCLIP serves as the pretrained reference elsewhere because CLIP-style pretraining is the most widely used (\cref{sec:encoders}), and the from-scratch encoder also matches WebSSL, the strongest objective here.
No objective leads every category, as self-supervised WebSSL is best on general and vision-centric questions and the from-scratch encoder on knowledge and on OCR and chart questions.
Every pretrained encoder trails the from-scratch encoder on OCR and chart questions, and only WebSSL and Cap gain from their priors, on vision-centric questions.
Freezing MetaCLIP gives slightly higher Avg.\ Text but noticeably lower Avg.\ VQA, so an encoder that adapts during multimodal training is the stronger choice.

\paragraph{At 4B.}
We repeat the comparison on the 4B backbone with three of the objectives, at the $5.8\times10^{21}$-FLOP budget of \cref{fig:encoders:upstream:b} (61 TPP), whose from-scratch and MetaCLIP points are the 4B rows of \cref{tab:encoders:objectives}.
WebSSL again scores highest among the objectives and MetaCLIP lowest, on both averages.
The from-scratch encoder leads all three on Avg.\ VQA and Avg.\ Text, and on every VQA category.

\begin{table}[!htbp]
  \centering
  \small
  \caption{
    \textbf{The from-scratch encoder matches the best pretrained encoders at 1.7B and leads every objective at 4B.}
    The models are trained to $1.04\times10^{22}$~FLOPs at 1.7B and $5.8\times10^{21}$~FLOPs at 4B, including encoder pretraining.
    ImageNet accuracy is measured before multimodal training.
    Bold: within one point of the best at each backbone size.
  }
  \label{tab:encoders:objectives}
  \label{tab:appx:pretraining-objectives-larger-backbone} %
  \label{tab:appx:upstream-imagenet} %
  
\setlength{\tabcolsep}{4.0pt}
\renewcommand{\arraystretch}{1.05}
\begin{tabular}{@{}llrrrrrrrrrrrr@{}}
  \toprule
  \multicolumn{2}{>{\columncolor{PaperGray100!55}[.5\tabcolsep][.5\tabcolsep]}c}{\textbf{Encoder recipe}}
    & \multicolumn{5}{>{\columncolor{PaperBlue100!55}[.5\tabcolsep][.5\tabcolsep]}c}{\textbf{VQA (\%) $\uparrow$}}
    & \multicolumn{5}{>{\columncolor{PaperGreen100!55}[.5\tabcolsep][.5\tabcolsep]}c}{\textbf{Text (\%) $\uparrow$}}
    & \multicolumn{2}{>{\columncolor{PaperGray100!55}[.5\tabcolsep][.5\tabcolsep]}c}{\textbf{ImageNet (\%) $\uparrow$}} \\
  \cmidrule(l{.5\tabcolsep}r{.5\tabcolsep}){1-2}\cmidrule(l{.5\tabcolsep}r{.5\tabcolsep}){3-7}\cmidrule(l{.5\tabcolsep}r{.5\tabcolsep}){8-12}\cmidrule(l{.5\tabcolsep}r{.5\tabcolsep}){13-14}
  Backbone & Variant & \multicolumn{1}{c}{Avg} & \multicolumn{1}{c}{Gen} & \multicolumn{1}{c}{Knw} & \multicolumn{1}{c}{OCR} & \multicolumn{1}{c}{Vis} & \multicolumn{1}{c}{Avg} & \multicolumn{1}{c}{Com} & \multicolumn{1}{c}{Sci} & \multicolumn{1}{c}{Knw} & \multicolumn{1}{c}{Rdg} & \multicolumn{1}{c}{Top-1} & \multicolumn{1}{c@{}}{Top-5} \\
  \midrule
  \rowcolor{PaperGray100!50}
  1.7B & \recipeglyph{scratch}Scratch encoder & \textbf{49.0} & \textbf{60.3} & \textbf{41.4} & \textbf{43.4} & 51.1 & \textbf{54.3} & \textbf{66.1} & \textbf{47.6} & 47.3 & 56.3 & -- & -- \\
   & \recipeglyph{metaclip}MetaCLIP & 47.5 & 59.4 & 40.3 & 40.3 & 50.2 & \textbf{54.7} & \textbf{65.9} & \textbf{48.3} & \textbf{48.0} & 56.5 & \textbf{73.0} & \textbf{93.1} \\
   & \recipeglyph{}MetaCLIP (frozen) & 45.5 & 56.5 & 40.0 & 36.6 & 48.8 & \textbf{55.1} & \textbf{66.4} & \textbf{48.0} & \textbf{47.5} & \textbf{58.7} & \textbf{73.0} & \textbf{93.1} \\
   & \recipeglyph{webssl}WebSSL & \textbf{49.3} & \textbf{60.6} & \textbf{41.3} & 39.8 & \textbf{55.5} & \textbf{54.6} & \textbf{66.1} & 46.5 & \textbf{48.5} & 57.3 & 61.5 & 86.1 \\
   & \recipeglyph{cap}Cap & \textbf{48.3} & 58.6 & \textbf{41.2} & 39.8 & 53.3 & \textbf{54.4} & \textbf{66.1} & \textbf{47.7} & 46.2 & 57.5 & \textbf{72.0} & 91.6 \\
   & \recipeglyph{cappa}CapPa & 47.2 & 58.7 & 39.8 & 40.1 & 50.1 & \textbf{54.3} & \textbf{66.5} & \textbf{47.5} & 46.3 & 57.0 & 69.6 & 90.2 \\
  \midrule
  \rowcolor{PaperGray100!50}
  4B & \recipeglyph{scratch}Scratch encoder & \textbf{51.2} & \textbf{62.7} & \textbf{44.5} & \textbf{43.5} & \textbf{54.0} & \textbf{56.4} & \textbf{67.5} & \textbf{49.8} & \textbf{49.4} & \textbf{58.8} & -- & -- \\
   & \recipeglyph{metaclip}MetaCLIP & 46.5 & 58.4 & 39.2 & 40.3 & 48.1 & 53.2 & 66.1 & 47.8 & 43.6 & 55.5 & \textbf{73.0} & \textbf{93.1} \\
   & \recipeglyph{webssl}WebSSL & 48.1 & 60.8 & 40.2 & 41.5 & 49.8 & 54.4 & \textbf{66.5} & 48.4 & 45.4 & 57.4 & 61.5 & 86.1 \\
   & \recipeglyph{cap}Cap & 46.6 & 58.3 & 40.3 & 39.2 & 48.4 & 53.7 & 66.1 & 48.5 & 44.6 & 55.8 & \textbf{72.0} & 91.6 \\
  \bottomrule
\end{tabular}

\end{table}

\paragraph{Training trajectories.}
\Cref{fig:architecture:pretraining-objectives} follows the from-scratch, updated MetaCLIP, and frozen MetaCLIP encoders through training at the 15\% share.
Every checkpoint before the last precedes the final learning-rate decay (\cref{sec:appx:optimization}), so only the final points match \cref{tab:encoders:objectives}.

\begin{figure}[!htbp]
  \centering
  \includegraphics[width=\linewidth]{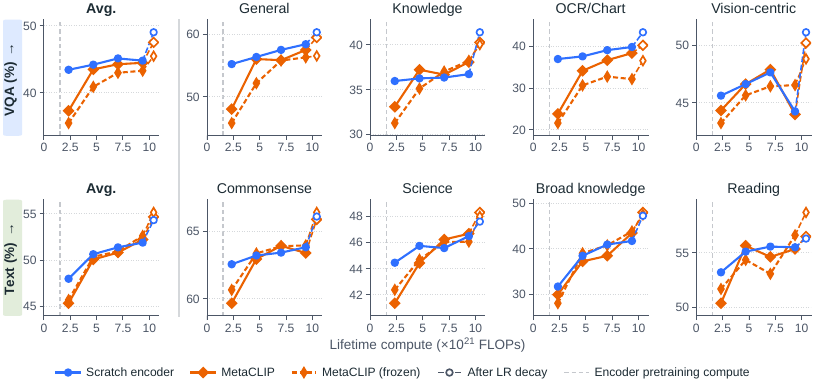}
  \caption{\textbf{Joint training with the scratch encoder is comparable to a pretrained encoder.}
  All models use a 1.7B backbone, and the curves start after the encoder's pretraining compute (dashed line).
  }
  \label{fig:architecture:pretraining-objectives}
\end{figure}

\paragraph{Compositional matching.}
SugarCrepe pairs each image with a correct caption and a hard negative that replaces, swaps, or adds an object, attribute, or relation~\citep{hsieh2023sugarcrepe}.
We score the multimodal-training checkpoints before instruction tuning by caption likelihood on over 7,511 items in seven subsets.
An item is correct when the correct caption's summed token loss is lower than the negative's loss.
A control scores the same items with an unrelated image shared by both captions.
Within each budget, every recipe reaches a comparable replace-and-swap mean, within about 1.5 points, and differs most on the small object-swap subset (\cref{tab:appx:sugarcrepe}).
Unlike VQA (\cref{sec:encoder-free}), this readout separates neither encoder-free from the encoders nor pretrained from from-scratch encoders.

\begin{table}[!htbp]
  \centering
  \small
  \caption{
    \textbf{All recipes are comparable on compositional caption matching.}
    Accuracy (\%) is measured before instruction tuning.
    The last two columns average the replace and swap subsets, leaving out the add subsets because the text alone decides them.
    Bold: within one point of the best in each block, except the shuffled control.
  }
  \label{tab:appx:sugarcrepe}
  
\renewcommand{\arraystretch}{1.05}
\setlength{\tabcolsep}{4.5pt}
\begin{tabular}{@{}lrrrrrrrrr@{}}
  \toprule
  & \multicolumn{3}{c}{Replace} & \multicolumn{2}{c}{Swap} & \multicolumn{2}{c}{Add} & \multicolumn{2}{c}{Replace and swap} \\
  \cmidrule(l{.5\tabcolsep}r{.5\tabcolsep}){2-4}\cmidrule(l{.5\tabcolsep}r{.5\tabcolsep}){5-6}\cmidrule(l{.5\tabcolsep}r{.5\tabcolsep}){7-8}\cmidrule(l{.5\tabcolsep}r{.5\tabcolsep}){9-10}
  Model & \multicolumn{1}{c}{Object} & \multicolumn{1}{c}{Attribute} & \multicolumn{1}{c}{Relation} & \multicolumn{1}{c}{Object} & \multicolumn{1}{c}{Attribute} & \multicolumn{1}{c}{Object} & \multicolumn{1}{c}{Attribute} & \multicolumn{1}{c}{Image} & \multicolumn{1}{c@{}}{Shuffled} \\
  \midrule
  \multicolumn{10}{@{}l}{\textit{1.7B, $1.04\times10^{22}$ FLOPs}} \\
  Scratch encoder & \textbf{79.2} & \textbf{82.0} & 77.0 & 71.4 & 84.8 & \textbf{95.5} & \textbf{96.5} & \textbf{78.9} & 64.7 \\
  Encoder-free & \textbf{79.7} & \textbf{82.2} & 77.3 & 72.2 & \textbf{86.2} & \textbf{95.9} & \textbf{97.4} & \textbf{79.5} & 66.8 \\
  MetaCLIP & 78.2 & 80.6 & 77.4 & 73.5 & 84.2 & \textbf{96.1} & \textbf{96.7} & 78.8 & 64.7 \\
  WebSSL & \textbf{78.8} & \textbf{81.2} & \textbf{79.2} & 73.9 & \textbf{86.6} & \textbf{96.0} & \textbf{96.4} & \textbf{79.9} & 65.1 \\
  Cap & 78.3 & 81.1 & 76.6 & 71.0 & 85.1 & \textbf{95.5} & 96.2 & 78.4 & 64.3 \\
  CapPa & 76.6 & 80.3 & 77.8 & \textbf{75.1} & 84.4 & \textbf{95.8} & 96.0 & \textbf{78.9} & 64.4 \\
  \arrayrulecolor{PaperGray400}\cmidrule{1-10}\arrayrulecolor{black}
  \multicolumn{10}{@{}l}{\textit{1.7B, $1.30\times10^{22}$ FLOPs}} \\
  Scratch encoder & 77.6 & 80.6 & \textbf{78.9} & \textbf{74.7} & \textbf{84.7} & \textbf{95.4} & \textbf{97.1} & \textbf{79.3} & 65.3 \\
  Encoder-free & \textbf{79.4} & \textbf{82.7} & 76.5 & 71.4 & \textbf{85.4} & \textbf{95.5} & \textbf{97.3} & \textbf{79.1} & 65.9 \\
  In-fusion & 78.1 & 80.1 & 77.7 & 68.6 & \textbf{85.0} & \textbf{95.4} & \textbf{96.7} & 77.9 & 64.9 \\
  \arrayrulecolor{PaperGray400}\cmidrule{1-10}\arrayrulecolor{black}
  \multicolumn{10}{@{}l}{\textit{4B, $1.16\times10^{22}$ FLOPs}} \\
  Scratch encoder & \textbf{79.5} & \textbf{81.2} & \textbf{78.2} & \textbf{72.2} & \textbf{85.4} & \textbf{95.6} & \textbf{97.4} & \textbf{79.3} & 64.3 \\
  MetaCLIP & 77.3 & 79.3 & \textbf{78.4} & 69.8 & \textbf{86.3} & \textbf{95.4} & \textbf{96.5} & 78.2 & 65.1 \\
  \bottomrule
\end{tabular}

\end{table}

\subsection{Encoder pretraining at fixed multimodal compute}
\label{sec:appx:upstream-fixed}

\Cref{fig:practical:free-pretraining} plots a sweep of experiments that holds multimodal training fixed and does not charge encoder pretraining, unlike the lifetime-budget comparisons of \cref{fig:encoders:upstream}.
Every model starts from a MetaCLIP checkpoint taken after 2, 4, 8, or 13 epochs over the image--text pairs (\cref{sec:appx:lifetime-flops}).
Every model then receives the same multimodal training: $2.7\times10^{21}$~FLOPs at 1.7B with the 70:30 mixture.
Longer pretraining raises the pretrained encoder's full-data ImageNet linear-probe accuracy (\cref{sec:appx:probe}) but does not improve Avg.\ VQA after multimodal training.
The shortest pretraining scores highest on Avg.\ VQA (\cref{tab:appx:upstream-fixed}).
A lower encoder learning rate also does not recover a benefit from the encoder pretrained prior.
Updating the shortest-pretrained MetaCLIP encoder at a tenth of the shared learning rate gives about the same Avg.\ VQA and Avg.\ Text as the full learning rate, and a from-scratch encoder with the same multimodal training matches both (\cref{tab:appx:upstream-fixed}).

\begin{table}[!htbp]
  \centering
  \small
  \caption{
    \textbf{At fixed multimodal compute, longer encoder pretraining improves ImageNet accuracy but not visual understanding.}
    All rows use a 1.7B backbone and $2.7\times10^{21}$~FLOPs of multimodal training; lifetime FLOPs add the encoder pretraining.
    ImageNet LP is a linear probe of the pretrained encoder before multimodal training.
    Bold: within one point of the best in each VQA and Text column.
  }
  \label{tab:appx:upstream-fixed}
  
\setlength{\tabcolsep}{2.6pt}
\renewcommand{\arraystretch}{1.05}
\begin{tabular}{@{}lrrrrrrrrrrrrr@{}}
  \toprule
  \multicolumn{4}{>{\columncolor{PaperGray100!55}[.5\tabcolsep][.5\tabcolsep]}c}{\textbf{Setup}}
    & \multicolumn{5}{>{\columncolor{PaperBlue100!55}[.5\tabcolsep][.5\tabcolsep]}c}{\textbf{VQA (\%) $\uparrow$}}
    & \multicolumn{5}{>{\columncolor{PaperGreen100!55}[.5\tabcolsep][.5\tabcolsep]}c}{\textbf{Text (\%) $\uparrow$}} \\
  \cmidrule(l{.5\tabcolsep}r{.5\tabcolsep}){1-4}\cmidrule(l{.5\tabcolsep}r{.5\tabcolsep}){5-9}\cmidrule(l{.5\tabcolsep}r{.5\tabcolsep}){10-14}
  Encoder & \makecell{Pretraining\\FLOPs} & \makecell{Lifetime\\FLOPs} & \makecell{ImageNet\\LP} & \multicolumn{1}{c}{Avg} & \multicolumn{1}{c}{Gen} & \multicolumn{1}{c}{Knw} & \multicolumn{1}{c}{OCR} & \multicolumn{1}{c}{Vis} & \multicolumn{1}{c}{Avg} & \multicolumn{1}{c}{Com} & \multicolumn{1}{c}{Sci} & \multicolumn{1}{c}{Knw} & \multicolumn{1}{c@{}}{Rdg} \\
  \midrule
  \recipeglyph{metaclip}MetaCLIP & $1.4{\times}10^{21}$ & $4.1{\times}10^{21}$ & 80.4 & \textbf{43.2} & 55.2 & 34.8 & \textbf{36.6} & 46.0 & \textbf{50.5} & \textbf{63.9} & \textbf{45.2} & \textbf{39.9} & 52.9 \\
   & $2.8{\times}10^{21}$ & $5.5{\times}10^{21}$ & 80.6 & 42.2 & 52.9 & 34.4 & \textbf{36.2} & 45.4 & \textbf{51.0} & \textbf{64.7} & \textbf{45.4} & 38.6 & \textbf{55.4} \\
   & $5.5{\times}10^{21}$ & $8.3{\times}10^{21}$ & 82.0 & \textbf{42.7} & 53.9 & 35.8 & 35.6 & 45.5 & \textbf{51.0} & \textbf{64.5} & \textbf{45.3} & \textbf{39.2} & \textbf{54.9} \\
   & $9.0{\times}10^{21}$ & $1.17{\times}10^{22}$ & 84.4 & 41.6 & 53.0 & 34.9 & 35.0 & 43.6 & \textbf{50.7} & \textbf{64.2} & \textbf{45.4} & 37.9 & \textbf{55.5} \\
    \arrayrulecolor{PaperGray400}\cmidrule{1-14}\arrayrulecolor{black}
  \recipeglyph{metaclip}MetaCLIP, 0.1$\times$ LR & $1.4{\times}10^{21}$ & $4.1{\times}10^{21}$ & 80.4 & \textbf{43.1} & 53.9 & \textbf{37.1} & 32.9 & \textbf{48.4} & \textbf{50.8} & \textbf{63.9} & \textbf{45.3} & \textbf{39.4} & \textbf{54.6} \\
    \arrayrulecolor{PaperGray400}\cmidrule{1-14}\arrayrulecolor{black}
  \recipeglyph{scratch}Scratch encoder & -- & $2.7{\times}10^{21}$ & -- & \textbf{43.5} & \textbf{56.3} & 35.5 & \textbf{37.2} & 45.0 & \textbf{50.6} & \textbf{64.3} & \textbf{46.2} & 36.3 & \textbf{55.5} \\
  \bottomrule
\end{tabular}

\end{table}

\section{Dedicated vision capacity}
\label{sec:appx:dedicated-capacity}

This section describes per-category breakdowns of evaluation metrics and experimental details for \cref{sec:encoder-free}.

\subsection{Encoder advantage by VQA category}
\label{sec:appx:encoder-free-categories}
\label{sec:appx:scaling:deficit} %

\Cref{fig:scaling:ef-categories} splits the encoder-free deficit compared to the from-scratch encoder by VQA category.
The deficit is largest on OCR and chart questions at every budget at both 1.7B and 4B backbone sizes.
Avg.\ Text stays within one point (\cref{fig:encoder-free:gap}).
\Cref{sec:appx:per-benchmark} gives every VQA and text benchmark for the runs of each main-text exhibit.

\begin{figure}[!htbp]
  \centering
  \includegraphics[width=\linewidth]{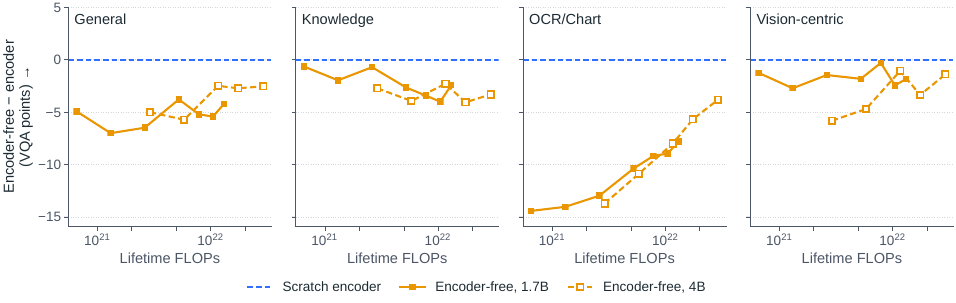}
  \caption{
    \textbf{The encoder's advantage over encoder-free is primarily in OCR and chart questions.}
    Each point compares encoder-free with the encoder at one budget of \cref{tab:appx:ladder}.
  }
  \label{fig:scaling:ef-categories}
\end{figure}

\subsection{Encoder-free with increased shared capacity}
\label{sec:appx:shared-capacity-dose}

\paragraph{Parameter-matched encoder-free.}
The parameter-matched encoder-free baselines in \cref{tab:encoder-free:steelman} give the encoder-free backbone 0.4B extra parameters.
The default 1.7B backbone has 28 layers of hidden size 2,048, the wider backbone has hidden size 2,304, the deeper backbone has 36 layers, and the wider and deeper backbone has 30 layers of hidden size 2,176, each with about 2.1B parameters.
\Cref{tab:appx:encoder-free:4b} shows how the wider variant does at 4B, using 122-TPP runs of \cref{tab:appx:ladder}.
Compared to the default 4B backbone with 36 layers of hidden size 2,560, the wider backbone has hidden size 2,688 and 4.4B parameters.
Here, wider means a larger hidden size, whereas the fixed-token study in \cref{tab:appx:shared-capacity-dose} widens only the FFNs and keeps the hidden size at 2,048.

\begin{table}[h]
  \centering
  \footnotesize
  \caption{
    \textbf{At 4B, a wider encoder-free backbone still trails the encoder on VQA.}
    All variants are trained to 122 TPP, i.e., $1.16\times10^{22}$~FLOPs.
    Bold: within one point of the best.
  }
  \label{tab:appx:encoder-free:4b}
  
\setlength{\tabcolsep}{3.0pt}
\renewcommand{\arraystretch}{1.05}
\begin{tabular}{@{}lrrrrrrrrrrrr@{}}
  \toprule
  \multicolumn{1}{>{\columncolor{PaperGray100!55}[.5\tabcolsep][.5\tabcolsep]}c}{\textbf{Architecture}}
    & \multicolumn{2}{>{\columncolor{PaperGray100!55}[.5\tabcolsep][.5\tabcolsep]}c}{\textbf{Added params (B)}}
    & \multicolumn{5}{>{\columncolor{PaperBlue100!55}[.5\tabcolsep][.5\tabcolsep]}c}{\textbf{VQA (\%) $\uparrow$}}
    & \multicolumn{5}{>{\columncolor{PaperGreen100!55}[.5\tabcolsep][.5\tabcolsep]}c}{\textbf{Text (\%) $\uparrow$}} \\
  \cmidrule(l{.5\tabcolsep}r{.5\tabcolsep}){1-1}\cmidrule(l{.5\tabcolsep}r{.5\tabcolsep}){2-3}\cmidrule(l{.5\tabcolsep}r{.5\tabcolsep}){4-8}\cmidrule(l{.5\tabcolsep}r{.5\tabcolsep}){9-13}
  Variant & Vision-only & Shared & \multicolumn{1}{c}{Avg} & \multicolumn{1}{c}{Gen} & \multicolumn{1}{c}{Knw} & \multicolumn{1}{c}{OCR} & \multicolumn{1}{c}{Vis} & \multicolumn{1}{c}{Avg} & \multicolumn{1}{c}{Com} & \multicolumn{1}{c}{Sci} & \multicolumn{1}{c}{Knw} & \multicolumn{1}{c}{Rdg} \\
  \midrule
  \recipeglyph{scratch}Scratch encoder & 0.4 & -- & \textbf{51.9} & \textbf{63.0} & \textbf{46.0} & \textbf{45.5} & \textbf{53.1} & \textbf{57.9} & \textbf{68.6} & \textbf{50.5} & \textbf{52.3} & \textbf{60.2} \\
  \recipeglyph{encoderfree}Encoder-free & -- & -- & 48.4 & 60.5 & 43.8 & 37.5 & 52.0 & 57.0 & \textbf{68.3} & 50.0 & \textbf{52.3} & 57.3 \\
  \recipeglyph{}\hspace{0.7em}+ wider & -- & 0.4 & 48.1 & 59.7 & 43.4 & 39.1 & 50.2 & \textbf{58.3} & \textbf{68.7} & \textbf{51.2} & \textbf{52.3} & \textbf{61.0} \\
  \bottomrule
\end{tabular}

\end{table}

\paragraph{Parameter-matched encoder-free at a fixed token count.}
\Cref{tab:encoder-free:steelman} matches lifetime compute, so each larger encoder-free variant trains on fewer tokens than the encoder.
We therefore also train 1.7B encoder-free backbones with wider FFNs or more layers on the encoder's token count, which gives each encoder-free variant more compute than the encoder (\cref{tab:appx:shared-capacity-dose}).
We evaluate these runs at a constant learning rate, without a final decay, so we compare rows here only within \cref{tab:appx:shared-capacity-dose}.
At approximately the encoder's parameter count, added FFN width or depth still leaves the backbone 2--3 Avg.\ VQA points behind the encoder.
Wider or deeper variants both narrow the gap, but wider FFNs need about twice the encoder's parameters to match it, and the deeper backbone still trails it slightly.
Every wider or deeper variant scores higher than the encoder on Avg.\ Text, consistent with how the shared parameters can also be used by language.

\begin{table}[!htbp]
  \centering
  \small
  \caption{
    \textbf{Shared capacity needs about twice the encoder's parameters to match it on VQA.}
    All variants use a 1.7B backbone, train on the encoder's token count, and use a constant learning rate.
  }
  \label{tab:appx:shared-capacity-dose}
  
\setlength{\tabcolsep}{3.6pt}
\renewcommand{\arraystretch}{1.05}
\begin{tabular}{@{}lrrrrrrr@{}}
  \toprule
  \multicolumn{3}{>{\columncolor{PaperGray100!55}[.5\tabcolsep][.5\tabcolsep]}c}{\textbf{Architecture}}
    & \multicolumn{2}{>{\columncolor{PaperGray100!55}[.5\tabcolsep][.5\tabcolsep]}c}{\textbf{Added params (B)}}
    & \multicolumn{1}{>{\columncolor{PaperGray100!55}[.5\tabcolsep][.5\tabcolsep]}c}{\textbf{Compute}}
    & \multicolumn{1}{>{\columncolor{PaperBlue100!55}[.5\tabcolsep][.5\tabcolsep]}c}{\textbf{VQA (\%) $\uparrow$}}
    & \multicolumn{1}{>{\columncolor{PaperGreen100!55}[.5\tabcolsep][.5\tabcolsep]}c}{\textbf{Text (\%) $\uparrow$}} \\
  \cmidrule(l{.5\tabcolsep}r{.5\tabcolsep}){1-3}\cmidrule(l{.5\tabcolsep}r{.5\tabcolsep}){4-5}\cmidrule(l{.5\tabcolsep}r{.5\tabcolsep}){6-6}\cmidrule(l{.5\tabcolsep}r{.5\tabcolsep}){7-7}\cmidrule(l{.5\tabcolsep}r{.5\tabcolsep}){8-8}
  Variant & Layers & FFN width & Vision-only & Shared & FLOPs & \multicolumn{1}{c}{Avg} & \multicolumn{1}{c@{}}{Avg} \\
  \midrule
  \recipeglyph{scratch}Scratch encoder & 28 & 6,144 & 0.4 & -- & $4.7{\times}10^{21}$ & 44.2 & 50.6 \\
  \recipeglyph{encoderfree}Encoder-free & 28 & 6,144 & -- & -- & $4.7{\times}10^{21}$ & 40.2 & 49.8 \\
  \recipeglyph{}\hspace{0.7em}+ wider FFN & 28 & 9,024 & -- & 0.5 & $5.6{\times}10^{21}$ & 41.1 & 51.5 \\
  \recipeglyph{}\hspace{0.7em}+ wider FFN & 28 & 10,752 & -- & 0.8 & $6.3{\times}10^{21}$ & 44.4 & 51.7 \\
  \recipeglyph{}\hspace{0.7em}+ wider FFN & 28 & 11,904 & -- & 1.0 & $6.7{\times}10^{21}$ & 44.6 & 53.3 \\
  \recipeglyph{}\hspace{0.7em}+ deeper & 36 & 6,144 & -- & 0.4 & $5.6{\times}10^{21}$ & 41.8 & 52.8 \\
  \recipeglyph{}\hspace{0.7em}+ deeper & 44 & 6,144 & -- & 0.8 & $6.6{\times}10^{21}$ & 43.6 & 54.7 \\
  \bottomrule
\end{tabular}

\end{table}

\subsection{Richer patch embeddings}
\label{sec:appx:encoder-free-tokenization}

The default encoder-free patch embedding is a single linear projection followed by RMSNorm and a learned gate (\cref{sec:appx:architecture}).
We compare it with two richer patch embedding designs: a normalized two-layer projection inspired by Gemma~4~\citep{gemmateam2026gemma4}, and a four-layer hierarchical MLP~\citep{touvron2022three} inspired by Inkling~\citep{thinkingmachines2026inkling}.
The lower part of \cref{tab:appx:patch-embedding-specs} describes each variant.
The normalized two-layer projection changes patch size, resolution, position embeddings, and normalization together.
It normalizes each patch with LayerNorm~\citep{ba2016layernorm}, projects it linearly, adds learned factorized 2D position embeddings, and then applies LayerNorm, a parameter-free RMSNorm, and a linear map into the backbone.
The hierarchical MLP has four linear layers that fold neighboring pixels into channels in stages, with RMSNorm and GELU~\citep{hendrycks2016gelu} between them, a final RMSNorm, and no position embedding.
In \cref{tab:architecture:encoder-free}, both richer variants read larger images in larger patches ($768\times768$ in $48\times48$ patches and $640\times640$ in $40\times40$ patches, rather than $224\times224$ in $14\times14$ patches), so they produce the same $16\times16$ grid of 256 vision tokens as the encoder.
At $5.2\times10^{21}$~FLOPs, both narrow the gap to the encoder, and the hierarchical MLP matches it on Avg.\ VQA (\cref{tab:architecture:encoder-free}).
The largest gain is on OCR and chart questions, where the default patch embedding trails the encoder most.
Avg.\ Text is comparable across patch embeddings.

However, when training on $224\times224$ images at $2.6\times10^{21}$~FLOPs, the two-layer projection gains 1.5 Avg.\ VQA over the default patch embedding whereas the hierarchical MLP loses 3.1, and both remain far behind the encoder (\cref{tab:encoder-free:steelman}).
The hierarchical MLP helps only when it compresses a higher-resolution image into the same number of tokens.
With $224\times224$ instruction tuning, it matches the default patch embedding on $224\times224$ images, while processing $640\times640$ images into the same 256 tokens raises it to 38.8, mostly on OCR and chart questions (24.0 against 10.2).
Raising only the instruction-tuning resolution to $448\times448$ quadruples the tokens without compressing them, and the hierarchical MLP then gains 1.4 points, against 4.4 for the default patch embedding.
The richer patch embeddings therefore help most with higher-resolution images.

\begin{table}[!htbp]
  \centering
  \small
  \caption{
    \textbf{Richer patch embeddings narrow the gap to the encoder only when processing higher-resolution images.}
    All variants train a 1.7B backbone at 315 TPP to $5.2\times10^{21}$~FLOPs.
    VQA uses $224\times224$ instruction tuning.
    Bold: within one point of the best.
    Each architecture variant is described in the table below.
  }
  \label{tab:architecture:encoder-free}
  \label{tab:appx:patch-embedding-specs} %
  
\setlength{\tabcolsep}{3.0pt}
\renewcommand{\arraystretch}{1.05}
\begin{tabular}{@{}lrrrrrrrrrrrr@{}}
  \toprule
  \multicolumn{1}{>{\columncolor{PaperGray100!55}[.5\tabcolsep][.5\tabcolsep]}c}{\textbf{Architecture}}
    & \multicolumn{2}{>{\columncolor{PaperGray100!55}[.5\tabcolsep][.5\tabcolsep]}c}{\textbf{Params}}
    & \multicolumn{5}{>{\columncolor{PaperBlue100!55}[.5\tabcolsep][.5\tabcolsep]}c}{\textbf{VQA (\%) $\uparrow$}}
    & \multicolumn{5}{>{\columncolor{PaperGreen100!55}[.5\tabcolsep][.5\tabcolsep]}c}{\textbf{Text (\%) $\uparrow$}} \\
  \cmidrule(l{.5\tabcolsep}r{.5\tabcolsep}){1-1}\cmidrule(l{.5\tabcolsep}r{.5\tabcolsep}){2-3}\cmidrule(l{.5\tabcolsep}r{.5\tabcolsep}){4-8}\cmidrule(l{.5\tabcolsep}r{.5\tabcolsep}){9-13}
  Variant
    & \makecell{Vision-only\\(M)}
    & \makecell{Total\\(B)}
    & \multicolumn{1}{c}{Avg}
    & \multicolumn{1}{c}{Gen}
    & \multicolumn{1}{c}{Knw}
    & \multicolumn{1}{c}{OCR}
    & \multicolumn{1}{c}{Vis}
    & \multicolumn{1}{c}{Avg}
    & \multicolumn{1}{c}{Com}
    & \multicolumn{1}{c}{Sci}
    & \multicolumn{1}{c}{Knw}
    & \multicolumn{1}{c@{}}{Rdg}
  \\
  \midrule
  \recipeglyph{scratch}Scratch encoder & 401 & 2.1 & \textbf{40.2} & \textbf{54.6} & 34.5 & \textbf{26.9} & \textbf{44.9} & \textbf{52.5} & \textbf{65.0} & \textbf{47.2} & 41.5 & \textbf{56.4} \\
  \recipeglyph{encoderfree}Encoder-free & 1 & 1.7 & 36.9 & 50.8 & 34.3 & 17.5 & \textbf{45.2} & \textbf{52.6} & \textbf{65.3} & \textbf{46.6} & 42.9 & \textbf{55.6} \\
  \recipeglyph{}Two-layer projection & 18 & 1.7 & 38.8 & 52.8 & 34.4 & 23.7 & \textbf{44.2} & \textbf{52.3} & \textbf{65.9} & \textbf{46.7} & \textbf{43.5} & 53.2 \\
  \recipeglyph{}Hierarchical MLP & 44 & 1.8 & \textbf{40.2} & \textbf{54.2} & \textbf{35.9} & \textbf{26.7} & \textbf{44.2} & \textbf{52.8} & \textbf{65.1} & 46.1 & \textbf{44.3} & \textbf{55.7} \\
  \bottomrule
\end{tabular}

  \vspace{0.8em}
  
\setlength{\tabcolsep}{3.0pt}
\renewcommand{\arraystretch}{1.05}
\begin{tabular}{@{}llrlrrr@{}}
  \toprule
  \multicolumn{4}{>{\columncolor{PaperGray100!55}[.5\tabcolsep][.5\tabcolsep]}c}{\textbf{Architecture}}
    & \multicolumn{3}{>{\columncolor{PaperGray100!55}[.5\tabcolsep][.5\tabcolsep]}c}{\textbf{Input}} \\
  \cmidrule(l{.5\tabcolsep}r{.5\tabcolsep}){1-4}\cmidrule(l{.5\tabcolsep}r{.5\tabcolsep}){5-7}
  Variant & \makecell[l]{Vision-only\\layers} & \makecell{Vision-only\\params (M)} & \makecell[l]{Position\\embedding} & \makecell{Patch\\size} & Resolution & \makecell{Vision\\tokens} \\
  \midrule
  \recipeglyph{scratch}Scratch encoder & 27 Transformer blocks & 401 & learned $+$ 2D RoPE & $14\times14$ & $224\times224$ & 256 \\
  \recipeglyph{encoderfree}Encoder-free & 1 linear & 1 & fixed 2D sine--cosine & $14\times14$ & $224\times224$ & 256 \\
  \recipeglyph{}Two-layer projection & 2 linear & 18 & learned factorized 2D & $48\times48$ & $768\times768$ & 256 \\
  \recipeglyph{}Hierarchical MLP & 4 linear & 44 & none & $40\times40$ & $640\times640$ & 256 \\
  \bottomrule
\end{tabular}

\end{table}

\subsection{Comparison with prior work}
\label{sec:appx:shukor}
Prior work, \citet{shukor2025scaling}, find that encoder-free models match or slightly outperform an encoder trained from scratch, whereas we find that the encoder leads (\cref{sec:encoder-free}).
To investigate this discrepancy, we examine two choices in their setup that may work against the encoder, a reduced encoder learning rate and using the averaged validation loss.
We also study a third difference, using alt-text captions rather than regenerated captions.
First, we evaluate the runs of the scaling ladder (\cref{sec:appx:size-ladder}) on their averaged loss.
We also retrain the ladder's 0.6B and 1B backbones, sizes within the range of their study, with a reduced encoder learning rate, alt-text captions, or both (\cref{tab:appx:shukor}).
Every other setting matches the ladder's runs, which serve as baselines.

\paragraph{Encoder learning rate.}
\citet{shukor2025scaling} train their from-scratch encoder at a reduced learning rate, the best of the $0.1\times$, $0.01\times$, and $0.001\times$ multipliers they compare.
We train the encoder at $0.01\times$ the backbone's peak learning rate, and also at $0.1\times$ at 1B, with the projector at the full rate.
Instruction tuning trains every encoder at its full rate.
We find that lowering the rate shrinks the encoder's lead (\cref{tab:appx:shukor}), and most of the drop comes between the full rate and $0.1\times$, whereas their sweep starts at $0.1\times$ and never trains the encoder at the full rate.
At $0.01\times$, the encoder is comparable to encoder-free on Avg.\ VQA at 1B and loses its caption-loss lead entirely at 0.6B.

\paragraph{Averaged validation loss.}
\citet{shukor2025scaling} fit their scaling laws to validation loss averaged over interleaved, caption, and text data, with captions that are web alt-text.
The text loss does not depend on the image, and in our ladder the interleaved loss barely differs between the encoder and encoder-free.
Differences between vision components therefore have a weaker effect on the average, which makes it a noisier signal for comparing them.
In our scaling ladder, the averaged loss barely separates the encoder from encoder-free, whereas held-out caption loss separates them clearly and tracks Avg.\ VQA (\cref{fig:scaling:loss-vqa}).
At 1.7B and 128 tokens per parameter, the two reach the same averaged loss while the encoder leads by 4.8 Avg.\ VQA points (\cref{fig:appx:shukor-loss}).

\begin{figure}[!htbp]
  \centering
  \includegraphics[width=\linewidth]{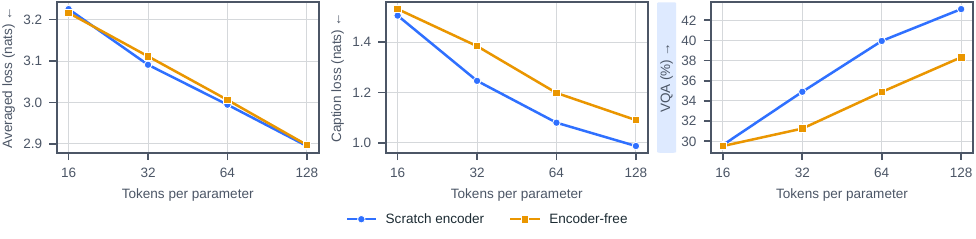}
  \caption{
    \textbf{Comparing on averaged loss misses the encoder's lead on caption loss and Avg.\ VQA.}
    All runs use the 1.7B model from the scaling ladder.
    The averaged loss is the mean of the alt-text caption, interleaved, and text losses.
  }
  \label{fig:appx:shukor-loss}
\end{figure}

\paragraph{Alt-text captions.}
We also train on the images' original alt-text instead of the regenerated captions, for the encoder and encoder-free alike.
This delays the encoder's lead, as the two tie on alt-text caption loss up to 32 tokens per parameter.
With both changes (a $0.01\times$ encoder learning rate and alt-text captions), encoder-free reaches a lower alt-text caption loss than the encoder at both sizes and every budget, by 0.04 to 0.07 nats, as \citet{shukor2025scaling} report.
Overall, their finding may reflect an under-trained encoder or an averaged loss that can hide its benefit.

\begin{table}[!htbp]
  \centering
  \small
  \caption{
    \textbf{Reducing encoder learning rate removes most or all of the encoder's lead.}
    All variants use the 1B backbone from the scaling ladder at 64 or 128 TPP, i.e., $3.7\times10^{20}$ or $7.3\times10^{20}$~FLOPs of lifetime compute.
    The encoder learning rate scales only the encoder's peak learning rate.
    Alt-text rows use the original alt-text instead of regenerated captions, in training and in the held-out loss.
    Bold: best in each column within each caption source.
  }
  \label{tab:appx:shukor}
  
\renewcommand{\arraystretch}{1.05}
\begin{tabular}{@{}llrrrr@{}}
  \toprule
  \multicolumn{2}{>{\columncolor{PaperGray100!55}[.5\tabcolsep][.5\tabcolsep]}c}{\textbf{Setup}}
    & \multicolumn{2}{>{\columncolor{PaperBlue100!55}[.5\tabcolsep][.5\tabcolsep]}c}{\textbf{Held-out caption loss $\downarrow$}}
    & \multicolumn{2}{>{\columncolor{PaperBlue100!55}[.5\tabcolsep][.5\tabcolsep]}c}{\textbf{Avg.\ VQA (\%) $\uparrow$}} \\
  \cmidrule(l{.5\tabcolsep}r{.5\tabcolsep}){1-2}\cmidrule(l{.5\tabcolsep}r{.5\tabcolsep}){3-4}\cmidrule(l{.5\tabcolsep}r{.5\tabcolsep}){5-6}
  Recipe & Encoder learning rate & \multicolumn{1}{c}{64 TPP} & \multicolumn{1}{c}{128 TPP} & \multicolumn{1}{c}{64 TPP} & \multicolumn{1}{c@{}}{128 TPP} \\
  \midrule
  \multicolumn{6}{@{}l}{\textit{Regenerated captions}} \\
  Scratch encoder & $1\times$ (ours) & \textbf{1.229} & \textbf{1.096} & \textbf{32.9} & \textbf{36.3} \\
  Scratch encoder & $0.1\times$ & 1.347 & 1.211 & 30.6 & 32.8 \\
  Scratch encoder & $0.01\times$~\citep{shukor2025scaling} & 1.372 & 1.230 & 28.8 & 31.2 \\
  Encoder-free & -- & 1.388 & 1.243 & 29.1 & 31.0 \\
  \arrayrulecolor{PaperGray400}\cmidrule{1-6}\arrayrulecolor{black}
  \multicolumn{6}{@{}l}{\textit{Alt-text captions}} \\
  Scratch encoder & $1\times$ (ours) & \textbf{3.167} & \textbf{2.826} & \textbf{29.4} & \textbf{35.5} \\
  Scratch encoder & $0.01\times$~\citep{shukor2025scaling} & 3.291 & 3.167 & 27.6 & 28.5 \\
  Encoder-free & -- & 3.247 & 3.126 & 26.9 & 27.3 \\
  \bottomrule
\end{tabular}

\end{table}

  \FloatBarrier

\section{In-fusion vision capacity}
\label{sec:appx:visual-capacity}

This section describes the \ours variants of \cref{sec:capacity}, which add vision-only modules inside the backbone.

\subsection{Implementation and sizing}
\label{sec:vis-capacity:parameterization}

In each augmented layer, \oursffn adds a vision-only SwiGLU branch of width $F$ beside the shared FFN, reading the same normalized input.
For backbone width $D$ and $L_a$ augmented layers, it adds $P_{\mathrm{FFN}}=L_a(3DF+D)$ parameters.
\oursqkvostyled also adds zero-initialized vision-only query, key, value, and output projections with the shapes of the shared ones.
For $H_Q$ query heads, $H_{KV}$ key--value heads, and head dimension $d_h$, they add $P_{\mathrm{QKVO}}=2L_aDd_h(H_Q+H_{KV})$ parameters.
Attention runs once over the combined projections, so there is no second attention stream.
Unless otherwise noted, every allocation augments the first eight layers ($L_a=8$), and \cref{tab:capacity:qkvo} lists the resulting widths and parameter counts.

The depth ablation uses \oursffn because its branch can be narrowed to fit the budget in any number of layers.
Vision-only attention projections have the shapes of the shared ones, so in every layer they would take 0.35B of the 0.4B budget at 1.7B (28 layers) and 0.94B at 4B (36 layers), more than the whole budget.

\begin{table}[!htbp]
  \centering
  \small
  \caption{
    \textbf{The \oursqkvo variant is matched with \oursffn.}
    Both have a fixed added-parameter budget.
    Note that the proportion spent on vision-only attention projections grows with backbone width.
    All rows are checkpoints before the learning rate decay, at $2.34\times10^{21}$~FLOPs (1.7B) and $1.04\times10^{22}$~FLOPs (4B) of lifetime compute.
    Bold: within one point of the better allocation at each backbone scale.
  }
  \label{tab:capacity:qkvo}
  \resizebox{\linewidth}{!}{
\setlength{\tabcolsep}{2.6pt}
\renewcommand{\arraystretch}{1.05}
\begin{tabular}{@{}lrrrrrrrrrrrrr@{}}
  \toprule
  \multicolumn{2}{>{\columncolor{PaperGray100!55}[.5\tabcolsep][.5\tabcolsep]}c}{\textbf{Architecture}}
    & \multicolumn{2}{>{\columncolor{PaperGray100!55}[.5\tabcolsep][.5\tabcolsep]}c}{\textbf{Added params (M)}}
    & \multicolumn{5}{>{\columncolor{PaperBlue100!55}[.5\tabcolsep][.5\tabcolsep]}c}{\textbf{VQA (\%) $\uparrow$}}
    & \multicolumn{5}{>{\columncolor{PaperGreen100!55}[.5\tabcolsep][.5\tabcolsep]}c}{\textbf{Text (\%) $\uparrow$}} \\
  \cmidrule(l{.5\tabcolsep}r{.5\tabcolsep}){1-2}\cmidrule(l{.5\tabcolsep}r{.5\tabcolsep}){3-4}\cmidrule(l{.5\tabcolsep}r{.5\tabcolsep}){5-9}\cmidrule(l{.5\tabcolsep}r{.5\tabcolsep}){10-14}
  Variant & \makecell{FFN\\width} & FFN & QKVO & \multicolumn{1}{c}{Avg} & \multicolumn{1}{c}{Gen} & \multicolumn{1}{c}{Knw} & \multicolumn{1}{c}{OCR} & \multicolumn{1}{c}{Vis} & \multicolumn{1}{c}{Avg} & \multicolumn{1}{c}{Com} & \multicolumn{1}{c}{Sci} & \multicolumn{1}{c}{Knw} & \multicolumn{1}{c@{}}{Rdg} \\
  \midrule
  \multicolumn{14}{@{}l}{\textit{1.7B backbone}} \\
  \recipeglyph{infusionffn}\oursffnstyled{} & 8,192 & 402.7 & 0.0 & 41.5 & 53.8 & 33.2 & \textbf{33.9} & 45.1 & \textbf{48.8} & \textbf{62.5} & \textbf{44.4} & 35.9 & \textbf{52.6} \\
  \recipeglyph{infusionqkvo}\oursqkvostyled{} & 6,144 & 302.0 & 100.7 & \textbf{43.1} & \textbf{54.9} & \textbf{36.7} & \textbf{33.8} & \textbf{47.0} & \textbf{48.7} & \textbf{62.2} & \textbf{43.4} & \textbf{37.1} & \textbf{51.9} \\
  \arrayrulecolor{PaperGray400}\cmidrule{1-14}\arrayrulecolor{black}
  \multicolumn{14}{@{}l}{\textit{4B backbone}} \\
  \recipeglyph{infusionffn}\oursffnstyled{} & 6,528 & 401.1 & 0.0 & \textbf{49.3} & 59.0 & \textbf{43.8} & \textbf{40.0} & \textbf{54.2} & \textbf{55.8} & \textbf{66.5} & \textbf{49.0} & \textbf{49.5} & 58.1 \\
  \recipeglyph{infusionqkvo}\oursqkvostyled{} & 3,136 & 192.7 & 209.7 & \textbf{49.4} & \textbf{61.9} & \textbf{43.9} & \textbf{40.5} & 51.2 & \textbf{56.7} & \textbf{67.3} & \textbf{49.5} & \textbf{49.8} & \textbf{60.0} \\
  \bottomrule
\end{tabular}

}
\end{table}

\subsection{Routing, depth, and attention-projection controls}
\label{sec:appx:capacity-design-controls}
\label{sec:appx:capacity-variants} %

In \cref{tab:capacity:design}, shared FFN widening grows the shared FFN of layers 1--8 from width 6,144 to 14,336.
The FFN branches have width 8,192 when placed in eight layers and 2,304 when placed in all 28.

\begin{table}[!htbp]
  \centering
  \small
  \caption{
    \textbf{Reserving the FFN branch for vision tokens keeps its lead with twice the training.}
    All rows use the 1.7B backbone, decayed at the lifetime compute of their block.
    Both FFN branches add 0.4B parameters to layers 1--8 of the encoder-free backbone.
    Bold: within one point of the best in each block.
  }
  \label{tab:appx:capacity:routing-longer}
  
\setlength{\tabcolsep}{3.2pt}
\renewcommand{\arraystretch}{1.05}
\begin{tabular}{@{}llrrrrrrrrrr@{}}
  \toprule
  \multicolumn{2}{>{\columncolor{PaperGray100!55}[.5\tabcolsep][.5\tabcolsep]}c}{\textbf{Architecture}}
    & \multicolumn{5}{>{\columncolor{PaperBlue100!55}[.5\tabcolsep][.5\tabcolsep]}c}{\textbf{VQA (\%) $\uparrow$}}
    & \multicolumn{5}{>{\columncolor{PaperGreen100!55}[.5\tabcolsep][.5\tabcolsep]}c}{\textbf{Text (\%) $\uparrow$}} \\
  \cmidrule(l{.5\tabcolsep}r{.5\tabcolsep}){1-2}\cmidrule(l{.5\tabcolsep}r{.5\tabcolsep}){3-7}\cmidrule(l{.5\tabcolsep}r{.5\tabcolsep}){8-12}
  Variant & Used by & \multicolumn{1}{c}{Avg} & \multicolumn{1}{c}{Gen} & \multicolumn{1}{c}{Knw} & \multicolumn{1}{c}{OCR} & \multicolumn{1}{c}{Vis} & \multicolumn{1}{c}{Avg} & \multicolumn{1}{c}{Com} & \multicolumn{1}{c}{Sci} & \multicolumn{1}{c}{Knw} & \multicolumn{1}{c@{}}{Rdg} \\
  \midrule
  \multicolumn{12}{@{}l}{\textit{$2.6\times10^{21}$ FLOPs (158 TPP)}} \\
  \rowcolor{PaperGray100!50}
  \recipeglyph{scratch}Scratch encoder & Vision & \textbf{44.4} & \textbf{57.2} & 35.7 & \textbf{38.2} & 46.5 & 49.7 & \textbf{63.9} & 45.0 & 36.7 & 53.4 \\
  \rowcolor{PaperGray100!50}
  \recipeglyph{encoderfree}Encoder-free & -- & 39.0 & 50.7 & 35.0 & 25.2 & 45.0 & \textbf{50.7} & \textbf{64.1} & \textbf{45.8} & 38.2 & \textbf{54.8} \\
  \recipeglyph{}Separate FFN branch & All & 42.3 & 54.1 & \textbf{37.4} & 30.8 & 46.7 & \textbf{51.2} & \textbf{64.8} & \textbf{46.3} & \textbf{41.6} & 52.0 \\
  \recipeglyph{infusionffn}\oursffnstyled{} & Vision & \textbf{45.0} & \textbf{57.0} & \textbf{36.7} & 36.0 & \textbf{50.4} & \textbf{50.4} & 63.0 & \textbf{45.3} & \textbf{40.6} & 52.9 \\
  \addlinespace[2pt]
  \multicolumn{12}{@{}l}{\textit{$5.2\times10^{21}$ FLOPs (315 TPP)}} \\
  \rowcolor{PaperGray100!50}
  \recipeglyph{scratch}Scratch encoder & Vision & \textbf{46.5} & \textbf{58.5} & \textbf{38.9} & \textbf{39.4} & 49.4 & \textbf{52.5} & 65.0 & \textbf{47.2} & 41.5 & \textbf{56.4} \\
  \rowcolor{PaperGray100!50}
  \recipeglyph{encoderfree}Encoder-free & -- & 41.9 & 54.7 & 36.3 & 29.0 & 47.6 & \textbf{52.6} & 65.3 & \textbf{46.6} & 42.9 & \textbf{55.6} \\
  \recipeglyph{}Separate FFN branch & All & 45.4 & 57.3 & \textbf{39.7} & 34.9 & \textbf{49.7} & \textbf{53.5} & \textbf{66.5} & \textbf{46.4} & \textbf{46.6} & 54.5 \\
  \recipeglyph{infusionffn}\oursffnstyled{} & Vision & \textbf{46.7} & \textbf{59.3} & \textbf{38.9} & 38.1 & \textbf{50.5} & \textbf{52.7} & 64.8 & \textbf{47.2} & \textbf{46.3} & 52.6 \\
  \bottomrule
\end{tabular}

\end{table}

\paragraph{Routing at matched compute.}
\oursffnstyled and the separate FFN branch differ only in which tokens use the branch.
Because the all-token branch also runs on text tokens, it trains on fewer tokens at matched compute.
Routing the branch to vision tokens leads on Avg.\ VQA at both budgets, while the all-token branch is slightly ahead on Avg.\ Text (\cref{tab:appx:capacity:routing-longer}).
At the longer budget, routing also raises downstream ImageNet probe accuracy from 61.2 to 66.0 (\cref{sec:appx:probe}).
With the same parameters, the separate FFN branch beats shared FFN widening on Avg.\ VQA (\cref{tab:capacity:design}).

\paragraph{Attention-projection controls.}
\label{sec:vis-capacity:qkvo-controls}

In two 4B contrasts at matched training tokens, the vision-only attention projections improve visual understanding both on their own and alongside an FFN branch (\cref{tab:capacity:qkvo-mechanisms}).
The first adds only the attention projections of \oursqkvo to the encoder-free model, and the second compares \oursqkvo with a model that has only its FFN branch.

\begin{table}[!htbp]
  \centering
  \small
  \caption{
    \textbf{Vision-only attention projections improve visual understanding both alone and alongside an FFN branch.}
    Each row is the difference between two 4B models trained on the same number of tokens, before the final decay.
    Caption-loss changes are in nats, where negative is better.
  }
  \label{tab:capacity:qkvo-mechanisms}
  
\setlength{\tabcolsep}{6pt}
\renewcommand{\arraystretch}{1.05}
\begin{tabular}{@{}lrrr@{}}
  \toprule
  \multicolumn{1}{>{\columncolor{PaperGray100!55}[.5\tabcolsep][.5\tabcolsep]}c}{\textbf{Contrast}}
    & \multicolumn{1}{>{\columncolor{PaperBlue100!55}[.5\tabcolsep][.5\tabcolsep]}c}{\textbf{VQA (\%) $\uparrow$}}
    & \multicolumn{2}{>{\columncolor{PaperGray100!55}[.5\tabcolsep][.5\tabcolsep]}c}{\textbf{Other metrics}} \\
  \cmidrule(l{.5\tabcolsep}r{.5\tabcolsep}){1-1}\cmidrule(l{.5\tabcolsep}r{.5\tabcolsep}){2-2}\cmidrule(l{.5\tabcolsep}r{.5\tabcolsep}){3-4}
  {} & \makecell{$\Delta$ Avg\\{[}95\% CI{]}}
    & \makecell{$\Delta$ OOD\\caption loss}
    & \makecell{$\Delta$ ImageNet\\top-1} \\
  \midrule
  QKVO-only $-$ encoder-free & +1.8 [1.0, 2.5] & $-$0.059 & +3.73 \\
  In-fusion $-$ same-width FFN branch & +1.1 [0.4, 1.9] & $-$0.013 & +0.75 \\
  \bottomrule
\end{tabular}

\end{table}

\subsection{Comparison at longer training}
\label{sec:appx:capacity-long-horizon}

\begin{table}[!htbp]
  \centering
  \small
  \caption{
    \textbf{\oursqkvostyled is comparable to the scratch encoder at longer training budgets.}
    Both model scales train to 1.11T token-equivalents.
    Bold: within one point of the best at each scale.
  }
  \label{tab:capacity:endpoints}
  \label{tab:capacity:long-horizon} %
  \label{tab:capacity:encoder-endpoints}
  
\setlength{\tabcolsep}{6.2pt}
\renewcommand{\arraystretch}{1.05}
\begin{tabular}{@{}lrrrrrrrrrr@{}}
  \toprule
  \multicolumn{1}{>{\columncolor{PaperGray100!55}[.5\tabcolsep][.5\tabcolsep]}c}{\textbf{Architecture}}
    & \multicolumn{5}{>{\columncolor{PaperBlue100!55}[.5\tabcolsep][.5\tabcolsep]}c}{\textbf{VQA (\%) $\uparrow$}}
    & \multicolumn{5}{>{\columncolor{PaperGreen100!55}[.5\tabcolsep][.5\tabcolsep]}c}{\textbf{Text (\%) $\uparrow$}} \\
  \cmidrule(l{.5\tabcolsep}r{.5\tabcolsep}){1-1}\cmidrule(l{.5\tabcolsep}r{.5\tabcolsep}){2-6}\cmidrule(l{.5\tabcolsep}r{.5\tabcolsep}){7-11}
  Model & \multicolumn{1}{c}{Avg} & \multicolumn{1}{c}{Gen} & \multicolumn{1}{c}{Knw} & \multicolumn{1}{c}{OCR} & \multicolumn{1}{c}{Vis} & \multicolumn{1}{c}{Avg} & \multicolumn{1}{c}{Com} & \multicolumn{1}{c}{Sci} & \multicolumn{1}{c}{Knw} & \multicolumn{1}{c@{}}{Rdg} \\
  \midrule
  \multicolumn{11}{@{}l}{\textit{1.7B backbone, $1.30{\times}10^{22}$ FLOPs}} \\
  \rowcolor{PaperGray100!50}
  \recipeglyph{scratch}Scratch encoder & \textbf{48.9} & \textbf{60.6} & 41.3 & \textbf{42.5} & 51.2 & \textbf{55.3} & \textbf{66.2} & \textbf{48.1} & 49.3 & 57.6 \\
  \recipeglyph{encoderfree}Encoder-free & 44.8 & 56.4 & 38.8 & 34.7 & 49.3 & \textbf{55.7} & \textbf{66.5} & \textbf{48.1} & 48.6 & \textbf{59.4} \\
  \recipeglyph{infusionffn}\oursffnstyled{} & \textbf{48.6} & \textbf{60.1} & 40.9 & 40.7 & \textbf{52.8} & \textbf{55.6} & \textbf{66.1} & \textbf{48.5} & \textbf{50.5} & 57.3 \\
  \recipeglyph{infusionqkvo}\oursqkvostyled{} & \textbf{49.2} & \textbf{60.0} & \textbf{43.0} & \textbf{41.6} & \textbf{52.1} & \textbf{55.9} & \textbf{66.9} & \textbf{48.5} & 49.3 & \textbf{59.0} \\
  \arrayrulecolor{PaperGray400}\cmidrule{1-11}\arrayrulecolor{black}
  \multicolumn{11}{@{}l}{\textit{4B backbone, $2.90{\times}10^{22}$ FLOPs}} \\
  \rowcolor{PaperGray100!50}
  \recipeglyph{scratch}Scratch encoder & \textbf{54.5} & \textbf{66.4} & \textbf{49.1} & \textbf{45.5} & \textbf{57.0} & \textbf{60.4} & \textbf{69.5} & 51.5 & \textbf{57.5} & \textbf{63.0} \\
  \recipeglyph{encoderfree}Encoder-free & 51.7 & 63.9 & 45.8 & 41.7 & 55.6 & \textbf{61.0} & \textbf{69.5} & \textbf{52.9} & \textbf{58.3} & \textbf{63.3} \\
  \recipeglyph{infusionffn}\oursffnstyled{} & 53.4 & 65.1 & 46.9 & \textbf{45.3} & 56.5 & \textbf{60.2} & \textbf{70.0} & \textbf{52.2} & 56.4 & \textbf{62.4} \\
  \recipeglyph{infusionqkvo}\oursqkvostyled{} & \textbf{54.3} & \textbf{65.6} & 47.9 & \textbf{45.7} & \textbf{57.9} & \textbf{60.7} & \textbf{69.2} & \textbf{52.6} & \textbf{57.8} & \textbf{63.1} \\
  \bottomrule
\end{tabular}

\end{table}

At the longest training budget, 1.11T token-equivalents, \oursqkvo is comparable to the encoder on both Avg.\ Text and Avg.\ VQA at both model scales (\cref{tab:capacity:endpoints}).
\oursffnstyled is comparable to the encoder at 1.7B and slightly behind it on Avg.\ VQA at 4B, where it also trails \oursqkvo at every budget (\cref{fig:capacity:qkvo-scale}).

\subsection{Where vision tokens are processed}
\label{sec:appx:layer-probes}
To see where each design processes vision tokens, we probe the 1.7B models at the longest training budget (\cref{fig:appx:layer-probes}), adapting the probes of \citet{chen2026encoderfree}.
For each backbone block, we measure how much the vision tokens change as one minus the cosine similarity between their hidden state before and after the block, and how much attention caption tokens place on them.
In both in-fusion models, vision tokens change mostly in the blocks with vision-only modules and little after them.
The encoder-free model keeps changing them through its early and middle shared blocks.
The encoder's tokens enter the backbone already processed, so they change mostly in the first few blocks.
After the in-fusion blocks, caption tokens attend to vision tokens more than in the encoder and encoder-free models.
The same pattern holds at a fifth of the training budget.
These probes suggest that in-fusion capacity performs vision-specific processing in its own early blocks and passes stable vision tokens to the shared backbone, whereas an encoder-free model builds them up across its shared layers.

\begin{figure}[!htbp]
  \centering
  \includegraphics[width=\linewidth]{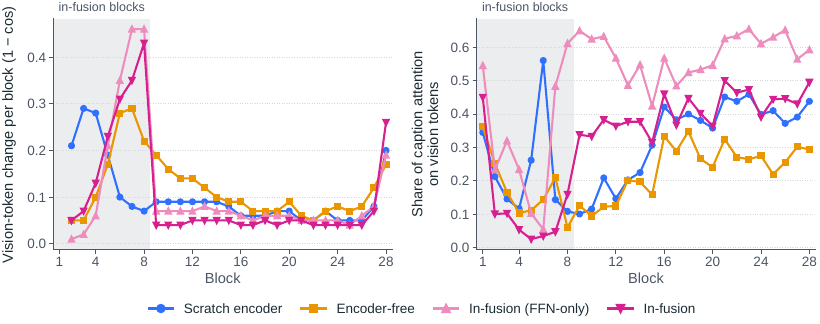}
  \caption{
    \textbf{In-fusion capacity confines vision processing to the earliest blocks, while encoder-free spreads it across shared layers.}
    Vision-token change is defined as $1-\cos(h_L, h_{L-1})$ between consecutive blocks.
    The shaded band marks blocks 1--8, where the in-fusion modules sit.
    All models use the 1.7B backbone at $1.30\times10^{22}$~FLOPs of lifetime compute.
  }
  \label{fig:appx:layer-probes}
\end{figure}

  \FloatBarrier
  
\section{Decay, learning rate, and caption loss}
\label{sec:appx:scaling}
This section lists the evaluation performance of each endpoint (post-decay) of our main comparisons, shows that the shared peak learning rate (\cref{sec:protocol:length}) does not decide the comparisons, and relates held-out caption loss to Avg.\ VQA (\cref{sec:encoder-free}).

\subsection{Decayed endpoints}
\label{sec:appx:scaling:ladder}

\Cref{tab:appx:ladder} lists the evaluation performance of each recipe's endpoint at each budget of \cref{sec:protocol:length}, next to a text-only reference trained to the same lifetime compute.
Each endpoint is evaluated after learning rate decay.

\begin{table}[!htbp]
  \centering
  \small
  \caption{
    \textbf{Encoder-free trails the encoder at every compute budget, while in-fusion is comparable to it.}
    Enc is the from-scratch encoder, EF encoder-free, and FFN and QKVO mark \oursffn and \oursqkvo.
    All models in a row share the same lifetime compute.
    Dashes mark models not trained to that budget.
    Bold: within one point of the best in each row, excluding the text-only reference.
  }
  \label{tab:appx:ladder}
  
\setlength{\tabcolsep}{4.0pt}
\renewcommand{\arraystretch}{1.05}
\begin{tabular}{@{}rrrrrrrrrrr@{}}
  \toprule
  \multicolumn{2}{>{\columncolor{PaperGray100!55}[.5\tabcolsep][.5\tabcolsep]}c}{\textbf{Budget}} & \multicolumn{4}{>{\columncolor{PaperBlue100!55}[.5\tabcolsep][.5\tabcolsep]}c}{\textbf{VQA (\%) $\uparrow$}} & \multicolumn{5}{>{\columncolor{PaperGreen100!55}[.5\tabcolsep][.5\tabcolsep]}c}{\textbf{Text (\%) $\uparrow$}} \\
  \cmidrule(l{.5\tabcolsep}r{.5\tabcolsep}){1-2}\cmidrule(l{.5\tabcolsep}r{.5\tabcolsep}){3-6}\cmidrule(l{.5\tabcolsep}r{.5\tabcolsep}){7-11}
  FLOPs & TPP & \multicolumn{1}{c}{Enc} & \multicolumn{1}{c}{EF} & \multicolumn{1}{c}{FFN} & \multicolumn{1}{c}{QKVO} & \multicolumn{1}{c}{Text-only} & \multicolumn{1}{c}{Enc} & \multicolumn{1}{c}{EF} & \multicolumn{1}{c}{FFN} & \multicolumn{1}{c@{}}{QKVO} \\
  \midrule
  \multicolumn{11}{@{}l}{\textit{1.7B backbone}} \\
  $6.5{\times}10^{20}$ & 39 & \textbf{37.3} & 32.0 & 36.3 & \textbf{37.4} & -- & 45.3 & \textbf{46.4} & \textbf{45.9} & \textbf{45.7} \\
  $1.30{\times}10^{21}$ & 79 & \textbf{41.2} & 34.8 & \textbf{41.4} & \textbf{40.4} & 48.9 & 47.3 & \textbf{48.3} & \textbf{48.4} & \textbf{48.1} \\
  $2.60{\times}10^{21}$ & 158 & \textbf{44.4} & 39.0 & \textbf{45.0} & 42.6 & 52.4 & \textbf{49.7} & \textbf{50.7} & \textbf{50.4} & \textbf{50.7} \\
  $5.21{\times}10^{21}$ & 315 & \textbf{46.5} & 41.9 & \textbf{46.7} & \textbf{46.5} & 54.4 & \textbf{52.5} & \textbf{52.6} & \textbf{52.7} & \textbf{53.4} \\
  $7.81{\times}10^{21}$ & 473 & \textbf{47.5} & 43.0 & -- & -- & 54.9 & \textbf{54.4} & \textbf{53.7} & -- & -- \\
  $1.04{\times}10^{22}$ & 631 & \textbf{49.0} & 43.8 & -- & -- & 56.5 & \textbf{54.3} & \textbf{55.3} & -- & -- \\
  $1.30{\times}10^{22}$ & 788 & \textbf{48.9} & 44.8 & \textbf{48.6} & \textbf{49.2} & 56.5 & \textbf{55.3} & \textbf{55.7} & \textbf{55.6} & \textbf{55.9} \\
  \arrayrulecolor{PaperGray400}\cmidrule{1-11}\arrayrulecolor{black}
  \multicolumn{11}{@{}l}{\textit{4B backbone}} \\
  $1.45{\times}10^{21}$ & 15 & -- & -- & -- & -- & 51.2 & -- & -- & -- & -- \\
  $2.90{\times}10^{21}$ & 31 & \textbf{47.8} & 41.0 & 44.8 & 45.5 & 54.8 & \textbf{52.0} & \textbf{51.5} & 50.8 & \textbf{51.9} \\
  $5.79{\times}10^{21}$ & 61 & \textbf{51.2} & 44.9 & 48.4 & 49.0 & 57.4 & \textbf{56.4} & \textbf{56.0} & \textbf{55.8} & \textbf{55.9} \\
  $1.16{\times}10^{22}$ & 122 & \textbf{51.9} & 48.4 & 49.8 & \textbf{51.8} & 59.0 & \textbf{57.9} & \textbf{57.0} & \textbf{57.1} & \textbf{57.7} \\
  $1.74{\times}10^{22}$ & 183 & \textbf{54.2} & 50.2 & -- & -- & -- & \textbf{58.9} & \textbf{59.2} & -- & -- \\
  $2.90{\times}10^{22}$ & 306 & \textbf{54.5} & 51.7 & 53.4 & \textbf{54.3} & -- & \textbf{60.4} & \textbf{61.0} & \textbf{60.2} & \textbf{60.7} \\
  \bottomrule
\end{tabular}

\end{table}

\subsection{Learning rate}
\label{sec:appx:scaling:lr}
All recipes share one peak learning rate per backbone size, a choice that could bias compute comparisons~\citep{porian2024resolving}.
Therefore, at 1.7B backbone size, we repeat the encoder, \oursqkvo, and encoder-free recipes at half and twice the peak learning rate (encoder-free also at four times), at 39 and 79 TPP (\cref{fig:appx:lr-sensitivity}).
At every learning rate, the encoder leads encoder-free and \oursqkvo matches or exceeds the encoder.
Higher learning rates raise Avg.\ Text for all three recipes, and have lesser effects on Avg.\ VQA.
The shared learning rate therefore does not decide the visual-understanding comparisons of \cref{sec:encoder-free,sec:capacity}.

\begin{figure}[!htbp]
  \centering
  \includegraphics[width=0.62\linewidth]{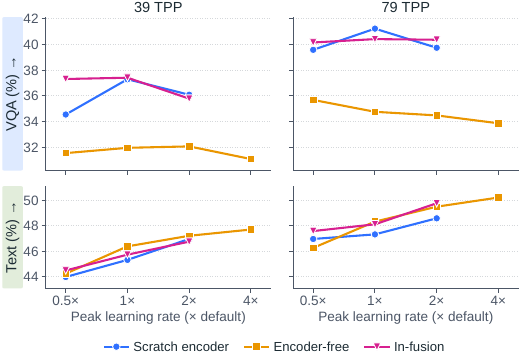}
  \caption{
    \textbf{The encoder maintains its lead over encoder-free, and \oursqkvo is comparable to the encoder, at every peak learning rate.}
    All models are 1.7B and trained to 39 and 79 TPP, i.e., $6.5\times10^{20}$ and $1.3\times10^{21}$~FLOPs of lifetime compute.
  }
  \label{fig:appx:lr-sensitivity}
\end{figure}

\subsection{Held-out caption loss}
\label{sec:appx:scaling:loss}
We find that across all four of the recipes, caption loss tracks Avg.\ VQA along one trend ($r=-0.96$; \cref{fig:scaling:loss-vqa}).
By category, OCR and chart questions track it most closely.

\begin{figure}[!htbp]
  \centering
  \includegraphics[width=0.62\linewidth]{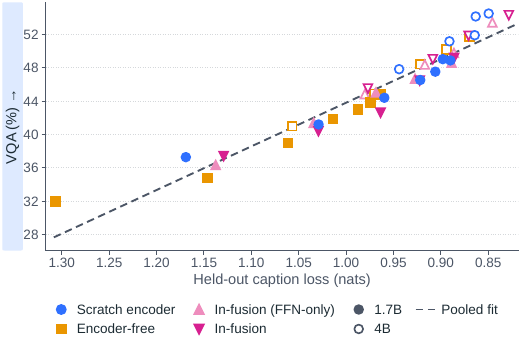}
  \caption{
    \textbf{Held-out caption loss tracks Avg.\ VQA along one trend shared by all four recipes.}
    Held-out caption loss uses images never seen in training.
    The line is a least-squares fit over all endpoints.
    All points are evaluated after learning rate decay.
  }
  \label{fig:scaling:loss-vqa}
\end{figure}

\section{Scaling ladder}
\label{sec:appx:ladder-small}
\label{sec:appx:size-ladder}

\Cref{sec:capacity:sizes} tests whether the comparisons of \cref{sec:encoder-free,sec:capacity} hold as the backbone grows, using a ladder of backbone sizes.
This section gives the ladder's setup, the scaling laws fitted to it, and how well the laws predict runs they were not fitted on.

\paragraph{Setup.}
The ladder trains the encoder-free, encoder, and \oursqkvo recipes under the protocol of \cref{sec:protocol} on backbones from 0.6B to 6.4B parameters.
The backbones follow the architecture of the 1.7B backbone, with query heads of dimension 128 spanning the hidden size, half as many key--value heads, and a feed-forward width three times the hidden size.
The vision module grows with the backbone (\cref{tab:appx:size-ladder}).
The encoder keeps the proportions of the 1.7B pairing: half the backbone's hidden size $d$, one layer fewer than the backbone, an MLP four times its own width, and $\mathrm{round}(0.75\,d/128)$ attention heads of dimension 128.
Its parameter count therefore stays at a nearly fixed ratio to the backbone's non-embedding parameters.
The in-fusion modules occupy the first $\mathrm{round}(8L/28)$ of the backbone's $L$ layers, 29\% of the depth, as at 1.7B.
Their widths are set so that they match the encoder's parameter count.

All recipes at a backbone size share one peak learning rate and one batch size (\cref{tab:appx:size-ladder}).
The peak learning rate is $3\times10^{-4}\sqrt{1.72\times10^{9}/N_{\mathrm{tot}}}$, rounded, where $N_{\mathrm{tot}}$ is the total backbone parameter count.
The batch size grows by about 1.5 times per backbone size.
Warmup lasts 2 tokens per non-embedding backbone parameter (2 TPP), \eg, 2.8B tokens for the 1.7B backbone, which has 1.4B non-embedding parameters.
The other settings follow \cref{tab:appx:hyperparameters}.
The 1.7B and 4B runs of \cref{sec:encoders,sec:encoder-free,sec:capacity} instead keep a batch of 1,024 sequences and 4,000 warmup updates at both sizes.
At 1.7B, where the ladder's batch is also 1,024, each ladder recipe is within two Avg.\ VQA points of those runs when interpolated to their budgets (\cref{sec:encoder-free,sec:capacity}).

Each recipe and size has one constant-learning-rate run, decayed over the final 10\% of training at several budgets (\cref{tab:appx:size-ladder-scores}).
Each point in \cref{fig:capacity:sizes,fig:appx:size-ladder-laws} is one decayed endpoint.
At 1.7B, the ladder's encoder is the 0.4B encoder of \cref{sec:protocol}.
At 6.4B, we also train the encoder recipe with the 0.4B encoder of the main runs, to check that this smaller encoder does not limit larger backbones (\cref{sec:encoders:encoder-architecture}).
\Cref{tab:appx:size-ladder-scores} lists the downstream scores of every run the laws are fitted on.

\paragraph{Scaling laws.}
For Avg.\ VQA, caption loss, and text loss, we fit a separate law for each recipe.
For the losses it takes the form $L(N,D)=E+A\,N^{-\alpha}+B\,D^{-\beta}$~\citep{hoffmann2022chinchilla}, with $N$ the non-embedding backbone parameters and $D$ the token-equivalents (\cref{sec:protocol:accounting}), fitted by minimizing a Huber loss on log loss.
For Avg.\ VQA, it is linear in $\ln N$ and $\ln D$, fitted by least squares.
Each is fitted on the recipe's 17 ladder runs (\cref{tab:appx:size-ladder-fits}).
From 1.7B to 6.4B, encoder-free needs 2.6 to 3.1 times the encoder's compute on Avg.\ VQA, 2.4 to 3.8 times on caption loss, and about the same on text loss.
\oursqkvostyled needs about the same compute as the encoder on Avg.\ VQA and text loss, and slightly less on caption loss.
\Cref{tab:appx:size-ladder-gof} gives the scaling laws' errors on every run and on each left-out backbone size.
Fitted on the ladder alone, the caption-loss laws also predict the separately trained 1.7B and 4B runs of \cref{sec:encoders,sec:encoder-free,sec:capacity} with an average absolute error of 1.6\% (\cref{tab:appx:size-ladder-heldout}).
Their warmup differs from the ladder's, and at 4B, so do the batch size and the size of the vision module.
\Cref{fig:appx:size-ladder-laws} shows each recipe's caption-loss law at every backbone size against the runs it is fitted on.

\begin{table}[!htbp]
  \centering
  \footnotesize
  \caption{
    \textbf{Architectures and optimization settings of the scaling ladder.}
    The vision module grows with the backbone size.
    Ratio is the vision-only parameter count over the backbone's non-embedding count, which stays at around 0.29.
    The in-fusion columns describe the FFN+QKVO variant (\cref{sec:capacity:infusion}).
    Batch counts sequences of 4,096 tokens, and warmup counts updates.
    Backbones are named by total parameters, including the tied embedding.
  }
  \label{tab:appx:size-ladder}
  \setlength{\tabcolsep}{3pt}
  \begin{tabular}{@{}lrrrrrrrrrr@{}}
    \toprule
    & \multicolumn{2}{c}{Backbone} & \multicolumn{3}{c}{Encoder} & \multicolumn{2}{c}{In-fusion} & \multicolumn{3}{c}{Optimization} \\
    \cmidrule(lr){2-3}\cmidrule(lr){4-6}\cmidrule(lr){7-8}\cmidrule(l){9-11}
    & \makecell[r]{Non-emb.\\(M)} & \makecell[r]{Width /\\layers} & \makecell[r]{Width /\\layers} & \makecell[r]{Vision-only\\(M)} & Ratio & Layers & Width & \makecell[r]{Peak\\LR} & Batch & Warmup \\
    \midrule
    0.6B & 393   & 1,280 / 20 & 640 / 19   & 115   & 0.29 & 1--6  & 3,584 & $5.1\times10^{-4}$ & 448   & 429 \\
    1B   & 793   & 1,536 / 28 & 768 / 27   & 227   & 0.29 & 1--8  & 4,544 & $3.9\times10^{-4}$ & 640   & 605 \\
    1.7B & 1,409 & 2,048 / 28 & 1,024 / 27 & 401   & 0.28 & 1--8  & 6,144 & $3.0\times10^{-4}$ & 1,024 & 672 \\
    3.2B & 2,831 & 2,560 / 36 & 1,280 / 35 & 809   & 0.29 & 1--10 & 7,936 & $2.2\times10^{-4}$ & 1,536 & 900 \\
    6.4B & 5,848 & 3,328 / 44 & 1,664 / 43 & 1,693 & 0.29 & 1--13 & 9,664 & $1.6\times10^{-4}$ & 2,304 & 1,239 \\
    \bottomrule
  \end{tabular}
\end{table}

\begin{table}[!htbp]
  \centering
  \footnotesize
  \caption{
    \textbf{In-fusion is comparable to the encoder at every size, whereas encoder-free trails both on VQA but stays comparable on text.}
    This shows metrics for every scaling ladder run (\cref{fig:appx:size-ladder-laws}).
    Bold: within one point of the best in each row.
  }
  \label{tab:appx:size-ladder-scores}
  
\setlength{\tabcolsep}{4pt}
\renewcommand{\arraystretch}{1.05}
\begin{tabular}{@{}lrrrrrrrr@{}}
  \toprule
  \multicolumn{3}{>{\columncolor{PaperGray100!55}[.5\tabcolsep][.5\tabcolsep]}c}{\textbf{Setup}} & \multicolumn{3}{>{\columncolor{PaperBlue100!55}[.5\tabcolsep][.5\tabcolsep]}c}{\textbf{VQA (\%) $\uparrow$}} & \multicolumn{3}{>{\columncolor{PaperGreen100!55}[.5\tabcolsep][.5\tabcolsep]}c}{\textbf{Text (\%) $\uparrow$}} \\
  \cmidrule(l{.5\tabcolsep}r{.5\tabcolsep}){1-3}\cmidrule(l{.5\tabcolsep}r{.5\tabcolsep}){4-6}\cmidrule(l{.5\tabcolsep}r{.5\tabcolsep}){7-9}
  Backbone & TPP & FLOPs & \makecell[r]{Scratch\\encoder} & \makecell[r]{Encoder-\\free} & In-fusion & \makecell[r]{Scratch\\encoder} & \makecell[r]{Encoder-\\free} & In-fusion \\
  \midrule
  0.6B & 128 & $2.09{\times}10^{20}$ & \textbf{31.6} & 26.1 & \textbf{30.7} & \textbf{43.3} & \textbf{42.4} & \textbf{42.7} \\
   & 256 & $4.18{\times}10^{20}$ & \textbf{33.4} & 27.9 & \textbf{32.6} & \textbf{44.1} & 43.3 & \textbf{44.4} \\
  \midrule
  1B & 64 & $3.66{\times}10^{20}$ & \textbf{32.9} & 29.1 & \textbf{33.6} & \textbf{43.3} & \textbf{43.5} & \textbf{44.0} \\
   & 128 & $7.32{\times}10^{20}$ & \textbf{36.3} & 31.0 & \textbf{36.0} & \textbf{46.1} & \textbf{45.7} & \textbf{45.7} \\
   & 256 & $1.46{\times}10^{21}$ & \textbf{40.1} & 33.4 & 38.0 & \textbf{47.6} & \textbf{47.9} & \textbf{47.3} \\
  \midrule
  1.7B & 32 & $5.29{\times}10^{20}$ & 34.9 & 31.2 & \textbf{36.1} & \textbf{46.4} & \textbf{45.6} & \textbf{46.0} \\
   & 64 & $1.06{\times}10^{21}$ & \textbf{39.9} & 34.9 & 38.4 & \textbf{48.2} & \textbf{48.3} & \textbf{48.1} \\
   & 128 & $2.12{\times}10^{21}$ & \textbf{43.1} & 38.3 & \textbf{42.2} & \textbf{50.5} & \textbf{50.4} & \textbf{50.5} \\
   & 256 & $4.23{\times}10^{21}$ & \textbf{44.8} & 41.0 & 43.5 & \textbf{52.4} & \textbf{52.8} & \textbf{52.7} \\
  \midrule
  3.2B & 16 & $9.78{\times}10^{20}$ & 36.0 & 31.7 & \textbf{39.3} & \textbf{48.1} & 46.9 & 46.5 \\
   & 32 & $1.96{\times}10^{21}$ & 41.3 & 35.7 & \textbf{42.8} & \textbf{50.4} & \textbf{51.1} & 49.8 \\
   & 64 & $3.91{\times}10^{21}$ & \textbf{46.4} & 39.9 & \textbf{45.9} & \textbf{53.0} & \textbf{53.7} & \textbf{52.7} \\
   & 128 & $7.82{\times}10^{21}$ & \textbf{50.6} & 43.6 & \textbf{50.2} & \textbf{55.4} & \textbf{56.2} & \textbf{55.9} \\
  \midrule
  6.4B & 16 & $3.90{\times}10^{21}$ & \textbf{46.0} & 40.4 & \textbf{46.8} & \textbf{53.0} & \textbf{52.6} & \textbf{52.1} \\
   & 32 & $7.81{\times}10^{21}$ & \textbf{50.2} & 44.3 & 48.9 & \textbf{56.4} & \textbf{56.4} & \textbf{56.2} \\
   & 64 & $1.56{\times}10^{22}$ & \textbf{52.8} & 47.5 & \textbf{52.0} & \textbf{59.1} & \textbf{58.9} & \textbf{59.1} \\
   & 128 & $3.12{\times}10^{22}$ & \textbf{54.6} & 49.1 & \textbf{54.2} & \textbf{61.4} & \textbf{60.7} & \textbf{61.5} \\
  \bottomrule
\end{tabular}

\end{table}

\begin{table}[!htbp]
  \centering
  \footnotesize
  \caption{
    \textbf{Encoder-free needs 2.6 to 3.1 times the encoder's compute on Avg.\ VQA, while in-fusion needs about as much as the encoder.}
    $N$ and $D$ are in billions.
    Each factor is the compute at which a recipe's law reaches a metric value, divided by the compute at which the encoder's law reaches the same value.
    The factors are read at the compute of each run at 64 TPP from 1.7B to 6.4B.
  }
  \label{tab:appx:size-ladder-fits}
  
\setlength{\tabcolsep}{4pt}
\begin{tabular}{@{}lllr@{}}
  \toprule
  Metric & Recipe & Law & Factor \\
  \midrule
  Avg.\ VQA & Scratch encoder & $13.7 + 4.40\ln N + 5.25\ln D$ & -- \\
   & Encoder-free & $10.8 + 4.40\ln N + 4.75\ln D$ & 2.6--3.1 \\
   & In-fusion & $19.3 + 5.29\ln N + 3.94\ln D$ & 1.0--1.1 \\
  \midrule
  Caption loss & Scratch encoder & $E=0.59$, $A=0.28$, $\alpha=0.27$, $B=7.54$, $\beta=0.77$ & -- \\
   & Encoder-free & $E=0.70$, $A=0.18$, $\alpha=0.59$, $B=4.18$, $\beta=0.54$ & 2.4--3.8 \\
   & In-fusion & $E=0.63$, $A=0.25$, $\alpha=0.40$, $B=3.39$, $\beta=0.61$ & 0.9--1.0 \\
  \midrule
  Text loss & Scratch encoder & $E=1.08$, $A=0.58$, $\alpha=0.19$, $B=1.03$, $\beta=0.19$ & -- \\
   & Encoder-free & $E=1.08$, $A=0.58$, $\alpha=0.19$, $B=1.00$, $\beta=0.19$ & 1.0 \\
   & In-fusion & $E=1.09$, $A=0.57$, $\alpha=0.19$, $B=1.03$, $\beta=0.19$ & 1.0 \\
  \bottomrule
\end{tabular}

\end{table}

\begin{table}[!htbp]
  \centering
  \footnotesize
  \caption{
    \textbf{One law per recipe fits every run and predicts each left-out backbone size.}
    Fit gives the RMS and largest residual over all runs.
    Size out gives the mean and largest absolute error over all runs, each predicted by the laws refitted without its backbone size.
    The 6.4B out columns give the same over the 6.4B runs, predicted by the laws fitted on 0.6B to 3.2B.
  }
  \label{tab:appx:size-ladder-gof}
  
\setlength{\tabcolsep}{4pt}
\begin{tabular}{@{}lrrrrrr@{}}
  \toprule
  & \multicolumn{2}{c}{Fit} & \multicolumn{2}{c}{Size out} & \multicolumn{2}{c}{6.4B out} \\
  \cmidrule(l{.5\tabcolsep}r{.5\tabcolsep}){2-3}\cmidrule(l{.5\tabcolsep}r{.5\tabcolsep}){4-5}\cmidrule(l{.5\tabcolsep}r{.5\tabcolsep}){6-7}
  Metric & \multicolumn{1}{c}{RMS} & \multicolumn{1}{c}{Max} & \multicolumn{1}{c}{Mean} & \multicolumn{1}{c}{Max} & \multicolumn{1}{c}{Mean} & \multicolumn{1}{c@{}}{Max} \\
  \midrule
  Avg.\ VQA (pts) & 0.93 & 2.27 & 1.03 & 3.40 & 0.95 & 1.78 \\
  Caption loss (\%) & 1.20 & 4.81 & 2.03 & 7.42 & 1.91 & 5.00 \\
  Text loss (\%) & 0.36 & 0.83 & 0.41 & 1.52 & 0.45 & 0.95 \\
  \bottomrule
\end{tabular}

\end{table}

\begin{table}[!htbp]
  \centering
  \footnotesize
  \caption{
    \textbf{The ladder's caption-loss laws predict the separately trained 1.7B and 4B runs within a few percent.}
    Each entry is the measured held-out caption loss divided by the loss that the recipe's law of \cref{tab:appx:size-ladder-fits} predicts, minus one, in percent.
  }
  \label{tab:appx:size-ladder-heldout}
  
\setlength{\tabcolsep}{4pt}
\begin{tabular}{@{}llrrrrrr@{}}
  \toprule
  & & \multicolumn{6}{c}{Token-equivalents} \\
  \cmidrule(l{.5\tabcolsep}){3-8}
  Backbone & Recipe & \multicolumn{1}{c}{111B} & \multicolumn{1}{c}{222B} & \multicolumn{1}{c}{444B} & \multicolumn{1}{c}{667B} & \multicolumn{1}{c}{889B} & \multicolumn{1}{c@{}}{1.11T} \\
  \midrule
  1.7B & Scratch encoder & $-$1.6 & $-$0.4 & $+$0.6 & $+$0.9 & $+$1.2 & $+$1.0 \\
   & Encoder-free & $-$2.7 & $-$1.3 & $+$0.9 & $+$1.4 & $+$1.9 & $+$2.1 \\
   & In-fusion & $-$0.8 & $-$0.7 & $-$0.5 & -- & -- & $-$0.6 \\
  \midrule
  4B & Scratch encoder & $-$4.3 & $-$1.5 & $+$0.9 & $+$3.0 & -- & $+$3.3 \\
   & Encoder-free & $-$5.0 & $-$4.0 & $-$1.9 & $-$1.7 & -- & $-$1.0 \\
   & In-fusion & $+$0.8 & $+$0.6 & $+$1.4 & -- & -- & $+$0.6 \\
  \bottomrule
\end{tabular}

\end{table}

\begin{figure}[!htbp]
  \centering
  \setlength{\parskip}{0pt}%
  \captionsetup[sub]{skip=0pt,justification=centering}%
  \includegraphics[width=\linewidth]{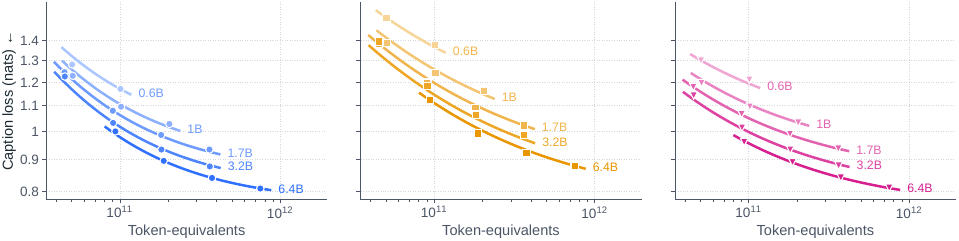}\par
  \nointerlineskip\vspace{1pt}%
  \hspace*{0.0489\linewidth}%
  \begin{minipage}[t]{0.2928\linewidth}
    \subcaption[Scratch encoder]{\textcolor{PaperBlue700}{Scratch encoder}}
    \label{fig:appx:size-ladder-laws:enc}
  \end{minipage}%
  \hspace*{0.0352\linewidth}%
  \begin{minipage}[t]{0.2928\linewidth}
    \subcaption[Encoder-free]{\textcolor{PaperAmber700}{Encoder-free}}
    \label{fig:appx:size-ladder-laws:ef}
  \end{minipage}%
  \hspace*{0.0351\linewidth}%
  \begin{minipage}[t]{0.2929\linewidth}
    \subcaption[In-fusion]{\textcolor{PaperPink700}{In-fusion}}
    \label{fig:appx:size-ladder-laws:inf}
  \end{minipage}%
  \hspace*{0.0023\linewidth}\par
  \caption{
    \textbf{A single caption-loss law per recipe fits the runs at every backbone size.}
    Points are the held-out caption loss of every ladder run.
    Each line is the recipe's law evaluated at one backbone size.
  }
  \label{fig:appx:size-ladder-laws}
\end{figure}

  \FloatBarrier

\section{Beyond lifetime compute: midtraining and inference}
\label{sec:appx:systems-full}

This section describes the full midtraining comparison of \cref{sec:capacity:late-onset} and the training and inference speed measurements behind \cref{sec:practical:serving}.

\subsection{Midtraining}
\label{sec:appx:midtraining}
\Cref{tab:appx:midtraining} provides the full midtraining comparison of \cref{sec:capacity:late-onset} with per-category metrics.
We also show the Avg.\ VQA of the three main recipes when trained with vision from the start.
Each recipe's schedule lowers the text mixture proportion from 100\% to 70\% over the first 20B tokens and then maintains the 70:30 mixture.
The pretrained MetaCLIP encoder stays frozen for the first 4,000 steps while its new linear projector and the backbone are updated.
It then unfreezes with a 2,000-step learning rate warmup.
Each midtraining run continues at a constant learning rate to 1.2T token-equivalents and then decays to 1.333T.
All midtraining runs use $1.56\times10^{22}$~FLOPs of lifetime compute, including the backbone's text-only pretraining but not MetaCLIP's encoder pretraining.
Vision-only parameters exclude an encoder's linear projection into the backbone and include the patch embedding of the other recipes.

\begin{table}[!htbp]
  \centering
  \small
  \caption{
    \textbf{In-fusion vision capacity and the from-scratch encoder maintain their VQA lead in midtraining.}
    From start shows VQA performance when training the same recipe with vision from the first step.
    Each recipe extends the 1.7B text-only model from 1T token-equivalents.
    Bold: within one point of the best.
  }
  \label{tab:capacity:late-onset}
  \label{tab:appx:midtraining}
  
\setlength{\tabcolsep}{2.5pt}
\renewcommand{\arraystretch}{1.05}
\begin{tabular}{@{}lrrrrrrrrrrrr@{}}
  \toprule
  \multicolumn{2}{>{\columncolor{PaperGray100!55}[.5\tabcolsep][.5\tabcolsep]}c}{\textbf{Recipe}} & \multicolumn{6}{>{\columncolor{PaperBlue100!55}[.5\tabcolsep][.5\tabcolsep]}c}{\textbf{VQA (\%) $\uparrow$}} & \multicolumn{5}{>{\columncolor{PaperGreen100!55}[.5\tabcolsep][.5\tabcolsep]}c}{\textbf{Text (\%) $\uparrow$}} \\
  \cmidrule(l{.5\tabcolsep}r{.5\tabcolsep}){1-2}\cmidrule(l{.5\tabcolsep}r{.5\tabcolsep}){3-8}\cmidrule(l{.5\tabcolsep}r{.5\tabcolsep}){9-13}
  Model & \makecell{Vision-only\\params (B)} & \multicolumn{1}{c}{Avg} & \multicolumn{1}{c}{Gen} & \multicolumn{1}{c}{Knw} & \multicolumn{1}{c}{OCR} & \multicolumn{1}{c}{Vis} & \multicolumn{1}{c}{\makecell{From\\start}} & \multicolumn{1}{c}{Avg} & \multicolumn{1}{c}{Com} & \multicolumn{1}{c}{Sci} & \multicolumn{1}{c}{Knw} & \multicolumn{1}{c@{}}{Rdg} \\
  \midrule
  \rowcolor{PaperGray100!50}
  \recipeglyph{scratch}Scratch encoder & 0.40 & 48.6 & 58.8 & 42.0 & 38.9 & \textbf{54.5} & \textbf{49.3} & \textbf{56.7} & \textbf{67.5} & \textbf{48.9} & 50.5 & \textbf{60.0} \\
  \recipeglyph{encoderfree}Encoder-free & $<$0.01 & 39.8 & 49.1 & 38.7 & 25.6 & 45.7 & 45.4 & \textbf{56.6} & \textbf{67.7} & \textbf{48.6} & \textbf{51.4} & 58.9 \\
  \recipeglyph{infusionqkvo}In-fusion & 0.40 & \textbf{50.1} & \textbf{61.0} & \textbf{43.1} & \textbf{41.0} & \textbf{55.2} & \textbf{49.2} & \textbf{56.4} & 66.7 & 47.8 & \textbf{51.0} & \textbf{60.2} \\
  \recipeglyph{}MoT, layers 1--8 & 0.40 & 48.5 & \textbf{60.1} & 41.8 & 39.5 & 52.5 & -- & \textbf{56.7} & 66.8 & \textbf{48.8} & \textbf{51.6} & 59.4 \\
  \recipeglyph{}MoT, all layers & 1.41 & 43.3 & 50.9 & 39.3 & 38.2 & 44.8 & -- & \textbf{56.6} & \textbf{67.4} & \textbf{48.0} & \textbf{50.9} & \textbf{59.9} \\
  \recipeglyph{metaclip}MetaCLIP (``free'') & 0.40 & 43.5 & 55.2 & 39.7 & 29.3 & 49.8 & -- & \textbf{56.5} & \textbf{67.3} & \textbf{48.0} & \textbf{50.9} & \textbf{59.6} \\
  \arrayrulecolor{PaperGray400}\cmidrule{1-13}\arrayrulecolor{black}
  \recipeglyph{}Text-only control & -- & -- & -- & -- & -- & -- & -- & \textbf{57.3} & \textbf{68.0} & \textbf{48.9} & \textbf{51.6} & \textbf{60.5} \\
  \bottomrule
\end{tabular}

\end{table}

\paragraph{Vision learning rate.}
\label{sec:appx:midtraining-lr}
\Cref{tab:appx:midtraining} uses the same peak learning rate for every parameter.
\Cref{tab:appx:midtraining-lr} instead adjusts the learning rate of the vision path alone, while every other parameter keeps the shared rate.
A higher vision learning rate brings the from-scratch encoder to within a point of in-fusion and partially closes MetaCLIP's gap to the from-scratch encoder, but the from-scratch encoder and in-fusion still lead.

\begin{table}[!htbp]
  \centering
  \small
  \caption{
    \textbf{Tuning the encoder learning rate does not impact midtraining findings.} 
    LR multiple scales the learning rate of each recipe's respective vision pathway, relative to the shared peak learning rate ($3\times10^{-4}$; \cref{tab:appx:hyperparameters}).
    Bold: within one point of the best on VQA.
    Avg.\ Text varies by only 0.6 points across all rows, so Text metrics are not bolded.
  }
  \label{tab:appx:midtraining-lr}
  
\setlength{\tabcolsep}{2.5pt}
\renewcommand{\arraystretch}{1.05}
\begin{tabular}{@{}lcrrrrrrrrrr@{}}
  \toprule
  \multicolumn{2}{>{\columncolor{PaperGray100!55}[.5\tabcolsep][.5\tabcolsep]}c}{\textbf{Recipe}} & \multicolumn{5}{>{\columncolor{PaperBlue100!55}[.5\tabcolsep][.5\tabcolsep]}c}{\textbf{VQA (\%) $\uparrow$}} & \multicolumn{5}{>{\columncolor{PaperGreen100!55}[.5\tabcolsep][.5\tabcolsep]}c}{\textbf{Text (\%) $\uparrow$}} \\
  \cmidrule(l{.5\tabcolsep}r{.5\tabcolsep}){1-2}\cmidrule(l{.5\tabcolsep}r{.5\tabcolsep}){3-7}\cmidrule(l{.5\tabcolsep}r{.5\tabcolsep}){8-12}
  Model & \makecell{LR\\multiple} & \multicolumn{1}{c}{Avg} & \multicolumn{1}{c}{Gen} & \multicolumn{1}{c}{Knw} & \multicolumn{1}{c}{OCR} & \multicolumn{1}{c}{Vis} & \multicolumn{1}{c}{Avg} & \multicolumn{1}{c}{Com} & \multicolumn{1}{c}{Sci} & \multicolumn{1}{c}{Knw} & \multicolumn{1}{c@{}}{Rdg} \\
  \midrule
  \rowcolor{PaperGray100!50}
  \recipeglyph{scratch}Scratch encoder & 1$\times$ & 48.6 & 58.8 & 42.0 & 38.9 & \textbf{54.5} & 56.7 & 67.5 & 48.9 & 50.5 & 60.0 \\
   & 3$\times$ & \textbf{49.4} & \textbf{61.4} & 41.4 & \textbf{42.1} & 52.5 & 56.3 & 66.8 & 48.0 & 50.9 & 59.3 \\
  \arrayrulecolor{PaperGray400}\cmidrule{1-12}\arrayrulecolor{black}
  \recipeglyph{infusionqkvo}In-fusion & 1$\times$ & \textbf{50.1} & \textbf{61.0} & 43.1 & 41.0 & \textbf{55.2} & 56.4 & 66.7 & 47.8 & 51.0 & 60.2 \\
   & 3$\times$ & \textbf{49.6} & \textbf{60.5} & \textbf{44.3} & 40.2 & 53.6 & 56.8 & 67.5 & 48.7 & 51.4 & 59.6 \\
  \arrayrulecolor{PaperGray400}\cmidrule{1-12}\arrayrulecolor{black}
  \recipeglyph{}MoT, layers 1--8 & 1$\times$ & 48.5 & 60.1 & 41.8 & 39.5 & 52.5 & 56.7 & 66.8 & 48.8 & 51.6 & 59.4 \\
   & 3$\times$ & \textbf{49.2} & \textbf{60.4} & 43.1 & 39.4 & 53.7 & 56.8 & 67.2 & 48.8 & 51.0 & 60.1 \\
  \arrayrulecolor{PaperGray400}\cmidrule{1-12}\arrayrulecolor{black}
  \recipeglyph{}MoT, all layers & 1$\times$ & 43.3 & 50.9 & 39.3 & 38.2 & 44.8 & 56.6 & 67.4 & 48.0 & 50.9 & 59.9 \\
   & 3$\times$ & 45.1 & 53.6 & 41.3 & 37.6 & 47.9 & 56.5 & 66.8 & 48.0 & 51.4 & 59.9 \\
  \arrayrulecolor{PaperGray400}\cmidrule{1-12}\arrayrulecolor{black}
  \recipeglyph{metaclip}MetaCLIP (``free'') & 0.3$\times$ & 38.8 & 46.6 & 39.2 & 23.9 & 45.7 & 56.7 & 67.0 & 48.8 & 51.5 & 59.7 \\
   & 1$\times$ & 43.5 & 55.2 & 39.7 & 29.3 & 49.8 & 56.5 & 67.3 & 48.0 & 50.9 & 59.6 \\
   & 3$\times$ & 46.1 & 57.9 & 40.2 & 35.4 & 50.7 & 56.6 & 67.1 & 48.6 & 50.7 & 59.8 \\
   & 10$\times$ & 42.9 & 52.9 & 39.9 & 28.1 & 50.7 & 56.5 & 67.6 & 47.9 & 50.7 & 60.0 \\
  \bottomrule
\end{tabular}

\end{table}

\subsection{Training time}
\label{sec:appx:systems-attribution}
\label{sec:appx:training-benchmark}

Here, we compare the training time of the encoder recipe with encoder-free, \oursffn, and \oursqkvo.
Training time can be compared in two ways: per-step comparisons fix the number of optimizer updates, whereas equal-FLOP comparisons fix total modeled training compute.
Because encoder computation counts toward that budget, the encoder recipe completes fewer updates at equal FLOPs, so its equal-FLOP time premium is smaller than its per-step premium.

We measure training time with complete optimizer updates on H200 GPUs with fully sharded data parallelism (FSDP)~\citep{zhao2023fsdp}.
The matched 4B comparison in \cref{tab:appx:capacity:systems-4b} runs the four recipes on 32 GPUs with a global batch of 1,024 sequences.
Each row projects training time from measured steady-state update times to the same modeled lifetime compute, $1.04\times10^{22}$ FLOPs (400B token-equivalents at 4B; \cref{sec:appx:compute-conversion}).

\Cref{tab:appx:capacity:systems-4b} times the encoder and the in-fusion configurations of \cref{tab:capacity:qkvo}, which match the encoder's 0.4B vision-only parameters, under the 4B training setup above.
At equal lifetime compute, neither in-fusion variant measurably changes training time relative to encoder-free.
The encoder takes 2.9\% longer than encoder-free and 2.6\% longer than \oursqkvo at equal FLOPs.
Per step, the encoder takes 5.3\% longer than encoder-free.
Note that the encoder used here is an unoptimized reference implementation, and encoder-specific optimization can narrow its gap.
For instance, a compiled~\citep{ansel2024pytorch2} version of the Base encoder in \cref{tab:architecture:encoder-architecture} trains within 0.2\% of an \oursffn model with the same vision-only parameters at 4B.

\paragraph{Encoder pipelining.}
Overlapping or pipelining~\citep{huang2019gpipe} the encoder recovers little of its cost.
At 4B, running the next microbatch's encoder forward pass alongside the backbone hides 20--36\% of the encoder's per-step cost.
Only the forward pass overlaps, because the encoder's backward pass waits for the backbone's gradients.
With or without an encoder, a four-stage pipeline has about half the throughput of the encoder recipe under FSDP on the same 32 GPUs.
FSDP optimizations applied to the encoder alone, such as prefetching parameters or skipping gradient synchronization between accumulation steps, change step time by less than 0.7\%.

\begin{table}[!htbp]
  \centering
  \small
  \caption{
    \textbf{In-fusion vision capacity trains as fast as encoder-free, while the encoder-based model takes longer.}
    Training time is for a 4B model projected to $1.04\times10^{22}$~FLOPs of lifetime compute (400B token-equivalents).
    Intervals are paired 95\% bootstrap intervals over 17 common timing windows.
  }
  \label{tab:appx:capacity:systems-4b}
  
\setlength{\tabcolsep}{6pt}
\renewcommand{\arraystretch}{1.05}
\begin{tabular}{@{}lrc@{}}
  \toprule
  \multicolumn{1}{>{\columncolor{PaperGray100!55}[.5\tabcolsep][.5\tabcolsep]}c}{\textbf{Architecture}} & \multicolumn{2}{>{\columncolor{PaperGray100!55}[.5\tabcolsep][.5\tabcolsep]}c}{\textbf{Training time}} \\
  \cmidrule(l{.5\tabcolsep}r{.5\tabcolsep}){1-1}\cmidrule(l{.5\tabcolsep}r{.5\tabcolsep}){2-3}
  Variant & $\Delta$ vs.\ encoder-free (\%) & 95\% CI \\
  \midrule
  Encoder & $+2.9$ & $[+1.9, +4.2]$ \\
  \oursffnstyled{} & $-0.3$ & $[-0.8, +0.4]$ \\
  \oursqkvostyled{} & $+0.3$ & $[-0.5, +1.1]$ \\
  \bottomrule
\end{tabular}

\end{table}

\paragraph{Visual workload.}
Per step, encoder-free and \oursffn both train faster than the encoder recipe.
Encoder-free saves more because the vision-only FFNs of \oursffn cost about 95\% of the encoder's FLOPs.
In an 8-GPU setup at 1.7B, raising the proportion of image--text data from 10\% to 50\% widens encoder-free's saving from 10.0\% to 13.2\%, while \oursffn saves 6--7\% throughout (\cref{fig:encoder-free:efficiency}).

\begin{figure}[!htbp]
  \centering
  \captionsetup[sub]{skip=1pt,justification=centering}%
  \begin{minipage}[t]{0.3333\linewidth}
    \centering
    \includegraphics[width=\linewidth]{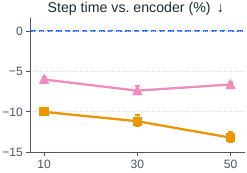}
    \subcaption*{Image--text sequences (\%)}
  \end{minipage}\par\vspace{2pt}
  \includegraphics[scale=1.192]{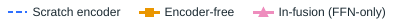}
  \caption{
    \textbf{Encoder-free and \oursffn reduce training step time relative to encoder.}
    All runs use the 1.7B backbone.
    Intervals are paired 95\% bootstrap intervals.
  }\label{fig:encoder-free:efficiency}\label{fig:encoder-free:efficiency:ratio}
\end{figure}

\FloatBarrier
\subsection{Inference latency}
\label{sec:appx:inference-latency}
\label{sec:appx:inference-benchmark}

\Cref{sec:practical:serving} compares how fast each recipe serves image requests.
We run every design in our own implementation on one H200, so all designs share the same inference code.
Each request has one image and 256 text tokens and generates 120 tokens.
Largely following the inference benchmark setup of Mono-InternVL~\citep{luo2025monointernvl}, we report time to first token (TTFT) and output throughput with 16 concurrent requests and 448-pixel images (1,024 vision tokens) unless noted.
Each configuration is measured in four runs, and we report medians with full ranges.

Without a separate encoder, the backbone no longer waits for each image to be encoded, so TTFT falls (\cref{fig:appx:ttft-resolution}).
\oursffnstyled still processes every vision token with its vision-only FFNs, and at matched vision-only parameters, it performs approximately the same computation per image as the encoder.
Its advantage comes from running in eight wide layers instead of the encoder's deeper stack of narrower layers (27 at the 0.4B budget), in our eager-mode implementation.
The inference time saving grows with resolution, because more vision tokens make the encoder a larger share of the time to first token.

\begin{figure}[!htbp]
  \centering
  \includegraphics[width=0.40\linewidth]{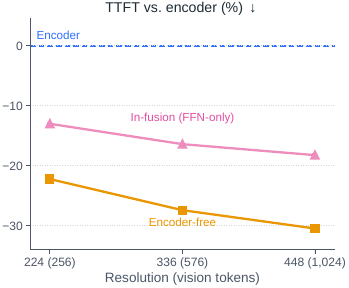}
  \caption{
    \textbf{Both encoder-free and \oursffn reduce TTFT at higher resolution.}
    TTFT is measured relative to the encoder at 1.7B with 16 concurrent requests, where intervals are full ranges over four runs.
  }
  \label{fig:appx:ttft-resolution}
\end{figure}

\begin{table}[!htbp]
  \centering
  \small
  \caption{
    \textbf{In-fusion capacity serves image requests faster than the encoder at every vision budget.}
    Encoder-free takes 146~ms at 1.7B and 340~ms at 4B, 31\% and 16\% less than the Large encoder.
    Each request has one 448-pixel image (1,024 vision tokens) and 256 text tokens, served on one H200.
    Budgets describe the encoder's vision-only parameters, which the in-fusion modules match.
    TTFT change relative to the scratch encoder is shown in parentheses.
  }
  \label{tab:appx:ttft-fixed-backbone}
  
\setlength{\tabcolsep}{5pt}
\renewcommand{\arraystretch}{1.05}
\begin{tabular}{@{}lrrr@{}}
  \toprule
  & \multicolumn{3}{c@{}}{TTFT with 16 requests (ms) $\downarrow$} \\
  \cmidrule(l{.5\tabcolsep}){2-4}
  Budget & \multicolumn{1}{c}{\recipeglyph{scratch}Scratch encoder} & \multicolumn{1}{c}{\recipeglyph{infusionffn}In-fusion (FFN-only)} & \multicolumn{1}{c@{}}{\recipeglyph{infusionqkvo}In-fusion} \\
  \midrule
  \multicolumn{4}{@{}l}{\textit{1.7B backbone}} \\
  Small (46M) & 158 & 152\,{\scriptsize($-$4\%)} & 154\,{\scriptsize($-$3\%)} \\
  Base (152M) & 176 & 157\,{\scriptsize($-$11\%)} & 169\,{\scriptsize($-$4\%)} \\
  \rowcolor{PaperGray100!50}
  Large (401M) & 210 & 171\,{\scriptsize($-$18\%)} & 182\,{\scriptsize($-$13\%)} \\
  Huge (924M) & 268 & 196\,{\scriptsize($-$27\%)} & 206\,{\scriptsize($-$23\%)} \\
  Wide (1.26B) & 273 & 212\,{\scriptsize($-$22\%)} & 219\,{\scriptsize($-$19\%)} \\
  \arrayrulecolor{PaperGray400}\cmidrule{1-4}\arrayrulecolor{black}
  \multicolumn{4}{@{}l}{\textit{4B backbone}} \\
  Base (152M) & 374 & 354\,{\scriptsize($-$5\%)} & 361\,{\scriptsize($-$3\%)} \\
  \rowcolor{PaperGray100!50}
  Large (401M) & 405 & 365\,{\scriptsize($-$10\%)} & 381\,{\scriptsize($-$6\%)} \\
  Huge (924M) & 464 & 392\,{\scriptsize($-$16\%)} & 405\,{\scriptsize($-$13\%)} \\
  \bottomrule
\end{tabular}

\end{table}

\paragraph{Varying the vision budget.}
We also try varying the vision-only parameters across the encoder sizes of \cref{sec:encoders:encoder-architecture}, with both in-fusion variants sized to match each encoder (\cref{tab:appx:ttft-fixed-backbone}).
In-fusion is faster than the encoder at every budget.
The time that vision adds over encoder-free is set by the vision budget rather than the backbone, so \oursffn saves as much time at 4B as at 1.7B.
Its relative inference time saving is smaller at 4B only because the backbone's own prefill time is longer.

  \FloatBarrier
  
\clearpage
\section{Per-benchmark results}
\label{sec:appx:per-benchmark}

This section reports every VQA and text benchmark of \cref{tab:appx:vqa-benchmarks,tab:appx:language-benchmarks} for the runs that each main-text exhibit shows, one table per exhibit.
Avg.\ VQA and Avg.\ Text are computed as in \cref{sec:appx:evaluation}, and FLOPs is lifetime compute.
Cap.\ is the held-out caption loss on regenerated captions of MetaCLIP 1.2 images never used in training, and Text the held-out text loss on unseen Nemotron-CC text weighted as in the training mixture (\cref{sec:appx:perplexity}); both are in nats per token and measured before VQA instruction tuning.

\paragraph{Column headers.}
\begin{center}
  
\small
\setlength{\tabcolsep}{4.0pt}
\renewcommand{\arraystretch}{1.08}
\begin{tabular}[t]{@{}ll@{}}
  \toprule
  Header & VQA benchmark \\
  \midrule
  \multicolumn{2}{@{}l}{\textit{General (Gen)}} \\
  MME-P & MME-P \\
  MMB & MMBench-EN \\
  SEED & SEED-Bench \\
  GQA & GQA \\
  \arrayrulecolor{PaperGray400}\cmidrule{1-2}\arrayrulecolor{black}
  \multicolumn{2}{@{}l}{\textit{Knowledge (Knw)}} \\
  SQA & ScienceQA (image) \\
  MMMU & MMMU \\
  MathV & MathVista \\
  AI2D & AI2D \\
  \arrayrulecolor{PaperGray400}\cmidrule{1-2}\arrayrulecolor{black}
  \multicolumn{2}{@{}l}{\textit{OCR and chart (OCR)}} \\
  Chart & ChartQA \\
  OCRB & OCRBench \\
  TextV & TextVQA \\
  DocV & DocVQA \\
  \arrayrulecolor{PaperGray400}\cmidrule{1-2}\arrayrulecolor{black}
  \multicolumn{2}{@{}l}{\textit{Vision-centric (Vis)}} \\
  MMVP & MMVP \\
  RWQA & RealWorldQA \\
  CV2D & CV-Bench-2D \\
  CV3D & CV-Bench-3D \\
  \bottomrule
\end{tabular}
\hspace{3em}%
\begin{tabular}[t]{@{}ll@{}}
  \toprule
  Header & Text benchmark \\
  \midrule
  \multicolumn{2}{@{}l}{\textit{Commonsense reasoning (Com)}} \\
  COPA & COPA \\
  HS & HellaSwag \\
  PIQA & PIQA \\
  SIQA & Social IQa \\
  WG & WinoGrande \\
  \arrayrulecolor{PaperGray400}\cmidrule{1-2}\arrayrulecolor{black}
  \multicolumn{2}{@{}l}{\textit{Science question answering (Sci)}} \\
  ARC-E & ARC-Easy \\
  ARC-C & ARC-Challenge \\
  OBQA & OpenBookQA \\
  \arrayrulecolor{PaperGray400}\cmidrule{1-2}\arrayrulecolor{black}
  \multicolumn{2}{@{}l}{\textit{Broad knowledge (Knw)}} \\
  MMLU & MMLU (57 subjects) \\
  \arrayrulecolor{PaperGray400}\cmidrule{1-2}\arrayrulecolor{black}
  \multicolumn{2}{@{}l}{\textit{Reading comprehension (Rdg)}} \\
  BoolQ & BoolQ \\
  RACEm & RACE-middle \\
  RACEh & RACE-high \\
  \bottomrule
\end{tabular}

\end{center}

\paragraph{Tables.}
\begin{center}
  
\small
\setlength{\tabcolsep}{4.0pt}
\renewcommand{\arraystretch}{1.08}
\begin{tabular}{@{}rllr@{}}
  \toprule
  Table & Main-text exhibit & Comparison & Page \\
  \midrule
  \ref{tab:appx:per-benchmark:upstream-share} & \cref{fig:encoders:upstream:a} & Encoder pretraining shares of a fixed lifetime budget & \pageref{tab:appx:per-benchmark:upstream-share} \\
  \ref{tab:appx:per-benchmark:upstream-lifetime} & \cref{fig:encoders:upstream:b} & From-scratch versus MetaCLIP-pretrained encoders across lifetime compute & \pageref{tab:appx:per-benchmark:upstream-lifetime} \\
  \ref{tab:appx:per-benchmark:encoder-free-gap} & \cref{fig:encoder-free:gap} & The from-scratch encoder versus encoder-free across lifetime compute & \pageref{tab:appx:per-benchmark:encoder-free-gap} \\
  \ref{tab:appx:per-benchmark:steelman} & \cref{tab:encoder-free:steelman} & Encoder-free backbones enlarged by the encoder's parameters & \pageref{tab:appx:per-benchmark:steelman} \\
  \ref{tab:appx:per-benchmark:design} & \cref{tab:capacity:design} & Placements of added in-fusion capacity & \pageref{tab:appx:per-benchmark:design} \\
  \ref{tab:appx:per-benchmark:qkvo-scale} & \cref{fig:capacity:qkvo-scale} & \oursqkvostyled{} versus \oursffn{} across lifetime compute & \pageref{tab:appx:per-benchmark:qkvo-scale} \\
  \ref{tab:appx:per-benchmark:trajectory-infusion} & \cref{fig:scaling:trajectory-infusion} & The from-scratch encoder, encoder-free, and \oursqkvostyled{} across lifetime compute & \pageref{tab:appx:per-benchmark:trajectory-infusion} \\
  \ref{tab:appx:per-benchmark:sizes} & \cref{fig:capacity:sizes} & Recipes across backbone sizes at 64 TPP & \pageref{tab:appx:per-benchmark:sizes} \\
  \ref{tab:appx:per-benchmark:free-pretraining} & \cref{fig:practical:free-pretraining} & Free MetaCLIP pretraining budgets before a fixed multimodal stage & \pageref{tab:appx:per-benchmark:free-pretraining} \\
  \ref{tab:appx:per-benchmark:midtraining} & \cref{fig:capacity:late-onset} & Recipes midtrained from a 1.7B text-only model & \pageref{tab:appx:per-benchmark:midtraining} \\
  \bottomrule
\end{tabular}

\end{center}

\begin{landscape}
\makeatletter
\newcommand{\appxstackedpage}{\setlength{\@fptop}{0pt}\setlength{\@fpsep}{10pt}\setlength{\@fpbot}{0pt plus 1fil}}
\newcommand{\appxcenteredpage}{\setlength{\@fptop}{0pt plus 1fil}\setlength{\@fpbot}{0pt plus 1fil}}
\makeatother

\appxstackedpage
\begin{table}[p]
  \centering
  \caption{
    \textbf{Per-benchmark results for \cref{fig:encoders:upstream:a}.}
    All runs use a 1.7B backbone and a $1.04\times10^{22}$-FLOP lifetime budget, of which Share is spent on encoder pretraining.
  }
  \label{tab:appx:per-benchmark:upstream-share}
  
\fontsize{6}{7}\selectfont
\setlength{\tabcolsep}{1.8pt}
\renewcommand{\arraystretch}{1.0}
\begin{tabular*}{\linewidth}{@{\extracolsep{\fill}}lrrrrrrrrrrrrrrrrrrrrrrrrrrrrrrrrr@{}}
  \toprule
  \multicolumn{2}{>{\columncolor{PaperGray100!55}[.5\tabcolsep][.5\tabcolsep]}c}{\textbf{Setup}} & \multicolumn{17}{>{\columncolor{PaperBlue100!55}[.5\tabcolsep][.5\tabcolsep]}c}{\textbf{VQA (\%) $\uparrow$}} & \multicolumn{13}{>{\columncolor{PaperGreen100!55}[.5\tabcolsep][.5\tabcolsep]}c}{\textbf{Text (\%) $\uparrow$}} & \multicolumn{2}{>{\columncolor{PaperGray100!55}[.5\tabcolsep][.5\tabcolsep]}c}{\textbf{Held-out loss $\downarrow$}} \\
  \cmidrule(l{.5\tabcolsep}r{.5\tabcolsep}){1-2}\cmidrule(l{.5\tabcolsep}r{.5\tabcolsep}){3-19}\cmidrule(l{.5\tabcolsep}r{.5\tabcolsep}){20-32}\cmidrule(l{.5\tabcolsep}r{.5\tabcolsep}){33-34}
   &  &  & \multicolumn{4}{c}{Gen} & \multicolumn{4}{c}{Knw} & \multicolumn{4}{c}{OCR} & \multicolumn{4}{c}{Vis} &  & \multicolumn{5}{c}{Com} & \multicolumn{3}{c}{Sci} & \multicolumn{1}{c}{Knw} & \multicolumn{3}{c}{Rdg} &  \\
\bottomrule
\end{tabular*}

\end{table}

\begin{table}[p]
  \centering
  \caption{
    \textbf{Per-benchmark results for \cref{fig:encoders:upstream:b}.}
    MetaCLIP's lifetime FLOPs include its $1.54\times10^{21}$-FLOP encoder pretraining.
  }
  \label{tab:appx:per-benchmark:upstream-lifetime}
  
\fontsize{6}{7}\selectfont
\setlength{\tabcolsep}{1.8pt}
\renewcommand{\arraystretch}{1.0}
\begin{tabular*}{\linewidth}{@{\extracolsep{\fill}}lrrrrrrrrrrrrrrrrrrrrrrrrrrrrrrrrr@{}}
  \toprule
  \multicolumn{2}{>{\columncolor{PaperGray100!55}[.5\tabcolsep][.5\tabcolsep]}c}{\textbf{Setup}} & \multicolumn{17}{>{\columncolor{PaperBlue100!55}[.5\tabcolsep][.5\tabcolsep]}c}{\textbf{VQA (\%) $\uparrow$}} & \multicolumn{13}{>{\columncolor{PaperGreen100!55}[.5\tabcolsep][.5\tabcolsep]}c}{\textbf{Text (\%) $\uparrow$}} & \multicolumn{2}{>{\columncolor{PaperGray100!55}[.5\tabcolsep][.5\tabcolsep]}c}{\textbf{Held-out loss $\downarrow$}} \\
  \cmidrule(l{.5\tabcolsep}r{.5\tabcolsep}){1-2}\cmidrule(l{.5\tabcolsep}r{.5\tabcolsep}){3-19}\cmidrule(l{.5\tabcolsep}r{.5\tabcolsep}){20-32}\cmidrule(l{.5\tabcolsep}r{.5\tabcolsep}){33-34}
   &  &  & \multicolumn{4}{c}{Gen} & \multicolumn{4}{c}{Knw} & \multicolumn{4}{c}{OCR} & \multicolumn{4}{c}{Vis} &  & \multicolumn{5}{c}{Com} & \multicolumn{3}{c}{Sci} & \multicolumn{1}{c}{Knw} & \multicolumn{3}{c}{Rdg} &  \\
\bottomrule
\end{tabular*}

\end{table}
\clearpage

\appxstackedpage
\begin{table}[p]
  \centering
  \caption{
    \textbf{Per-benchmark results for \cref{fig:encoder-free:gap}.}
  }
  \label{tab:appx:per-benchmark:encoder-free-gap}
  
\fontsize{6}{7}\selectfont
\setlength{\tabcolsep}{1.8pt}
\renewcommand{\arraystretch}{1.0}
\begin{tabular*}{\linewidth}{@{\extracolsep{\fill}}lrrrrrrrrrrrrrrrrrrrrrrrrrrrrrrrrr@{}}
  \toprule
  \multicolumn{2}{>{\columncolor{PaperGray100!55}[.5\tabcolsep][.5\tabcolsep]}c}{\textbf{Setup}} & \multicolumn{17}{>{\columncolor{PaperBlue100!55}[.5\tabcolsep][.5\tabcolsep]}c}{\textbf{VQA (\%) $\uparrow$}} & \multicolumn{13}{>{\columncolor{PaperGreen100!55}[.5\tabcolsep][.5\tabcolsep]}c}{\textbf{Text (\%) $\uparrow$}} & \multicolumn{2}{>{\columncolor{PaperGray100!55}[.5\tabcolsep][.5\tabcolsep]}c}{\textbf{Held-out loss $\downarrow$}} \\
  \cmidrule(l{.5\tabcolsep}r{.5\tabcolsep}){1-2}\cmidrule(l{.5\tabcolsep}r{.5\tabcolsep}){3-19}\cmidrule(l{.5\tabcolsep}r{.5\tabcolsep}){20-32}\cmidrule(l{.5\tabcolsep}r{.5\tabcolsep}){33-34}
   &  &  & \multicolumn{4}{c}{Gen} & \multicolumn{4}{c}{Knw} & \multicolumn{4}{c}{OCR} & \multicolumn{4}{c}{Vis} &  & \multicolumn{5}{c}{Com} & \multicolumn{3}{c}{Sci} & \multicolumn{1}{c}{Knw} & \multicolumn{3}{c}{Rdg} &  \\
\bottomrule
\end{tabular*}

\end{table}

\begin{table}[p]
  \centering
  \caption{
    \textbf{Per-benchmark results for \cref{tab:encoder-free:steelman}.}
    All rows use a 1.7B backbone and $2.6\times10^{21}$ lifetime FLOPs.
  }
  \label{tab:appx:per-benchmark:steelman}
  
\fontsize{6}{7}\selectfont
\setlength{\tabcolsep}{1.8pt}
\renewcommand{\arraystretch}{1.0}
\begin{tabular*}{\linewidth}{@{\extracolsep{\fill}}lrrrrrrrrrrrrrrrrrrrrrrrrrrrrrrrr@{}}
  \toprule
  \multicolumn{1}{>{\columncolor{PaperGray100!55}[.5\tabcolsep][.5\tabcolsep]}c}{\textbf{Architecture}} & \multicolumn{17}{>{\columncolor{PaperBlue100!55}[.5\tabcolsep][.5\tabcolsep]}c}{\textbf{VQA (\%) $\uparrow$}} & \multicolumn{13}{>{\columncolor{PaperGreen100!55}[.5\tabcolsep][.5\tabcolsep]}c}{\textbf{Text (\%) $\uparrow$}} & \multicolumn{2}{>{\columncolor{PaperGray100!55}[.5\tabcolsep][.5\tabcolsep]}c}{\textbf{Held-out loss $\downarrow$}} \\
  \cmidrule(l{.5\tabcolsep}r{.5\tabcolsep}){1-1}\cmidrule(l{.5\tabcolsep}r{.5\tabcolsep}){2-18}\cmidrule(l{.5\tabcolsep}r{.5\tabcolsep}){19-31}\cmidrule(l{.5\tabcolsep}r{.5\tabcolsep}){32-33}
   &  & \multicolumn{4}{c}{Gen} & \multicolumn{4}{c}{Knw} & \multicolumn{4}{c}{OCR} & \multicolumn{4}{c}{Vis} &  & \multicolumn{5}{c}{Com} & \multicolumn{3}{c}{Sci} & \multicolumn{1}{c}{Knw} & \multicolumn{3}{c}{Rdg} &  \\
\bottomrule
\end{tabular*}

\end{table}
\clearpage

\appxstackedpage
\begin{table}[p]
  \centering
  \caption{
    \textbf{Per-benchmark results for \cref{tab:capacity:design}.}
    All rows use a 1.7B backbone and $2.6\times10^{21}$ lifetime FLOPs.
  }
  \label{tab:appx:per-benchmark:design}
  
\fontsize{6}{7}\selectfont
\setlength{\tabcolsep}{1.8pt}
\renewcommand{\arraystretch}{1.0}
\begin{tabular*}{\linewidth}{@{\extracolsep{\fill}}lrrrrrrrrrrrrrrrrrrrrrrrrrrrrrrrr@{}}
  \toprule
  \multicolumn{1}{>{\columncolor{PaperGray100!55}[.5\tabcolsep][.5\tabcolsep]}c}{\textbf{Architecture}} & \multicolumn{17}{>{\columncolor{PaperBlue100!55}[.5\tabcolsep][.5\tabcolsep]}c}{\textbf{VQA (\%) $\uparrow$}} & \multicolumn{13}{>{\columncolor{PaperGreen100!55}[.5\tabcolsep][.5\tabcolsep]}c}{\textbf{Text (\%) $\uparrow$}} & \multicolumn{2}{>{\columncolor{PaperGray100!55}[.5\tabcolsep][.5\tabcolsep]}c}{\textbf{Held-out loss $\downarrow$}} \\
  \cmidrule(l{.5\tabcolsep}r{.5\tabcolsep}){1-1}\cmidrule(l{.5\tabcolsep}r{.5\tabcolsep}){2-18}\cmidrule(l{.5\tabcolsep}r{.5\tabcolsep}){19-31}\cmidrule(l{.5\tabcolsep}r{.5\tabcolsep}){32-33}
   &  & \multicolumn{4}{c}{Gen} & \multicolumn{4}{c}{Knw} & \multicolumn{4}{c}{OCR} & \multicolumn{4}{c}{Vis} &  & \multicolumn{5}{c}{Com} & \multicolumn{3}{c}{Sci} & \multicolumn{1}{c}{Knw} & \multicolumn{3}{c}{Rdg} &  \\
\bottomrule
\end{tabular*}

\end{table}

\begin{table}[p]
  \centering
  \caption{
    \textbf{Per-benchmark results for \cref{fig:capacity:qkvo-scale}.}
  }
  \label{tab:appx:per-benchmark:qkvo-scale}
  
\fontsize{6}{7}\selectfont
\setlength{\tabcolsep}{1.8pt}
\renewcommand{\arraystretch}{1.0}
\begin{tabular*}{\linewidth}{@{\extracolsep{\fill}}lrrrrrrrrrrrrrrrrrrrrrrrrrrrrrrrrr@{}}
  \toprule
  \multicolumn{2}{>{\columncolor{PaperGray100!55}[.5\tabcolsep][.5\tabcolsep]}c}{\textbf{Setup}} & \multicolumn{17}{>{\columncolor{PaperBlue100!55}[.5\tabcolsep][.5\tabcolsep]}c}{\textbf{VQA (\%) $\uparrow$}} & \multicolumn{13}{>{\columncolor{PaperGreen100!55}[.5\tabcolsep][.5\tabcolsep]}c}{\textbf{Text (\%) $\uparrow$}} & \multicolumn{2}{>{\columncolor{PaperGray100!55}[.5\tabcolsep][.5\tabcolsep]}c}{\textbf{Held-out loss $\downarrow$}} \\
  \cmidrule(l{.5\tabcolsep}r{.5\tabcolsep}){1-2}\cmidrule(l{.5\tabcolsep}r{.5\tabcolsep}){3-19}\cmidrule(l{.5\tabcolsep}r{.5\tabcolsep}){20-32}\cmidrule(l{.5\tabcolsep}r{.5\tabcolsep}){33-34}
   &  &  & \multicolumn{4}{c}{Gen} & \multicolumn{4}{c}{Knw} & \multicolumn{4}{c}{OCR} & \multicolumn{4}{c}{Vis} &  & \multicolumn{5}{c}{Com} & \multicolumn{3}{c}{Sci} & \multicolumn{1}{c}{Knw} & \multicolumn{3}{c}{Rdg} &  \\
\bottomrule
\end{tabular*}

\end{table}
\clearpage

\appxcenteredpage
\begin{table}[p]
  \centering
  \caption{
    \textbf{Per-benchmark results for \cref{fig:scaling:trajectory-infusion}.}
  }
  \label{tab:appx:per-benchmark:trajectory-infusion}
  
\fontsize{6}{7}\selectfont
\setlength{\tabcolsep}{1.8pt}
\renewcommand{\arraystretch}{1.0}
\begin{tabular*}{\linewidth}{@{\extracolsep{\fill}}lrrrrrrrrrrrrrrrrrrrrrrrrrrrrrrrrr@{}}
  \toprule
  \multicolumn{2}{>{\columncolor{PaperGray100!55}[.5\tabcolsep][.5\tabcolsep]}c}{\textbf{Setup}} & \multicolumn{17}{>{\columncolor{PaperBlue100!55}[.5\tabcolsep][.5\tabcolsep]}c}{\textbf{VQA (\%) $\uparrow$}} & \multicolumn{13}{>{\columncolor{PaperGreen100!55}[.5\tabcolsep][.5\tabcolsep]}c}{\textbf{Text (\%) $\uparrow$}} & \multicolumn{2}{>{\columncolor{PaperGray100!55}[.5\tabcolsep][.5\tabcolsep]}c}{\textbf{Held-out loss $\downarrow$}} \\
  \cmidrule(l{.5\tabcolsep}r{.5\tabcolsep}){1-2}\cmidrule(l{.5\tabcolsep}r{.5\tabcolsep}){3-19}\cmidrule(l{.5\tabcolsep}r{.5\tabcolsep}){20-32}\cmidrule(l{.5\tabcolsep}r{.5\tabcolsep}){33-34}
   &  &  & \multicolumn{4}{c}{Gen} & \multicolumn{4}{c}{Knw} & \multicolumn{4}{c}{OCR} & \multicolumn{4}{c}{Vis} &  & \multicolumn{5}{c}{Com} & \multicolumn{3}{c}{Sci} & \multicolumn{1}{c}{Knw} & \multicolumn{3}{c}{Rdg} &  \\
\bottomrule
\end{tabular*}

\end{table}
\clearpage

\appxcenteredpage
\begin{table}[p]
  \centering
  \caption{
    \textbf{Per-benchmark results for \cref{fig:capacity:sizes}.}
    All runs train to 64 TPP.
  }
  \label{tab:appx:per-benchmark:sizes}
  
\fontsize{6}{7}\selectfont
\setlength{\tabcolsep}{1.8pt}
\renewcommand{\arraystretch}{1.0}
\begin{tabular*}{\linewidth}{@{\extracolsep{\fill}}lrrrrrrrrrrrrrrrrrrrrrrrrrrrrrrrrr@{}}
  \toprule
  \multicolumn{2}{>{\columncolor{PaperGray100!55}[.5\tabcolsep][.5\tabcolsep]}c}{\textbf{Setup}} & \multicolumn{17}{>{\columncolor{PaperBlue100!55}[.5\tabcolsep][.5\tabcolsep]}c}{\textbf{VQA (\%) $\uparrow$}} & \multicolumn{13}{>{\columncolor{PaperGreen100!55}[.5\tabcolsep][.5\tabcolsep]}c}{\textbf{Text (\%) $\uparrow$}} & \multicolumn{2}{>{\columncolor{PaperGray100!55}[.5\tabcolsep][.5\tabcolsep]}c}{\textbf{Held-out loss $\downarrow$}} \\
  \cmidrule(l{.5\tabcolsep}r{.5\tabcolsep}){1-2}\cmidrule(l{.5\tabcolsep}r{.5\tabcolsep}){3-19}\cmidrule(l{.5\tabcolsep}r{.5\tabcolsep}){20-32}\cmidrule(l{.5\tabcolsep}r{.5\tabcolsep}){33-34}
   &  &  & \multicolumn{4}{c}{Gen} & \multicolumn{4}{c}{Knw} & \multicolumn{4}{c}{OCR} & \multicolumn{4}{c}{Vis} &  & \multicolumn{5}{c}{Com} & \multicolumn{3}{c}{Sci} & \multicolumn{1}{c}{Knw} & \multicolumn{3}{c}{Rdg} &  \\
\bottomrule
\end{tabular*}

\end{table}
\clearpage

\appxstackedpage
\begin{table}[p]
  \centering
  \caption{
    \textbf{Per-benchmark results for \cref{fig:practical:free-pretraining}.}
    All runs use a 1.7B backbone and the same $2.7\times10^{21}$-FLOP multimodal stage; Pretraining is the encoder's pretraining compute in FLOPs.
  }
  \label{tab:appx:per-benchmark:free-pretraining}
  
\fontsize{6}{7}\selectfont
\setlength{\tabcolsep}{1.8pt}
\renewcommand{\arraystretch}{1.0}
\begin{tabular*}{\linewidth}{@{\extracolsep{\fill}}lrrrrrrrrrrrrrrrrrrrrrrrrrrrrrrrrr@{}}
  \toprule
  \multicolumn{2}{>{\columncolor{PaperGray100!55}[.5\tabcolsep][.5\tabcolsep]}c}{\textbf{Setup}} & \multicolumn{17}{>{\columncolor{PaperBlue100!55}[.5\tabcolsep][.5\tabcolsep]}c}{\textbf{VQA (\%) $\uparrow$}} & \multicolumn{13}{>{\columncolor{PaperGreen100!55}[.5\tabcolsep][.5\tabcolsep]}c}{\textbf{Text (\%) $\uparrow$}} & \multicolumn{2}{>{\columncolor{PaperGray100!55}[.5\tabcolsep][.5\tabcolsep]}c}{\textbf{Held-out loss $\downarrow$}} \\
  \cmidrule(l{.5\tabcolsep}r{.5\tabcolsep}){1-2}\cmidrule(l{.5\tabcolsep}r{.5\tabcolsep}){3-19}\cmidrule(l{.5\tabcolsep}r{.5\tabcolsep}){20-32}\cmidrule(l{.5\tabcolsep}r{.5\tabcolsep}){33-34}
   &  &  & \multicolumn{4}{c}{Gen} & \multicolumn{4}{c}{Knw} & \multicolumn{4}{c}{OCR} & \multicolumn{4}{c}{Vis} &  & \multicolumn{5}{c}{Com} & \multicolumn{3}{c}{Sci} & \multicolumn{1}{c}{Knw} & \multicolumn{3}{c}{Rdg} &  \\
\bottomrule
\end{tabular*}

\end{table}

\begin{table}[p]
  \centering
  \caption{
    \textbf{Per-benchmark results for \cref{fig:capacity:late-onset}.}
    Each run continues a 1.7B text-only model from 1T to 1.33T token-equivalents.
  }
  \label{tab:appx:per-benchmark:midtraining}
  
\fontsize{6}{7}\selectfont
\setlength{\tabcolsep}{1.8pt}
\renewcommand{\arraystretch}{1.0}
\begin{tabular*}{\linewidth}{@{\extracolsep{\fill}}lrrrrrrrrrrrrrrrrrrrrrrrrrrrrrrrr@{}}
  \toprule
  \multicolumn{1}{>{\columncolor{PaperGray100!55}[.5\tabcolsep][.5\tabcolsep]}c}{\textbf{Setup}} & \multicolumn{17}{>{\columncolor{PaperBlue100!55}[.5\tabcolsep][.5\tabcolsep]}c}{\textbf{VQA (\%) $\uparrow$}} & \multicolumn{13}{>{\columncolor{PaperGreen100!55}[.5\tabcolsep][.5\tabcolsep]}c}{\textbf{Text (\%) $\uparrow$}} & \multicolumn{2}{>{\columncolor{PaperGray100!55}[.5\tabcolsep][.5\tabcolsep]}c}{\textbf{Held-out loss $\downarrow$}} \\
  \cmidrule(l{.5\tabcolsep}r{.5\tabcolsep}){1-1}\cmidrule(l{.5\tabcolsep}r{.5\tabcolsep}){2-18}\cmidrule(l{.5\tabcolsep}r{.5\tabcolsep}){19-31}\cmidrule(l{.5\tabcolsep}r{.5\tabcolsep}){32-33}
   &  & \multicolumn{4}{c}{Gen} & \multicolumn{4}{c}{Knw} & \multicolumn{4}{c}{OCR} & \multicolumn{4}{c}{Vis} &  & \multicolumn{5}{c}{Com} & \multicolumn{3}{c}{Sci} & \multicolumn{1}{c}{Knw} & \multicolumn{3}{c}{Rdg} &  \\
\bottomrule
\end{tabular*}

\end{table}
\end{landscape}

  \FloatBarrier
  \FloatBarrier
\fi

\end{document}